\documentclass[twocolumn, switch]{article} %

\usepackage{preprint}

\usepackage{amsmath, amsthm, amssymb, amsfonts}

\usepackage[numbers,square]{natbib}
\usepackage[utf8]{inputenc}	%
\usepackage[T1]{fontenc}	%
\usepackage{xcolor}		%
\usepackage{url}                %
\usepackage{xurl}               %
\usepackage[colorlinks = true,
            linkcolor = purple,
            urlcolor  = cyan,
            citecolor = cyan,
            anchorcolor = black]{hyperref}	%

\AtBeginEnvironment{thebibliography}{\hypersetup{linkcolor=cyan,urlcolor=cyan}}
\AtEndEnvironment{thebibliography}{\hypersetup{linkcolor=purple,urlcolor=cyan}}
\usepackage{booktabs} 		%
\usepackage{nicefrac}		%
\usepackage{microtype}		%
\usepackage{lineno}		%
\usepackage{graphicx}		%
\usepackage{float}			%
\usepackage{placeins}		%

\usepackage{siunitx}
\usepackage{array}
\usepackage{makecell}
\usepackage{multirow}
\usepackage{caption}
\usepackage{subcaption}
\usepackage{etoolbox}
\usepackage{adjustbox}
\usepackage{tabularx}
\usepackage{colortbl}
\usepackage[breakable,skins]{tcolorbox}
\tcbuselibrary{listings, breakable}
\usepackage{listings}

\usepackage{newfloat}
\DeclareFloatingEnvironment[name={Supplementary Figure}]{suppfigure}
\usepackage{sidecap}
\sidecaptionvpos{figure}{c}
\AtBeginEnvironment{figure}{%
  \setlength{\abovecaptionskip}{4pt}%
  \setlength{\belowcaptionskip}{\baselineskip}%
  \setlength{\intextsep}{0pt}%
  \setlength{\textfloatsep}{\baselineskip}%
}
\AtEndEnvironment{figure}{%
  \setlength{\abovecaptionskip}{4pt}%
  \setlength{\belowcaptionskip}{2pt}%
  \setlength{\intextsep}{4pt plus 1pt minus 1pt}%
  \setlength{\textfloatsep}{4pt plus 1pt minus 1pt}%
}

\usepackage{titlesec}
\titlespacing\section{0pt}{6pt}{4pt}
\titlespacing\subsection{0pt}{5pt}{3pt}
\titlespacing\subsubsection{0pt}{4pt}{2pt}

\definecolor{gold}{RGB}{212,175,55}
\definecolor{silver}{RGB}{192,192,192}

\newtcolorbox{promptbox}{
  enhanced,
  breakable,
  colback=gray!10,
  colframe=black!50,
  fontupper=\ttfamily\footnotesize,
  listing only,
  listing options={
    basicstyle=\ttfamily\footnotesize,
    breaklines=true,
    showstringspaces=false,
    columns=flexible
  },
  left=2pt, right=2pt, top=2pt, bottom=2pt,
}

\usepackage{tikz}
\definecolor{gsblue}{HTML}{4285F4}
\DeclareRobustCommand{\gsicon}{%
	\begin{tikzpicture}[baseline=-0.35em]
	\draw[gsblue, fill=gsblue] (0,0) circle [radius=0.16];
	\node[white] at (0,0) {\fontfamily{phv}\selectfont\bfseries\tiny G};
	\end{tikzpicture}%
}
\newcommand{\scholarauthorA}{vU6oBhwAAAAJ}  %
\newcommand{\scholarauthorB}{rn7gJxMAAAAJ}  %
\newcommand{\scholarA}{\hspace{0.25em}\href{https://scholar.google.com/citations?user=\scholarauthorA}{\gsicon}}
\newcommand{\scholarB}{\hspace{0.25em}\href{https://scholar.google.com/citations?user=\scholarauthorB}{\gsicon}}

\title{Retrieval Augmented (Knowledge Graph), and Large Language Model-Driven Design Structure Matrix (DSM) Generation of Cyber-Physical Systems}
\newcommand{\shorttitle}{Retrieval Augmented (KG), and LLM-Driven DSM Generation of CPS}
\usepackage{authblk}

\author[1\thanks{\tt{sinan.bank@colostate.edu}}]{Hasan Sinan Bank\scholarA}
\author[1]{Daniel R. Herber, PhD\scholarB}

\affil[1]{Department of Systems Engineering, Colorado State University, Fort Collins, CO 80523, USA}

\begin{document}

\twocolumn[ %
  \begin{@twocolumnfalse} %

\maketitle

\begin{abstract}
We explore the potential of Large Language Models (LLMs), Retrieval-Augmented Generation (RAG), and Graph-based RAG (GraphRAG) for generating Design Structure Matrices (DSMs). We test these methods on two distinct use cases---a power screwdriver and a CubeSat with known architectural references---evaluating their performance on two key tasks: determining relationships between predefined components, and the more complex challenge of identifying components and their subsequent relationships. We measure the performance by assessing each element of the DSM and overall architecture. Despite design and computational challenges, we identify opportunities for automated DSM generation, with all code publicly available for reproducibility and further feedback from the domain experts.
\end{abstract}
\keywords{retrieval augmented knowledge graph \and natural language processing \and architecture synthesis \and design structure matrix}
\vspace{0.35cm}

  \end{@twocolumnfalse} %
] %

\noindent\textbf{Abbreviations:} DSM---Design Structure Matrix; LLM---Large Language Model; AI---Artificial Intelligence; CAD---Computer Aided Design; CPS---Cyber-Physical System; NLP---Natural Language Processing; RAG---Retrieval Augmented Generation; MBSE---Model-Based Systems Engineering; PLM---Product Lifecycle Management

\section{Introduction}
Since the Industrial Revolution, humanity has increasingly created sophisticated technological tools that shape our world, inspired by natural patterns and scientific advancements. As technology has evolved, the growing complexity of these developments has necessitated more advanced approaches to engineering and design \cite{incose2023incose}. This complexity prompted the establishment of formalized education programs, with the University of Michigan awarding the first MSc \cite{peckham1994making} in 1859 and Yale University granting the first PhD \cite{geiger2017advance} in 1861, while concurrently spurring the development of foundational conceptual frameworks. For decades, engineering and scientific disciplines have driven the pursuit of understanding technologies and their interrelated parts. In response to these needs, significant frameworks emerged, including von Bertalanffy's introduction of general system theory \cite{von1972history} in the 1930s, and Bell Laboratories coining the term "systems engineering" in the 1940s \cite{honour2018historical, hossain2020historical}. As these foundational concepts evolved to address increasingly integrated technologies, they have become particularly relevant for today's complex systems. Building on this historical foundation, modern architectural representation methods for cyber-physical systems (CPS) require clear definitions of systems, complexity, and architecture to address today's design challenges. \par
\begin{figure*}[htbp]
  \centering
  \includegraphics[width=1\textwidth]{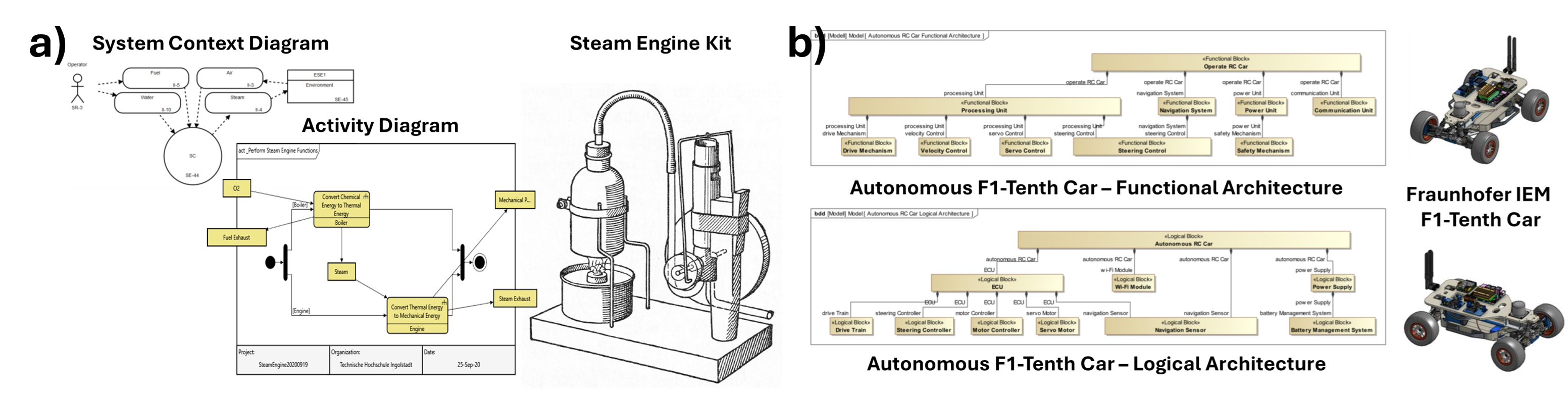}
  \caption{Architectural Representations of Sample Systems \textbf{a)} Steam Engine \cite{di2021evaluating} \textbf{b)} F1Tenth Autonomous Car \cite{von2024toward}}
  \label{fig:1}
 \end{figure*}
To effectively examine modern architectural representation methods, particularly in the context of CPS, we need to clarify the definitions for the core concepts of systems engineering, such as systems, complexity, and architecture. The term "system" has over 180 definitions, applied in various contexts, and the notion of "complexity" is similarly multifaceted \cite{dori2017system}. Our goal in providing these definitions is to establish clarity regarding our perspective, avoiding the ambiguity often associated with phrases such as "I can't define it for you, but I know it when I see it." \cite{corning1998complexity} A complex system has many elements or entities that are highly interrelated, interconnected, or interwoven \cite{crawley2015system}. This complexity is driven into systems by demanding more in terms of function, performance, robustness, and flexibility, as well as through the increasing need for systems to work together and interconnect. Such characteristics create challenges not only in development but also in comprehension and management. \par
Similar to "system" and "complexity," architecture has numerous definitions. Its origins span various industries, but in the context of systems, it is best described as "an abstract description of the entities of a system and the relationship between those entities" \cite{whitney2004influence} or as defined in ISO/IEC 42010, "fundamental concepts or properties of a system in its environment embodied in its elements, relationships, and in the principles of its design and evolution." Every system has an architecture that may arise through deliberate de novo design, evolution from previous designs with legacy constraints, adherence to regulations and standards, accretion of smaller systems, or dialogue between stakeholders and architects. Architecture strongly influences system behavior, determining both its basic functions and other properties such as durability, maintainability, and flexibility - collectively called the "ilities." \cite{de2011engineering} \par
When we examine the literature, the formation of architectural representation mostly deals with qualitative and hard-to-measure aspects such as function, allocation, structure, and physical elements. Within the context of architecting or the definition of system architecture, the architect finds a design to solve the problem from requirements to physical aspects while overlaying these in slightly different perspectives. Maier \cite{maier2009art}  characterizes architecting methodologies as both art and science - where science is embodied in normative and rational methods, while art manifests in participative and heuristic approaches, particularly when dealing with "immeasurables" and complex system conceptualization. Buede \cite{buede2024engineering}  further emphasizes this duality, demonstrating how system architectures comprise functional (what the system must do), physical (resource partitioning), and allocated (mapping functions to resources) architectures that, while distinct, must be "developed in parallel but with close interaction" to ensure meaningful integration. \par
These architectural challenges have become particularly evident in CPS, which represent one of the most complex integration challenges in modern engineering. As defined by NSF \cite{NSF2024}, the defining characteristic of CPS is the "seamless integration of computation algorithms and physical hardware," while NIST \cite{NIST2017} elaborates that "Cyber-Physical Systems (CPS) are smart systems that include co-engineered interacting networks of physical and computational components." VDI \cite{graessler2020new} (Verein Deutscher Ingenieure) describes CPS as an evolution beyond traditional mechatronic systems by incorporating internet connectivity and adaptive properties - they are defined as "interconnected mechatronic systems that are additionally connected to the Internet of Things and Services and adapt their properties during operation. This enables them to communicate with each other and use Internet services." This connectivity between physical processes and digital networks creates diverse applications with high economic potential and innovative strength. \par
The evolution of systems from mechanical to cyber-physical has introduced new challenges in architectural representation and analysis, as illustrated in Fig. \ref{fig:1}. Using the steam engine \cite{di2021evaluating} and F1Tenth Autonomous Car \cite{von2024toward} as representative examples, we can see how system architectures have evolved from purely mechanical representations to complex cyber-physical integrations. As these systems integrate computational elements, networking capabilities, and adaptive behaviors, we must examine how traditional and emerging methods address their architectural representation. \par
The challenges of representing and analyzing such CPS architectures motivate our exploration of new approaches for automated architectural generation. The following sections examine key architectural representation methods and automation techniques, starting from graph-based approaches while examining its matrix-based derivatives such as Design Structure Matrix (DSM), and expanding to language-based methods. This background establishes the foundation for our proposed methodology that combines baseline models and augmented techniques to generate DSMs that capture the complexity of modern CPS. \par
\section{Literature Review}
\subsection{The Architectural Representation}
Modern CPS present significant challenges in architectural representation and analysis, particularly in generating effective architectural models during the design phase because of their inherent complexities. Researchers have approached this challenge from various angles, with most studies focusing on analyzing existing architectural structures. Among these contributions, Raz \cite{raz2018system} introduced a notable framework for architecture design space characterization that integrates decision alternatives across functional, physical, and allocational spaces. Their work demonstrated that the traditional dichotomous treatment of system architecture and design trade-off analysis is insufficient for complex systems with interacting and interwoven functions and elements. However, their approach, while providing a substantial framework for representing and analyzing architectural design space, focuses primarily on characterizing and evaluating architectural alternatives rather than their initial formation. \par
To address these architectural representation challenges, researchers have developed numerous methodologies for modern complex systems. Early efforts focused on creating integrated model environments, while later work explored specialized analysis techniques. For example, the Total System Model (TSM) concept introduced by Bajaj \cite{bajaj2017graph} provides a graph-based digital blueprint that federates information from engineering models across multiple repositories. Schummer \cite{schummer2022approach} focused on developing a schema and MBSE modeling approach for using graph databases to analyze system architectures. However, these approaches face practical implementation challenges, particularly when integrating existing engineering tools and processes. As Chadzynski \cite{chadzynski2018enhancing} demonstrated, effective automation requires not only modeling capabilities but also seamless integration with Product Lifecycle Management (PLM) systems where critical implementation details reside. \par
While the above approaches focus on model-based representations, another significant contribution to architectural analysis comes from matrix-based methods. The DSM, introduced by Steward in the 1960s, is a compact, matrix-based representation of a system that is particularly valuable for analyzing and managing complex interdependencies \cite{steward1981design}. While structurally similar to an adjacency matrix from graph theory, a DSM's primary application is in architectural analysis, such as identifying modules through clustering or managing process flows by reordering elements (e.g., to minimize feedback loops). However, these approaches face significant practical limitations. For example, as highlighted by Eppinger \cite{eppinger2012design}, multiple real-world cases demonstrate the labor-intensive nature of DSM creation: the development process of Pratt \& Whitney's jet engine required four months of doctoral student work to extract the necessary DSM interactions. Creating a DSM with 84 components for Xerox's iGen3 Digital Printing System architecture required 140 person-hours. Even more notably, at NASA, a master's student spent five months researching seven robotic spacecraft missions to generate the Technology Risk DSM for Mars Pathfinder Technology Readiness Assessment. These cases underscore a persistent challenge: while DSM generation is important for understanding the system architecture, the process remains labor-intensive and time-consuming. \par
Additional challenges exist with domain-specific language methods, such as SysML, where users must understand specific structures in a generic manner while deriving their own extensions for specific uses \cite{santos2023survey}. This requirement for significant expert input makes scalability and the general use of these approaches persistently challenging. While some research studies claim to automate trade studies by combining PLM and SysML models, they often fail to address the fundamental challenge identified by Chadzynski \cite{chadzynski2018enhancing}: the need to semantically connect elements across different modeling domains while maintaining traceability throughout the Requirements-Functional-Logical-Physical (RFLP) core. \par
All of these examples highlight two critical gaps: 1) irrespective of the state of the CPS (e.g., physically existing or fresh in the initial design phase), extensive time and human expertise are required for the generation of systems architecture; and 2) integration challenges between system models, implementation data, and existing PLM workflows such as computer-aided design, engineering, and manufacturing. The labor-intensive nature of these processes is demonstrated in industry practice, where generating architectural representations can take months of expert work while maintaining semantic connections and traceability across different modeling domains, data structures, and modalities remains a persistent challenge. \par
All of these gaps suggest the need for a generalizable and automated approach to generating architectural representations that can reduce manual effort while maintaining accuracy and semantic compatibility with existing practices. The emergence of advanced language models and multi-modal capabilities offers promising new directions for addressing these challenges, particularly when combined with proven approaches for integrating PLM systems and maintaining traceability across the development life-cycle. \par
These gaps in current architectural representation methods have led researchers to investigate automated approaches to address the challenges of manual effort, scalability, and integration. These automation efforts have primarily followed two directions: graph-based methods that leverage mathematical foundations for structural representation and analysis and language-based methods that harness semantic understanding for architectural generation. Each approach offers distinct advantages while addressing different aspects of the identified challenges. The following sections examine these approaches in detail, beginning with graph-based methods that have historically provided a foundation for system representation. \par
\subsection{Automated Architecture Generation}
\subsubsection{Graph-based Methods}
Graph theory provides a rigorous mathematical foundation for representing and analyzing relationships between components in a system. Through its fundamental concepts of vertices (nodes) and edges (connections), it enables engineers and scientists to model complex systems from physical structures to information networks, excelling in representing both qualitative relationships (e.g., functional dependencies, spatial arrangements) and quantitative metrics (e.g., flow rates, connection strengths). The flexibility of its components allows for the depiction of diverse properties and relationships, while the graph framework provides a holistic overview. Moreover, its mathematical rigor and structural adaptability facilitate seamless translation into formats such as matrices for computational purposes, preserving inherent modularity and adaptation, solidifying its role as a foundational apparatus across domains. \par
Building on this foundation, graph-theoretic approaches provide a systematic way to generate and analyze system architectures. While research studies have investigated methods ranging from transforming generated shapes into simulation models \cite{hooshmand2012steps} to intricate relationship mapping through generative algorithms \cite{khetan2015managing} and Boolean satisfiability problems \cite{munzer2013automatically}, the difficulties are especially pronounced in CPS. In these domains, component relationships must satisfy multiple constraints simultaneously, extending beyond simple feasibility constraints \cite{wyatt2014scheme} to include uniqueness requirements \cite{schmidt1998graph} and the need to accommodate various models and design variables in the optimization process \cite{deshmukh2015bridging}. These multi-domain relationships and constraints can make traditional graph representations difficult to interpret and analyze without a layout algorithm (e.g., ForceAtlas2  \cite{jacomy2014forceatlas2}, YifanHu \cite{hu2005efficient}, etc.) or software tool (e.g., Gephi \cite{ICWSM09154}), especially during initial architectural formation, as the visual complexity and the interactions increase with the number of components or geometrical elements (e.g., vertex, edges, faces, etc.) of each part. \par
An alternative to these complex node-link visualizations is to use a matrix representation, such as the previously introduced DSM. The DSM excels at representing structural relationships and enabling quantitative analysis by maintaining clarity through its predefined and structured tabular format. This approach inherently avoids the visual clutter of node-link diagrams in large-scale systems, without requiring a layout algorithm or tool. However, despite these advantages in representing architectures, capturing the initial semantic information and domain expertise required to create the DSM remains a significant challenge. DSMs face inherent constraints in the initial formation of the architecture for capturing semantic complexity and contextual information---all essential elements for comprehensive architectural understanding. For example, Petnga's work \cite{petnga2019graph} demonstrates the analytical power of graph-based approaches once implemented, and it also reveals key challenges in the initial capture and encoding of semantic information. His framework requires significant manual effort to extract and transform SysML model data into labeled property graphs, with users needing expertise across multiple domains, including systems modeling, graph databases (e.g., Neo4j), and specialized query languages (e.g., Cypher). Peterson \cite{peterson2015understanding} similarly acknowledges these expertise barriers, noting that while SysML offers robust modeling capabilities, its limited accessibility hampers wider comprehension of the systems-of-interest. These challenges in capturing and encoding semantic information point to a fundamental limitation in graph-based or DSM approaches to initial architectural generation. As systems grow in complexity, incorporating language-based methods into the architectural generation process may provide the semantic richness needed to complement the structural precision of DSM or graph-based methods.
\subsubsection{Language-Based Methods}
Chomsky's transformational grammar \cite{chomsky2002syntactic} laid the groundwork for understanding how language structure might relate to system architecture. By hypothesizing that all human languages share an underlying structure consisting of three fundamental components---a noun as the instrument of action, a verb describing the action, and a noun as the object of the action---Chomsky had already established a parallel to what we would recognize as key elements of system architecture: form, process, and operand \cite{crawley2015system}. This foundational insight about structural parallels between language and systems has evolved significantly over time. For example, more recent work by Akay and Kim \cite{akay2021reading} builds on this foundation using advanced language models to demonstrate how design knowledge embedded in textual documentation can be systematically extracted and structured through their "Design Reading" system. Their research shows that when functional requirements (FRs) and design parameters (DPs) are expressed in natural language documentation, they can be automatically identified and hierarchically organized through a recursive decomposition process using question-answering techniques. For instance, they demonstrated how BERT \cite{devlin2018bert}, fine-tuned on question-answering tasks, can extract the highest-level FRs and corresponding DPs, then recursively decompose them into structured hierarchies. This systematic approach to processing design documentation extends Ulrich's framework of product architecture \cite{ulrich1995role} as a scheme mapping function to physical components. From Chomsky's initial insights about language structure to Akay and Kim's demonstration that language models can automatically extract and structure design knowledge from text, we see an evolution in understanding how language, having evolved to describe parts and their relationships, serves as more than a descriptive tool---it potentially encodes the architectural relationships that define system structure. The semantic domain of language has proven to be a faithful mirror of the functional domain of design, allowing modern language processing systems to automatically discern and represent system architectures while preserving their essential relationships and hierarchies. In many cases, natural language processing (NLP), especially LLMs, is utilized to translate existing textual information about an engineering system to a target format such as Enhanced Function-Means (E F-M) Modeling \cite{panarotto2022using} as an extended derivative of axiomatic design \cite{suh1990principles} or Unified Modeling Language (UML) diagrams \cite{gomez2024large} as the basis of system modeling language (SysML v1), and Function Behavior Structure (FBS) of systems \cite{wang2023task}, to name a few. \par
In a similar sense, recent efforts have been to integrate NLP methods into model-based systems engineering workflows, i.e., SysML-based approaches. By analyzing textual descriptions of systems, NLP methods can help automate certain aspects of the model-based systems engineering process, reducing the manual workload and helping to scale architecture generation. For example, Zhong \cite{zhong2023natural} investigated automatically generating SysML v1 models, focusing particularly on structure and requirement diagrams through natural language processing. While their approach achieved automation in diagram generation, the preparation steps require significant manual work due to their reliance on traditional NLP approaches. Additionally, current research demonstrates a significant increase in the utilization of large language models, mostly to interpret one form of information to another. Furthermore, DeHart \cite{dehart2024leveraging} explores leveraging Large Language Models (LLMs) for direct interaction with SysML v2, demonstrating how LLMs can serve as an efficient interface for model manipulation and reduce the complexity typically associated with API interactions. While their work shows promising results regarding efficiency and workflow integration, it lacks specific metrics for analyzing LLM output quality and does not address critical challenges such as hallucination prevention and validation methodology for ensuring model accuracy---key considerations for implementing LLMs in technical domains. VanGundy et al. demonstrated the effectiveness of combining LLMs with graph databases for requirement discovery and traceability, showing improved accuracy through validated database queries compared to LLM approaches \cite{vangundy2024requirement}. \par
These approaches focus on translating textual information into various modeling formats such as graphs, UML diagrams, and SysML models, another important line of research has explored automated generation of architectural representations through matrix-based approaches. One of the first studies to automatically generate DSMs from design matrices was Dong and Whitney's work in 2001, which converted requirement-parameter relationships into component dependencies. However, their approach was limited to square matrices with an equal number of requirements and parameters. Wilschut \cite{wilschut2018generation} demonstrated Multi-Domain Matrix (MDM) generation from structured textual specifications in 2018, though their method required manual conversion of natural language documentation into their prescribed grammatical format. Koh \cite{koh2024auto} recently explored automated DSM generation using LLMs to reduce the traditional time-intensive process of manual DSM creation, though the approach relies on predefined questions, off-the-shelf LLMs, and lacks shared source code or query history. \par
The evidence demonstrates the potential of natural language processing approaches to translate one form of information to another and create architectural representations while reducing the manual labor necessary to know the domain-specific languages. The question remains: what approaches can best ensure reliability and semantic consistency in these translations while minimizing manual intervention in the process, particularly regarding validating LLM outputs and preventing hallucinations? As we explore the challenges and possibilities of automating architectural generation, we must recognize that no single approach can solve every aspect of system complexity. While valuable, many existing methods struggle to scale effectively as system complexity increases. Our goal in the following sections is not to present a final solution to all the challenges in system architecture generation but rather to offer an approach that can help streamline certain aspects of architectural formation and analysis. \par
Given the complexity of such modern systems, it is natural to approach their representation from an architectural perspective, aiming to understand and resolve system issues through this lens. However, the key question is how to implement this solution effectively. In the remainder of this paper, we present language- and augmented language models with their derivatives for doing so. We will highlight existing methodologies and our approach, comparing them programmatically with the available ones, while providing source code and log files of the experiments \cite{bankh2024xlm}.
\section{Method}
In the following sections, we examine various language model variants (LLM, Augmented Language Model-ALM) that enhance this process, their implementation specifics, and their contributions to addressing CPS architectural challenges, as shown in Fig. \ref{fig:3}.
\subsection{Large Language Models-LLMs}
As highlighted in the literature, LLMs have demonstrated significant capabilities in processing and understanding natural language text, with emerging abilities to extract semantic relationships from unstructured content. The key aspect that makes LLMs particularly interesting for systems engineering is their ability to process unstructured data (e.g., technical documents) and extract meaningful relationships, potentially requiring relatively less amount of pre-defined templates or structured formats.\par
\begin{figure}[htbp]
 \centering
 \includegraphics[width=0.5\textwidth]{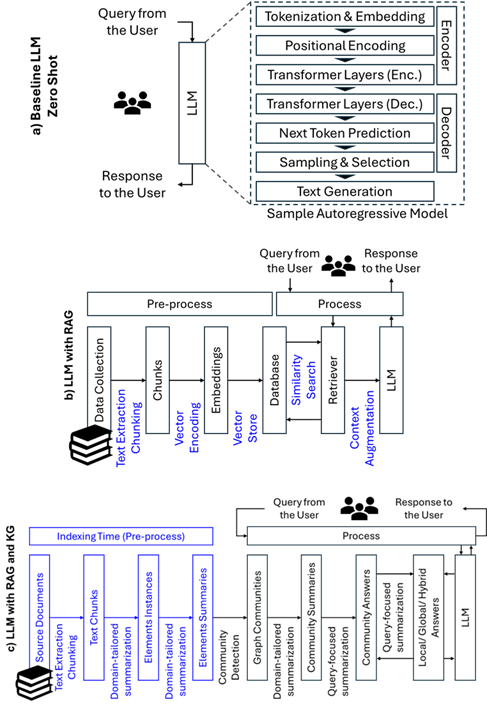}
 \captionsetup{width=\linewidth}
 \caption{Overview of \textbf{a)} Baseline LLM \textbf{b)} Baseline LLM with RAG \cite{lewis2020retrieval} \textbf{c)} LLM with KG and RAG \cite{edge2024local}}
 \label{fig:3}
\vspace{-\baselineskip}
\end{figure}
Many of these models are autoregressive, trained on extensive text corpora (e.g., FineWeb \cite{huggingface2023fineweb}) to predict the conditional probability of token sequences. To achieve this, they utilize Transformer architectures with self-attention mechanisms \cite{vaswani2017attention} to effectively capture long-range dependencies in text. Well-known foundational models, such as DeepSeek \cite{guo2025deepseek,liu2024deepseek}, ChatGPT \cite{achiam2023gpt}, Llama \cite{dubey2024llama}, and Claude \cite{anthropic2024claude}, leverage these architectures with vast datasets and scaling techniques \cite{kaplan2020scaling}. This enables them to perform a wide array of language-based downstream tasks, such as translation, summarization, and reasoning, by generalizing patterns learned during pretraining to novel inputs \cite{brown2020language}. However, the underlying statistical nature of these models introduces innate limitations for systems engineering applications. As Buchmann et al. succinctly state, "LLMs are statistical, as opposed to conceptual models." \cite{buchmann2024expectations} This core characteristic leads to two critical issues: hallucination and semantic drift. While LLMs can generate fluent and plausible-looking output, their probabilistic foundation means they may invent relationships or components that do not exist or fail to capture critical, deterministic technical dependencies accurately.\par
When applied to systems engineering tasks, the transformer processes engineering-relevant data (e.g., technical documents, graphs, geometrical information, etc.) through tokenization, converting text into numerical representations that maintain contextual relationships. The attention mechanism enables the model to assign varying levels of importance to different parts of the input, thereby facilitating the identification of crucial technical relationships and dependencies. The model generates output through decoding that converts these numerical representations back into human-readable text while maintaining learned semantics and relationships, as shown in Fig. \ref{fig:3}.a. \par
\begin{figure*}[!t]
  \centering
  \includegraphics[width=1\textwidth]{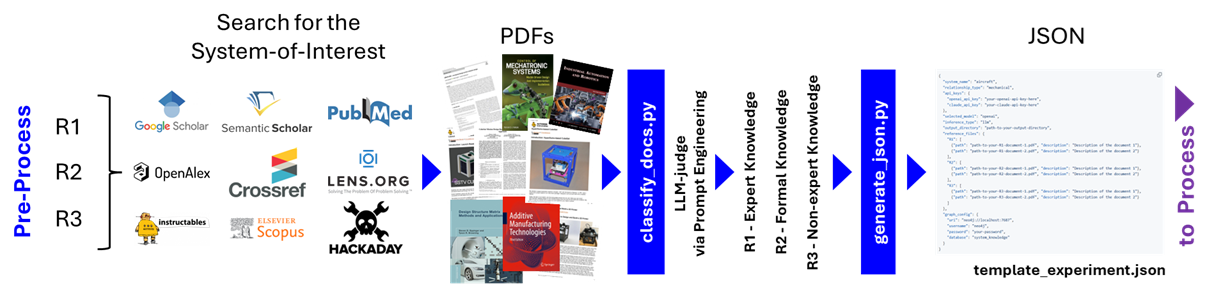}
  \caption{An Overview of Step 1 of the proposed approach}
  \label{fig:4}
\end{figure*}
\subsection{Augmented Language Models-ALMs}
ALMs represent an evolution in language model application that addresses one of the fundamental challenges in architectural generation: ensuring the accuracy and reliability of the generated content \cite{andriopoulos2023augmenting}. Extensive research has been conducted on identifying hallucinations in outputs generated by LLMs \cite{ji2023survey}, a concern that has grown increasingly significant in technical fields employing LLMs for systems architecture generation. \par
These issues in LLMs stem from multiple factors, including temporal limitations of training data, incorrect knowledge activation, and imperfect data distribution in the input space \cite{kandpal2023large,mallen2022not}. These challenges are particularly acute in technical domains (e.g., engineering system design) where precision is paramount, as inaccurate generation can propagate through subsequent errors in system design decisions. \par
Retrieval-Augmented Generation (RAG) has emerged to overcome these limitations by incorporating external knowledge sources into the generation process. This approach enhances input queries with retrieved documents from specific corpora, leveraging information retrieval mechanisms to supply relevant knowledge to generative LLMs \cite{lewis2020retrieval, guu2020retrieval}. RAG systems have progressed from specialized retrieval models to more integrated approaches that directly augment standard language models \cite{khandelwal2019generalization, borgeaud2022improving}. \par
Experimental evaluations have demonstrated RAG's empirical success across various application domains \cite{khattab2022demonstrate, ram2023context}. This approach has particular significance for architectural generation, where grounding in verified technical documentation and established system knowledge is crucial. The process encompasses both retrievals of relevant architectural documentation and validation of relationships through external knowledge bases \cite{min2023factscore}, as shown in Fig. \ref{fig:3}.b.\par
While traditional RAG approaches rely on direct text chunk retrieval and embedding-based similarity search, graph-based RAG architectures introduce a fundamentally different approach to knowledge organization and retrieval. The key innovation lies in converting source documents into a structured entity knowledge graph, where nodes represent key concepts and edges capture their relationships. This structural representation enables two critical capabilities: (1) the capture of complex relationships between concepts that may be distributed across different text chunks and (2) the ability to leverage graph modularity for efficient information organization. \par

A distinctive feature of graph-based RAG implementation is its hierarchical processing pipeline. Initially, the system processes source documents to extract entities and relationships using LLM-driven extraction techniques \cite{zhang2024causal}. These elements are then organized into a weighted graph structure where edge weights represent the strength of relationships between entities. The graph is subsequently partitioned into communities using algorithms (e.g., Leiden \cite{traag2019louvain}) to identify closely related concept clusters. This community structure enables a novel "map-reduce" approach to query processing: relevant communities are first identified, parallel summaries are generated for each community, and these summaries are then synthesized into a final response \cite{baek2023knowledge}. \par
Where traditional RAG might struggle due to its limited context windows and local relevance matching, graph-based implementations use community structure to maintain a broader context while reducing computational overhead. Practical implementations in both Neo4J \cite{neo4j_rag_application} and NebulaGraph \cite{nebulagraph_graph_rag} environments have demonstrated the scalability of this approach, particularly for queries requiring integration of distributed knowledge. \par
To realize all highlighted points in Fig. \ref{fig:3} as aligned with the details above while generating the target system's DSM, we employ a three-step procedure: \par
\textit{Step 1 - Preparation of the References and Design Configuration:} To utilize downstream LLM tasks (e.g., RAG and graph processes), we need appropriate technical context from known references. In this step, we semi-automate a procedure to extract information from well-known academic search engines (e.g., Google Scholar, Semantic Scholar, Web of Science, Scopus) and crowdsourced project repositories (e.g., Instructables, Hackaday) for the system-of-interest. Scientific literature with high citation counts typically contains accurate expert-level information, representing \textit{R1-type} knowledge that we can use for the target system-of-interest, while books commonly used in classroom instruction or industrial standards for the system-of-interest provide examples of \textit{R2-type} (formal or normative) knowledge. By "formal," we mean information typically used to teach students at various levels (e.g., from high school to graduate). Here, we utilize crowdsourced project repositories created mainly by enthusiasts, hobbyists, and makers---many of whom are experts in their fields. However, since these web pages generally aim to be used by entry-level users, their content is primarily suited for non-expert or novice users for the target system-of-interest, hence their classification as \textit{R3-type}(informal). \par
\begin{figure*}[!t]
  \centering
  \includegraphics[width=1\textwidth]{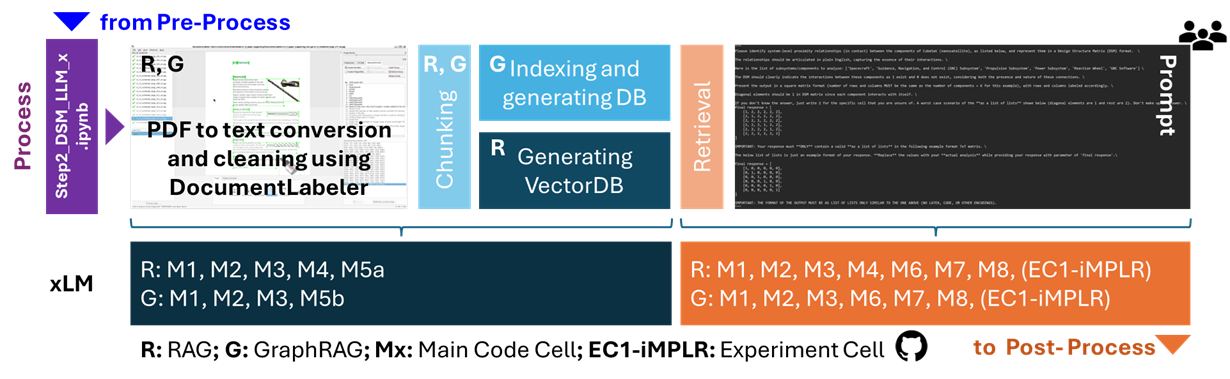}
  \caption{An Overview of Step 2 with associated repository \cite{bankh2024xlm} steps}
  \label{fig:5}
\end{figure*}
This methodological approach demonstrates that the choice of academic search engines and crowdsourced project repositories implicitly and not strictly pre-filters the types of references to be used before the document classification step. After collecting the documents, we proceed with the prompt-engineered \textit{classify\_docs.py} script to classify the downloaded documents for the specific use case. Similar to Ma \cite{ma2024exploring}, for enhanced results, we used prompt engineering while highlighting the expertise of the LLM model in our classification queries. Once classified, the documents are formatted as \textit{[Year Author] RType-Title.pdf}. We then use \textit{generate\_json.py} to create the required configuration files for \textit{Step 2 - Process} while combining the documents in the same category into a single document and listing them in the reference files parameter of the configuration file. We illustrate this workflow in Fig. 4, including a snippet of the template JSON file. Unlike Yan \cite{yan2024corrective}, where an LLM-judge verifies retrieved information during the retrieval process due to the use of the Internet as a reference, our approach pre-specifies and classifies the documents that will be used in downstream LLM tasks before the fact. \par
\textit{Step 2 - xLM Preparation and Process:} Subject to the selected method, we utilized the provided workflows in Fig. 3 (a-LLM, b-RAG, and c-GraphRAG) in this step. For RAG and GraphRAG, we first convert the documents to text and clean them by using DocumentLabeler \cite{bank2024catalogbank}. Then, we created chunks of the texts to be ingested by the target models (e.g., RAG and GraphRAG), while baseline LLM models utilized the existing knowledge of the model (without any documents or fine-tuning). For a comparable setup between RAG and GraphRAG, we used the same nominal chunk size of 1200 in both pipelines (characters in the RAG text splitter, tokens in GraphRAG), with an overlap of 300 characters for RAG and of 100 tokens (gpt-4-turbo-preview) or 150 tokens (Ollama-served models) for GraphRAG. Moreover, in GraphRAG, we autotuned the graph extraction prompts using the framework's built-in capabilities \cite{guevara2024graphrag} and indexed the target documents before sending our prompts. \par
Irrespective of the method, we asked the same prompts as shown in Appendix \ref{App:A}. While addressing the questions, we ensured that three aspects were covered in our queries: output format specification (e.g., the size of the DSM matrix and its structure), clear and concise instructions (without overlooking the specifics of the systems of interest and domain), and a first-shot example (e.g., an example reflecting uncertainties and another one to reflect the expected output format). The structure of the prompts varies based on the use case, method, and evaluation, as detailed in the supplemental information via configuration files. \par
The experiments with the ground truth consist of two scenarios (\textit{i}- with initial subcomponents and \textit{ii}-without any initial subcomponents). For (\textit{i}), where we provide the model with the list subcomponents (e.g., for a power screwdriver: ["Bit", "Transmission", "Motor", "Electrical System", "Battery Holder", "Housing", "External"]) and expect the model to determine the existing relationships (e.g., sub-system's spatial proximity, etc.) between those subcomponents, whereas in (\textit{ii}), we did not provide the list of subcomponents and allowed the model to identify the subcomponents. To do so, in (\textit{ii}), we utilized an initial prompt for identification of the components, an updated prompt, and a validator prompt, as their details are provided in Appendix \ref{App:A}. Once the model inferred the system's subcomponents by using the initial prompt, the updated prompt extracts the relationships between components, and the validator prompts checks the identified subcomponents. After the identification and validation process, the model determines the existing target relationships (e.g., proximity for power screwdriver and whole-part for CubeSat) between those parts. We provide the necessary details---such as which models and type of compute were used and why---in the Results and Discussion section while sharing our configuration files and process logs in the paper's repository. \par
Subject to the model, method, and case being run, we analyzed and visualized the results to assess the effectiveness of the models in Step 3. \par
\begin{figure*}[!t]
 \centering
 \includegraphics[width=1\textwidth]{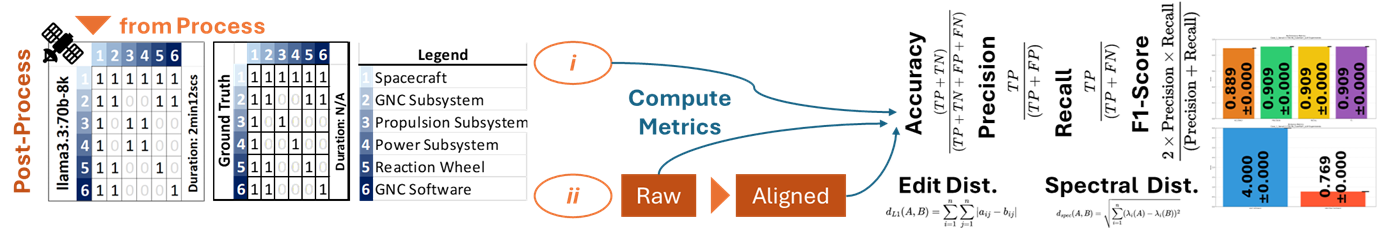}
 \caption{An Overview of Step 3 - Analysis and Visualization}
 \label{fig:6}
\end{figure*}
\textit{Step 3 - Analysis and Visualization of the Results:} It is imperative to understand the performance of the methods for the specific use cases that we have. In (\textit{i}), where we provide the model with the list of subcomponents, we analyzed the determination of the existing target relationships in a cell-by-cell manner by using the traditional metrics such as accuracy, precision, recall, and F1-Score. These metrics are calculated using true positives (\text{TP}), true negatives (\text{TN}), false positives (\text{FP}), and false negatives (\text{FN}) as follows:
\begin{align}
\text{Accuracy} &= \frac{\text{TP} + \text{TN}}{\text{TP} + \text{TN} + \text{FP} + \text{FN}} \\[1ex]
\text{Precision} &= \frac{\text{TP}}{\text{TP} + \text{FP}} \\[1ex]
\text{Recall} &= \frac{\text{TP}}{\text{TP} + \text{FN}} \\[1ex]
\text{F1-Score} &= \frac{2 \times \text{Precision} \times \text{Recall}}{\text{Precision} + \text{Recall}}
\end{align}
In (\textit{ii}), we did not provide the list of subcomponents and allowed the model to identify the subcomponents and their target relationships, we incorporated a cosine-similarity based LLM filtration followed by traditional metrics as in (\textit{i}) on a cell-by-cell basis. Moreover, due to the graph structure of DSMs, we utilized graph-based metrics to evaluate the effect of the subcomponents' alignment globally, compared to the original ground truth. We visualized all these metrics in a bar graph to clearly understand the impact of the method on the (\textit{i}-determination of the existing target relationships) or (\textit{ii}-identification of the subcomponents and determination of the existing target relationships) in Figs. \ref{fig:8} and Fig. \ref{fig:9}, while including their standard deviations. The graph-based metrics we employed include the edit distance ($d_{L1}$) and spectral distance ($d_{spec}$) between design structure matrices $A$ (ground truth) and $B$ (generated) (where $a_{ij}$ and $b_{ij}$ are individual elements of these matrices and A and B are n x n matrices):
\begin{align}
d_{L1}(A,B) &= \sum_{i=1}^n \sum_{j=1}^n |a_{ij} - b_{ij}| \\[1ex]
d_{spec}(A,B) &= \sqrt{\sum_{i=1}^n (\lambda_i(A) - \lambda_i(B))^2}
\end{align}

where $\lambda_{i}(A)$ and $\lambda_{i}(B)$ are the eigenvalues of matrices A and B as shown in Fig. 6.
\section{Use-case Studies}
We demonstrate the effectiveness of our approach through two distinct use cases and examine systems with established architectural structures from literature. There are many documented use cases across diverse domains, including micro-electromechanical systems (MEMS) \cite{akay2021reading}, aerospace control systems \cite{wang2024intelligent}, as well as automotive powertrains and manufacturing systems \cite{engel2015advancing}, among many others. For our validation, we specifically focus on architectural structures from CubeSAT systems \cite{friedenthal2017architecting} and a power tool example from Tilstra's product architecture analysis with High-Definition Design Structure Matrix (HDDSM) \cite{tilstra2012high}. These well-documented examples serve as ground truth for comparing and validating our results. While previous studies have investigated LLM effectiveness for various downstream tasks either with ground truth \cite{doris2024designqa} or without it \cite{xu2024good}, our work uses standard evaluation metrics to provide a comprehensive benchmark for local and global relationships of the DSM structures. \par
Our investigation of architectural generation for CPS would demonstrate appropriate cases since most complex CPS are iteratively designed from their existing structures rather than being designed from scratch \cite{suh2010technology,dennis2023machine}. Our approach enables us to dissect existing designs and provide an initial architecture that would lead to future structural modifications---and potentially assist in preventing obsolescence \cite{ozkan2022multi} or any other potential technical debts \cite{kleinwaks2023technical} of systems engineering efforts during the design process of CPS. While the solution provides a starting point for architectural development, the generated architectures serve as initial proposals that require expert validation, verification, and further refinement by system architects. \par
In the following sections, we detail our use cases and testing methodology. To ensure reproducibility, unlike many research studies, we provide our source code \cite{bankh2024xlm} with multiple test cases rather than claiming effectiveness based on a single use case with an insufficient number of experiments that supports the methodology. 
\subsection{Architectural Structure of Existing Systems}
The presented research examines two distinct systems that provide validation cases for our proposed methodology due to their well-documented component interactions and already-established system architectures. \par
\textbf{Power Screwdriver:} The Black \& Decker power screwdriver represents a consumer product with a documented architecture \cite{tilstra2010representing}. It consists of 42 components organized into six main subsystems (bit, transmission, motor, electrical system, battery holder, and housing). Tilstra \cite{tilstra2012high} utilized this product to demonstrate the HDDSM approach, providing detailed documentation of physical connections and functional interactions (including spatial proximity, energy transfer, material flow, and information exchange) which serve as ground truth for our proximity relationship analysis. \par
\textbf{CubeSat:} CubeSats are standardized nanosatellites based on a $10 \times 10 \times 10 \text{cm}$ form factor (1U), often scaled up (2U, 3U, etc.), serving diverse missions \cite{kitts1995initial, clarke1996picosat}. However, CubeSats face significant challenges, with failure rates of 35\% in academic missions and 25\% in industrial applications, often attributed to issues originating in the conceptual design phase \cite{girardello2024greatcube+}. To address these challenges, researchers have applied Model-Based Systems Engineering (MBSE) methodologies for formalized architectural documentation \cite{anderson2014enterprise, magone2025scalable}. Friedenthal \cite{friedenthal2017architecting} provided a SysML guide for small satellite architecture, which we use as a basis. CubeSats typically employ a modular architecture with standardized subsystems (structures/thermal, power, attitude control, communications, command/data handling, payload). This standardized structure with documented interfaces \cite{calpoly2022CDS, jamie2017basic}, especially the whole-part relationships between subsystems, makes CubeSats valuable for validating our architecture generation methods. \par
Both use cases provide valuable ground-truth architectural structures that can be used to validate generated DSMs based on the extracted information. Tilstra's power screwdriver example offers detailed component-level interactions in a consumer product context, while Friedenthal's spacecraft example demonstrates a more abstract example of a system-of-systems architectural structure in a space application context (e.g., CubeSat).
\begin{figure}[htbp]
 \centering
 \includegraphics[width=0.495\textwidth]{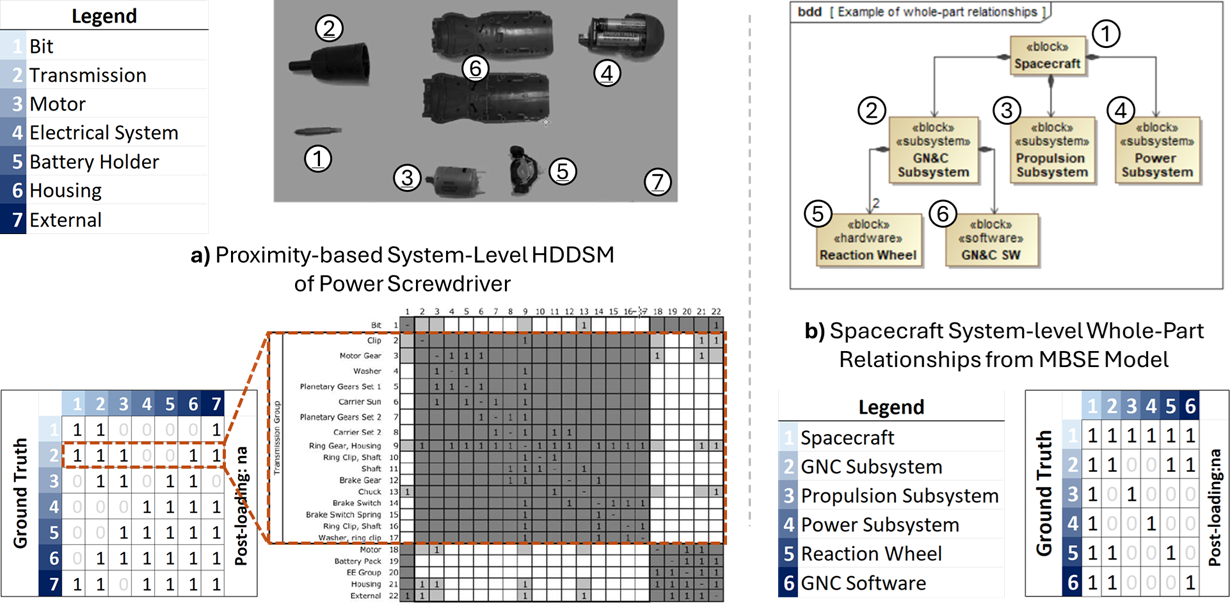}
 \captionsetup{width=\linewidth}
 \caption{Sample DSM Architecture for \textbf{a)} power screwdriver \cite{tilstra2012high} and \textbf{b)} CubeSat \cite{friedenthal2017architecting}}
 \label{fig:7}
\vspace{-1\baselineskip}
\end{figure}
\section{Results and Discussion}
Irrespective of the presented methods (LLM, RAG, or GraphRAG), the process of generating DSMs through language models involves two fundamental tasks: component identification and the determination of the existing relationship between pairs. Initially, the models must identify the relevant system components using the provided prompts and subject to the method the documentation that the use-case entails. Then, they must determine the target relationships to populate the DSM structure. At the implementation level, each of our tested methods processes these tasks differently. Basic LLMs depend exclusively on their pre-trained knowledge to identify components and classify their target relationships, while RAG utilizes chunks of relevant documents, and GraphRAG further structures this process by forming a knowledge graph from these document chunks. \par
The computational demands and model capabilities for these tasks vary significantly. GraphRAG, being the most complex approach, places the highest demands on model performance and computational resources, particularly during the knowledge graph construction (e.g., indexing phase). Therefore, we used GraphRAG performance as our initial constraint for the selection of the models as detailed in Appendix \ref{App:selection}, our reasoning is that the models capable of handling GraphRAG's demands would also perform adequately in simpler baseline LLM and RAG implementations. This approach allowed us to systematically evaluate different hardware configurations (from local GPU setups to cloud infrastructure) and model architectures (from open-source to proprietary) to identify the most suitable combinations for DSM generation. \par
To thoroughly evaluate these approaches, we conducted our experiments in two phases. First, we analyzed how each method performed in component identification, and then, the identification of the components and existing relationships separately. While doing so, we examined their end-to-end performance in generating complete DSMs. This systematic evaluation allows us to understand not just the final DSM accuracy but also where and why each approach succeeds or fails in the generation process while considering the practical constraints of computational resources and model selection (as presented in Appendix \ref{App:A}).
\subsection{Evaluations of LLM, RAG, and GraphRAG for DSM Generation}
In our experiments, we evaluated three distinct methodologies (LLM, RAG, GraphRAG) in parallel to the presented use cases (Power screwdriver and CubeSat). The presented use cases have established ground-truth architectural structures: a Black \& Decker power screwdriver's proximity (spatial relationship of the parts) architecture out of nine relationships \cite{tilstra2012high}, and a CubeSat nanosatellite system from a sample spacecraft's systems architecture \cite{friedenthal2017architecting}. Our goal is to evaluate the performance of the models for (\textit{i}) determining the existence of target relationships between the provided systems and in (\textit{ii}) identifying the components, followed by determining the relationships of identified components. For calculating the cell-level evaluation metrics discussed in Section 3 (Equations 1-4), model outputs corresponding to the "unsure" state ("2") were excluded to evaluate performance based only on confident predictions of existence ("1") or non-existence ("0"). Unsure outputs were observed primarily in GPT-4o, and in some configurations for deepseek-r1:14b, llama3.3:70b, and mixtral:8x22b; the percentages and cell counts are given in Appendix~\ref{App:remarks}, and the full per-model, per-configuration data for both tasks and both use cases are provided in the accompanying repository~\cite{bankh2024xlm}. Furthermore, to enhance the reliability of the results, we conducted each experiment five times and used the global and local evaluation metrics with their standard deviations in the following sections. \par
For the experiments described below, each API request is treated as an independent operation with no shared memory across calls. Therefore, in the context of chain-of-thought processes---implemented as model-to-model validation via a validator in \textit{ii}---it is necessary to explicitly provide both the query and its corresponding response if any further processing or validation is required (as detailed in Appendix~\ref{App:A}).

\begin{figure*}[htbp]
 \centering
 \includegraphics[width=1\textwidth]{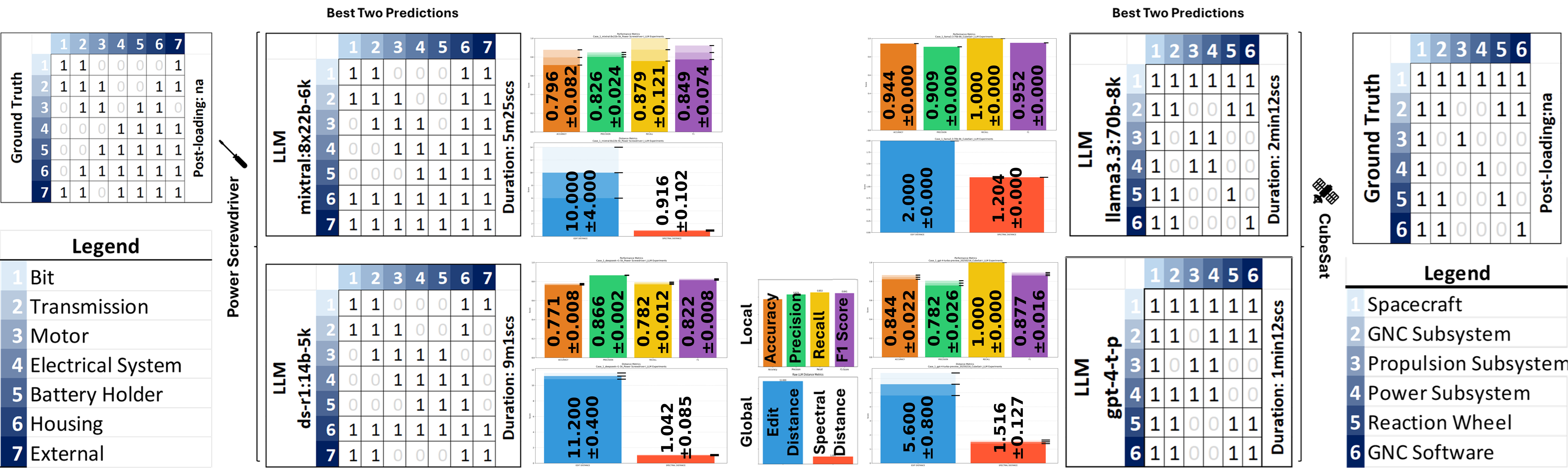}
 \caption{Best two predictions (highest mean accuracy, one configuration per model) of the models and methods for (\textit{i}) of power screwdriver and CubeSat. (Please see Table \ref{tab:case_i_cs_experiments} and Table \ref{tab:case_i_ps_experiments} in Appendix.)}
 \label{fig:8}
\end{figure*}
\textit{i- Determination of the Existing Relationships of the Pairs:} In this evaluation, we provided a list of components and explicitly indicated the types of relationships we aimed to recognize. For the power screwdriver scenario, the targeted relationships were primarily spatial, underscoring the system's physical structure. In contrast, for the CubeSat, the analysis concentrated on delineating system-level whole-part relationships. During experiments, we observed significant performance degradation when details, such as the relationship type or the specific object of interest (e.g., power screwdriver or CubeSat), were omitted. This sensitivity underscores the importance of precise prompt design, as discussed by \cite{khattab2023dspy}. Therefore, detailed prompts used in this study are provided in Appendix \ref{App:A} as supplementary material. \par
To assess relationship classification performance, we evaluated five models selected via the method detailed in Appendix \ref{App:selection}. Model sizes ranged from approximately $1.77$T parameters (OpenAI gpt-4-turbo-preview) down to $14.8$B parameters (deepseek-r1:14b). In baseline LLM experiments for spatial reasoning of the power screwdriver, mixtral:8x22b achieved the best overall performance (and best LLM performance), with an F1-Score of $0.849\pm0.074$, precision of $0.826\pm0.024$, and accuracy of $0.796\pm0.082$, despite being smaller than both GPT-4 variants ($\approx1.77$T parameters). Additionally, deepseek-r1:14b exhibited competitive accuracy ($0.771\pm0.008$) as second-best overall performance, surpassing larger models such as llama3.3:70b. These results suggest that architectural design choices and training methods may influence performance more significantly than model size alone. The high accuracy and precision observed with mixtral:8x22b and deepseek-r1:14b indicate their effectiveness in minimizing false positives---a critical factor for reliable DSM generation. \par
For the CubeSat scenario, characterized by abstract, high-level whole-part relationships, llama3.3:70b outperformed other models in baseline LLM, achieving accuracy of $0.944\pm0.000$, precision of $0.909\pm0.000$, recall of $1.000\pm0.000$, and an F1-score of $0.952\pm0.000$. The superior performance of llama3.3:70b likely stems from its capability to effectively handle higher abstraction levels, suitable for system-level analyses compared to the spatial proximity emphasis in the power screwdriver scenario, though a more detailed architectural or training data analysis would be needed to confirm these potential model specializations. \par
We further investigated RAG methods by systematically combining all reference sets (R1, R2, R3) to determine whether external knowledge retrieval improved relationship classification. Contrary to initial expectations, simply aggregating all references did not consistently enhance performance. For CubeSat, performance varied according to model size and architecture. Some configurations performed best with a single reference, while others benefited from multiple references. Notably, certain large-scale models exhibited diminished or plateaued performance when all three references (R1-R2-R3) were combined, highlighting a non-monotonic relationship between retrieved context volume and classification accuracy. Similar behavior was observed for the power screwdriver scenario. Adding more references did not reliably improve performance, indicating greater model sensitivity to specific reference configurations, resulting in significant performance variability. These findings align with the understanding that retrieval-augmented pipelines require careful tuning to avoid introducing irrelevant or conflicting information and pinpointing the exact textual elements within specific references that cause performance shifts (e.g., performance enhancements or bias) would require further detailed investigation beyond the scope of this work. Interestingly, smaller architectures sometimes benefited more from RAG than larger models, a phenomenon also observed in baseline experiments.  \par
This highlights that while the R1/R2/R3 classification showed some correlation with outcome quality, simply aggregating references based on these broad categories does not guarantee optimal performance. The specific combination of documents containing the necessary component and interaction details proved more critical. Future work could investigate more granular content-based relevance scoring for reference selection, potentially moving beyond these high-level source classifications to better predict utility for specific DSM generation tasks.
We additionally evaluated GraphRAG, integrating structured graph representations with retrieval-augmented generation. This approach utilized knowledge graphs created from different reference combinations (R1-R2 and R2-R3). GraphRAG showed notable performance gains in specific configurations. For the power screwdriver case, GraphRAG R2-R3 achieved strong performance throughout the models, significantly outperforming some of the baseline LLMs and standard RAG approaches. For the CubeSat scenario, GraphRAG R2-R3 with gpt-4-turbo-preview reached the accuracy of its RAG R2 configuration ($0.822$), which is below its best standard RAG configuration (RAG R3, $0.833$) and below its baseline LLM ($0.844$). \par
Additionally, structural similarity metrics (edit and spectral distances) revealed interesting patterns across all methods and models. For the CubeSat scenario, llama3.3:70B with baseline LLM achieved the lowest edit distance (edit distance: $2.0\pm0.0$, spectral distance: $1.204\pm0.000$), while mixtral:8x22b with baseline LLM showed competitive results for power screwdriver (edit distance: $10.0\pm4.0$, spectral distance: $0.916\pm0.102$). Various RAG implementations with specific reference combinations also demonstrated good structural correspondence depending on the model used. GPT4-Turbo-Preview with GraphRAG R2-R3 reduced the edit distance for the power screwdriver relative to its baseline ($14.0\pm1.265$ versus $15.2\pm0.980$), although its spectral distance did not improve ($1.019\pm0.115$ versus $0.819\pm0.231$). These varied results across different approaches suggest that structural alignment depends significantly on the combination of model architecture, retrieval method, and reference material selection, rather than being exclusive to any single approach. \par
All these findings are detailed in Fig. \ref{fig:8} and Appendix \ref{App:A}. \par
\begin{figure*}[htbp]
 \centering
 \includegraphics[width=\textwidth]{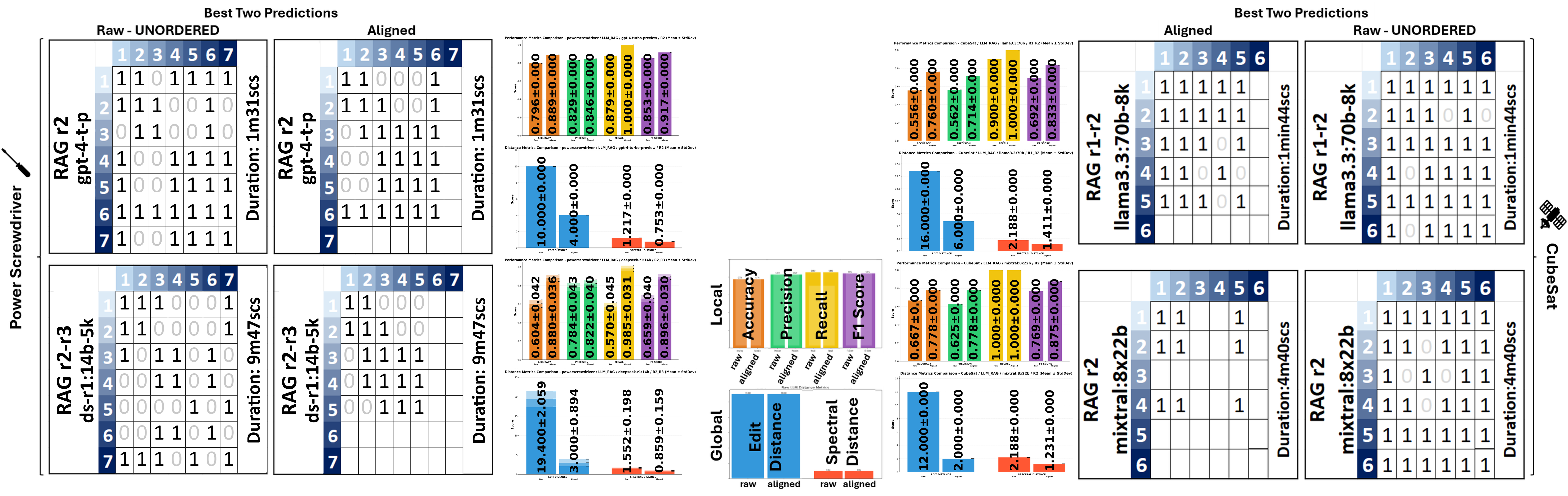}
 \caption{Best two aligned predictions (highest mean aligned accuracy, one configuration per model; llama3.3:70b: highest accuracy with five of six components aligned) of the models and methods for (\textit{ii}) of power screwdriver and CubeSat. (Please see Table \ref{tab:case_ii_raw_cs_experiments}, \ref{tab:case_ii_aligned_cs_experiments}, \ref{tab:case_ii_raw_ps_experiments}, and \ref{tab:case_ii_aligned_ps_experiments} for the detailed results and Figs. \ref{fig:11} and \ref{fig:12} for the details of raw and aligned DSMs in Appendix.)}
 \label{fig:9}
\end{figure*}
\textit{ii- Identification or the Parts and Determination of the Existing Relationships of the Pairs:} In this evaluation, models first predict the component list before determining relationship existence. Our analysis considers both raw predictions and aligned results after filtering with LLMs(achieved via cosine similarity between component name embeddings; further details are provided in the code repository \cite{bankh2024xlm}). We employ the same evaluation metrics as in \textit{i}. \par
Across 500 experiments (as $\text{\#ofModels} \times \text{\#ofRepetitions} \times \text{\#ofDocumentCombinations} \times \text{\#ofUseCases}$; where $5 \times 5 \times 1 \times 2$ in LLM; where $5 \times 5 \times 7 \times 2$ in RAG; and where $5 \times 5 \times 2 \times 2$ in GraphRAG) both for power screwdriver and CubeSat in (\textit{ii}), gpt-4-turbo-preview has consistent performance in the power screwdriver case, and it has the best raw accuracy in eight of the ten configurations and the best aligned accuracy in five of them. For our (\textit{i}) (only identification of existing relationships) and \textit{ii} (identification of components and their existing relationships) studies, the results differ significantly: for CubeSat, the experiments of (\textit{ii}) reach lower top accuracy scores than those of (\textit{i}) ($0.722$ raw, $0.778$ aligned, versus $0.944$), whereas for the power screwdriver the top scores of (\textit{ii}) ($0.878$ raw, $0.889$ aligned) exceed the top score of (\textit{i}) ($0.796$). The alignment process (filter with LLMs) in (\textit{ii}) improves the accuracy over the raw scores in 34 of the 38 comparable power screwdriver configurations, but in only 12 of the 38 CubeSat configurations, where the matching frequently covers only part of the six components. Fig. \ref{fig:9} presents the best two predictions of the complete set of experiments for both power screwdriver and CubeSat use-cases. For CubeSat, the second panel shows llama3.3:70b with RAG R1-R2, the most accurate prediction that aligned five of the six components, because the higher aligned accuracies of mixtral:8x22b and gpt-4-turbo-preview ($0.778$) rest on three aligned components only. \par
In (\textit{ii}), baseline RAG improves the aligned accuracy over the baseline LLM of the same model in 23 of the 51 comparable configurations (most often with R2-R3, R1-R2, and R1-R2-R3 as references), whereas the raw accuracy improves in 14 of the 52. However, in several experiments, the reason for higher accuracy is the result of the deduction of unmatched parts during the matching and alignment. Hence, one should check both the spectrum distance (to understand how close the system is to the ground truth one) and the DSM prediction of the model (to see the reduction in the number of components after the alignment step) while checking the DSM matrix to see the real changes (removal of the parts during the alignment). \par
The validator generally confirmed the predictions of most models without requiring corrections. However, in experiments involving the deepseek-r1:14b model, which occasionally showed confusion in its reasoning for task (\textit{ii}), the validator \textit{did} provide corrections, effectively guiding the model's reasoning process. This would suggest the validator's utility might be model-specific, being more beneficial for models prone to reasoning inconsistencies. \par
Based on the baseline RAG results available at the time of the experiments and the cost of GraphRAG indexing, we evaluated only the R1-R2 and R2-R3 combinations for baseline GraphRAG.
\section{Conclusion and Future Work}
In this paper, we investigated the capabilities of LLMs with their RAG, and GraphRAG derivatives in generating DSMs with two distinct use cases: power screwdriver and CubeSat with well-known architectural references, while evaluating two architectural generation tasks: (i) determining relationships between predefined components and (ii) identifying components and their relationships. We assessed performance using both local metrics for cell-level evaluation and global metrics to capture overall architecture. Despite design and computational challenges, we identify opportunities for automated DSM generation, with all code publicly available for reproducibility, enabling domain experts and interested parties to validate results, provide feedback, and contribute domain-specific improvements.\par
Our experiments demonstrated that model architecture often proves more critical than parameter count for relationship classification tasks. For instance, for the power screwdriver's spatial reasoning in task (\textit{i}), mixtral:8x22b achieved the highest performance, despite being smaller than gpt-4 variants. For CubeSat's whole-part relationships, llama3.3:70B showed best performance. RAG improved performance in specific configurations, but combining all references did not consistently yield improvements, suggesting that careful reference selection is essential. The top-performing models also had the lowest edit distances to the ground truth in five of the six tables (in the aligned power screwdriver table the lowest edit distance belongs to deepseek-r1:14b with RAG R2-R3), whereas spectral distance did not always favor the same models, which suggests that cell-level accuracy and global structural similarity capture different aspects of architectural quality. While these results suggest architectural choices may outweigh parameter count in some scenarios, further studies across more diverse tasks and models are needed to fully substantiate this trend.\par
Irrespective of the method, all approaches require consideration of prompt design, as query formulation significantly affects outcomes. This dependency on query design presents considerations for broader application across diverse system types. The methodology was tested on two distinct cases (power screwdriver focusing on spatial proximity, CubeSat focusing on whole-part relationships), providing valuable insights but also highlighting the need for future validation. Extending this work to a broader range of CPS architectures, varying documentation quality, and diverse relationship types (e.g., functional, logical flows) would be crucial to assess the generalizability of these automated DSM generation techniques. The effectiveness of RAG approaches depends on reference document selection and classification, which involves a balance between automation and domain expertise. Computational requirements vary across approaches, with cloud-based models offering faster processing but at a considerably higher cost, while local implementations require substantial hardware resources. The preparation of the experimental setup benefits from domain knowledge to ensure appropriate system representation.\par
Several research directions could enhance automated DSM generation. Addressing prompt dependency is imperative for broader application, with potential approaches including compiler-like frameworks such as DSpy \cite{khattab2023dspy} that can automate the creation of self-improving pipelines for language model calls. Incorporating reasoning models shows promise for competitive architectural results even with smaller models, potentially providing cost-effective alternatives to larger proprietary options. Furthermore, exploring a wider array of LLM architectures, potentially including models fine-tuned specifically for engineering tasks or complex reasoning, could yield deeper insights into the factors driving performance variations beyond just parameter count or reference data. For better integration with detailed design processes, improved connections with Product Lifecycle Management interfaces are needed, particularly focusing on physical viewpoints (e.g., geometric data representations and material aspects). Finally, while LLMs, ALMs, and their derivatives demonstrate understanding and reasoning abilities, they face challenges in real-world task execution. To overcome these limitations, structured agentic frameworks offer significant potential for enhanced performance (e.g., error correction, etc.). Agentic approaches with reasoning and programmatic xLM mechanisms might generate more accurate system architectures with multi-level granularity, which we leave for future study.\par

\section*{acknowledgements}
The authors gratefully acknowledge Dr. Steve Simske and Dr. Emily King for their insightful feedback.

\FloatBarrier
\bibliography{sample}
\newpage
\appendix

\section{Appendix A}
\label{App:A}
Below are the LLM prompts as used in `Step2\_DSM\_LLM\_RAG\_GRAPHRAG.ipynb` The reader can find the output from these prompts in the same IPython Notebook and provided log files (inside `$\text{/logs}$` folder). The structure of the prompts as single sentence per line is purposefully written as presented here and in the code repository. However, due to the limited space on the paper, we cannot reflect that verbatim and present it in the associated prompt boxes below. \\
Moreover, we present the details of the model selection procedure as well as additional details of the experiments and their complete results as part of the corresponding tables. 
\subsection{}
\textit{Power Screwdriver Example Prompt for \textit{i}:}
\begin{promptbox}
"""
Please identify system-level proximity relationships (in contact) between the components of Power Screwdriver, as listed below, and represent them in a Design Structure Matrix (DSM) format.  \
The relationships should be articulated in plain English, capturing the essence of their interactions. \
Here is the list of subsystems/components to analyze: ["Bit", "Transmission", "Motor", "Electrical System", "Battery Holder", "Housing", "External Environment"] \
The DSM should clearly indicate the interactions between these components as 1 exist and 0 does not exist, considering both the presence and nature of these connections. \   
Present the output in a square matrix format (number of rows and columns MUST be the same as the number of components = 7 for this example), with rows and columns labeled accordingly. \
Diagonal elements should be 1 in DSM matrix since each component interacts with itself. \
If you don't know the answer, just write 2 for the specific cell that you are unsure of. A worst case scenario of the **as a list of lists** shown below (diagonal elements are 1 and rest are 2). Don't make up an answer. \\
    final response = [
        [1, 2, 2, 2, 2, 2, 2],
        [2, 1, 2, 2, 2, 2, 2],
        [2, 2, 1, 2, 2, 2, 2],
        [2, 2, 2, 1, 2, 2, 2],
        [2, 2, 2, 2, 1, 2, 2],
        [2, 2, 2, 2, 2, 1, 2],
        [2, 2, 2, 2, 2, 2, 1]
    ] \\
IMPORTANT: Your response must **ONLY** contain a valid **as a list of lists** in the following example format 7x7 matrix. \\
The below list of lists is just an example format of your response. **Replace** the values with your **actual analysis** while providing your response with parameter of "final response".\\
    final response = [
        [1, 0, 0, 0, 0, 0, 0],
        [0, 1, 0, 0, 0, 0, 0],
        [0, 0, 1, 0, 0, 0, 0],
        [0, 0, 0, 1, 0, 0, 0],
        [0, 0, 0, 0, 1, 0, 0],
        [0, 0, 0, 0, 0, 1, 0],
        [0, 0, 0, 0, 0, 0, 1]
    ] \\
IMPORTANT: THE FORMAT OF THE OUTPUT MUST BE AS LIST OF LISTS ONLY SIMILAR TO THE ONE ABOVE (NO LATEX, CODE, OR OTHER ENCODINGS).
"""
\end{promptbox}
\textit{Power Screwdriver Example Prompt for \textit{ii}:}
\begin{promptbox}
"""
Please identify system-level proximity relationships between the components of power screwdriver, and represent them in a Design Structure Matrix (DSM) format.  

The relationships should be articulated in plain English, capturing the essence of their interactions. 

The DSM should clearly indicate the interactions between these components as 1 exist and 0 does not exist, considering both the presence and nature of these connections. 
    
Present the output in a square matrix format (number of rows and columns MUST be the same as the number of identified components), with rows and columns labeled accordingly. 

Diagonal elements should be 1 in DSM matrix since each component interacts with itself. 

If you don't know the answer, just write 2 for the specific cell that you are unsure of. A worst case scenario of the **as a list of lists** shown below (diagonal elements are 1 and rest are 2). Don't make up an answer. 

This is an example of the output format for 7 components (as 7x7 matrix)-based on our system-of-interest the size of the matrix can change. 

{
  "final\_response": [
                [1, 2, 2, 2, 2, 2, 2], 
                [2, 1, 2, 2, 2, 2, 2], 
                [2, 2, 1, 2, 2, 2, 2],
                [2, 2, 2, 1, 2, 2, 2],
                [2, 2, 2, 2, 1, 2, 2],
                [2, 2, 2, 2, 2, 1, 2],
                [2, 2, 2, 2, 2, 2, 1]]\\
}

IMPORTANT: Your response must **ONLY** contain a valid **as a list of lists** as JSON in the following example format 7x7 matrix. 

The below list of lists is just an example format of your response. **Replace** the values with your **actual analysis** while providing your response with parameter of "final response".

This is an example of the output format for 7 components (as 7x7 matrix)-based on our system-of-interest the size of the matrix can change. 

IMPORTANT: Your response MUST be a valid JSON object in the following format:
{
  "final\_response": [[1,0,0,0,0,0,0], [0,1,0,0,0,0,0], ...]
}
"""
\end{promptbox}
\textit{CubeSat Example Prompt 1 for \textit{i}:}
\begin{promptbox}
\_prompt = f"""
    Please identify system-level {config.relationship\_type} relationships between the components of {config.concept\_name}, as listed below, and represent them in a Design Structure Matrix (DSM) format.  \
    
    The relationships should be articulated in plain English, capturing the essence of their interactions. \
    
    Here is the list of subsystems/components to analyze: {config.predicted\_components} \
    
    The DSM should clearly indicate the interactions between these components as 1 exist and 0 does not exist, considering both the presence and nature of these connections. \
    
    Present the output in a square matrix format (number of rows and columns MUST be the same as the number of components = {len(config.predicted\_components)} for this example), with rows and columns labeled accordingly. \
    
    Diagonal elements should be 1 in DSM matrix since each component interacts with itself. \
    
    If you don't know the answer, just write 2 for the specific cell that you are unsure of. A worst case scenario of the **as a list of lists** shown below (diagonal elements are 1 and unsure ones are 2). Don't make up an answer. \
    
    final response = [[1, 2, 2, 2, 2, 2, 2],[2, 1, 2, 2, 2, 2, 2],[2, 2, 1, 2, 2, 2, 2],[2, 2, 2, 1, 2, 2, 2],[2, 2, 2, 2, 1, 2, 2],[2, 2, 2, 2, 2, 1, 2],[2, 2, 2, 2, 2, 2, 1]] for 7 components \
    
    final response = [[1, 2, 2],[2, 1, 2],[2, 2, 1]] for 3 components \
    
    IMPORTANT: Your response must **ONLY** contain a valid **as a list of lists** in the following example format 7x7 matrix. \
    
    The below list of lists is just an example format of a response. **Replace** the values with your **actual analysis** while providing your response with parameter of "final response".\
    
    final response = [[1, 0, 0, 0, 0, 0, 0],[0, 1, 0, 0, 0, 0, 0],[0, 0, 1, 0, 0, 0, 0],   [0, 0, 0, 1, 0, 0, 0],[0, 0, 0, 0, 1, 0, 0],[0, 0, 0, 0, 0, 1, 0],[0, 0, 0, 0, 0, 0, 1]] for 7 components \
    
    final response = [[1, 0, 0],[0, 1, 0],[0, 0, 1]] for 3 components \
     
    IMPORTANT: THE FORMAT OF THE OUTPUT MUST BE AS LIST OF LISTS ONLY SIMILAR TO THE ONE ABOVE (NO LATEX, CODE, OR OTHER ENCODINGS).
    """
\end{promptbox}
Above, `config.predicted\_components` is not empty, unlike in the (\textit{ii}) provided below. This way, we were able to generalize and parameterize for both cases. In (\textit{ii}), the identification prompt determines the predicted components, and hence, it is utilized accordingly in the following steps. \\
\textit{Identification prompt \textit{ii}:}
\begin{promptbox}
    prompt = f"""
            Please identify the major components of {self.config.concept\_name} in the {self.config.application\_domain} based on {self.config.relationship\_type} relationships. \
            
            The number of components must be {len(self.original\_components)} and must be a list of strings. \
            
            Please do not identify similar components that will work together under subsystem level since our interest is to identify the major components that form the core architecture of the system. \
            
            Here is an example of the components for made-up system: \
            
            final response = [ "component 1", "component 2", "component 3", "component 4", "component 5" ] \
            
            IMPORTANT: THE FORMAT OF THE OUTPUT MUST BE AS A LIST OF STRINGS ONLY SIMILAR TO THE ONE ABOVE (NO LATEX, NO CODE, NO DESCRIPTION, OR OTHER ENCODINGS).  \
            """
\end{promptbox}
\textit{CubeSat Example Update Prompt 1 for \textit{ii}:}
\begin{promptbox}
    \_prompt = f"""
    Please identify system-level {config.relationship\_type} relationships between the components of {config.concept\_name}, as listed below, and represent them in a Design Structure Matrix (DSM) format.  \
    
    The relationships should be articulated in plain English, capturing the essence of their interactions. \
    
    Here is the list of subsystems/components to analyze: {config.predicted\_components} \
    
    The DSM should clearly indicate the interactions between these components as 1 exist and 0 does not exist, considering both the presence and nature of these connections. \
    
    Present the output in a square matrix format (number of rows and columns MUST be the same as the number of components = {len(config.predicted\_components)} for this example), with rows and columns labeled accordingly. \
    
    Diagonal elements should be 1 in DSM matrix since each component interacts with itself. \
    
    If you don't know the answer, just write 2 for the specific cell that you are unsure of. A worst case scenario of the **as a list of lists** shown below (diagonal elements are 1 and unsure ones are 2). Don't make up an answer. \
    
    final response = [[1, 2, 2, 2, 2, 2, 2],[2, 1, 2, 2, 2, 2, 2],[2, 2, 1, 2, 2, 2, 2],[2, 2, 2, 1, 2, 2, 2],[2, 2, 2, 2, 1, 2, 2],[2, 2, 2, 2, 2, 1, 2],[2, 2, 2, 2, 2, 2, 1]] for 7 components \
    
    final response = [[1, 2, 2],[2, 1, 2],[2, 2, 1]] for 3 components \
    
    IMPORTANT: Your response must **ONLY** contain a valid **as a list of lists** in the following example format 7x7 matrix (NO LATEX, NO CODE, NO DESCRIPTION, OR OTHER ENCODINGS, JUST THE LIST OF LISTS). \
    
    The below list of lists is just examples format of a response. **Replace** the values with your **actual analysis** while providing your response with parameter of "final response".\
    
    final response = [[1, 0, 0, 0, 0, 0, 0],[0, 1, 0, 0, 0, 0, 0],[0, 0, 1, 0, 0, 0, 0],   [0, 0, 0, 1, 0, 0, 0],[0, 0, 0, 0, 1, 0, 0],[0, 0, 0, 0, 0, 1, 0],[0, 0, 0, 0, 0, 0, 1]] for 7 components \
    
    final response = [[1, 0, 0],[0, 1, 0],[0, 0, 1]] for 3 components \
     
    IMPORTANT: THE FORMAT OF THE OUTPUT MUST BE AS LIST OF LISTS ONLY SIMILAR TO THE ONE ABOVE (NO LATEX, NO CODE, NO DESCRIPTION, OR OTHER ENCODINGS, JUST THE LIST OF LISTS). \
    """
\end{promptbox}
\textit{Validator prompt:}
\begin{promptbox}
    validation\_prompt = f"""
        You are a validator. Review this response based on the requirements of the original prompt: \
        
        Original prompt: {original\_prompt}
        Response: {raw\_response}
        
        Requirements:
        1- Response must follow the format in the original prompt.
        2- Response must have the same number of components as the original prompt.
        
        Answer only with "Valid" or "Invalid".
        """
\end{promptbox}
\subsection{Selection of the Models and Compute for the Experiments}
\label{App:selection}
\begin{table*}[t!]
\centering
\caption{Model and Experimental Details (as more details provided in GitHub Repository)}
\begin{tabular}{|l|l|l|l|l|l|l|}
\hline
\multicolumn{1}{|c|}{\textbf{Model}} & \multicolumn{1}{c|}{\textbf{Embedding}} & \multicolumn{1}{c|}{\textbf{Compute}} & \multicolumn{1}{c|}{\textbf{CodeBase}} & \multicolumn{1}{c|}{\textbf{Cost|Duration}} & \multicolumn{1}{c|}{\textbf{Input}} & \multicolumn{1}{c|}{\textbf{Quality(\# of Nodes)}} \\
\hline
gpt4tp-OAI & TE3s-OAI & OAI Infra & MS-GraphRAG & 4.87 USD & CC - Dickens & High(208nodes)\\
\hline
gpt4o-OAI & TE3s-OAI & OAI Infra & MS-GraphRAG & 1.18 USD & CC - Dickens & High(267nodes)\\
\hline
mistral & TE3s-OAI & HP Victus & MS-GraphRAG & 33 mins 2secs & CC - Dickens & Low(460 nodes)\\
\hline
deepseek-llm & TE3s-OAI & HP Victus & MS-GraphRAG & 1hr 30mins & CC - Dickens & Low(17 nodes)\\
\hline
llama3.2:3b & TE3s-OAI & HP Victus & MS-GraphRAG & 24mins 6secs & CC - Dickens & Low(59 nodes)\\
\hline
mixtral-8x22b & NET-Ollama & Tensorcraft & MS-GraphRAG & 4hrs 11mins & CC - Dickens & High(587nodes)\\
\hline
llama3.3:70b & NET-Ollama & Tensorcraft & MS-GraphRAG & 2hrs 48mins & CC - Dickens & High(600nodes)\\
\hline
ds-r1:14b & TE3s-OAI & Tensorcraft & MS-GraphRAG & 1hr 18mins & CC - Dickens & High(499nodes)\\
\hline
\multicolumn{7}{|l|}{\footnotesize\textbf{TE3}:Text-embedding-small, \textbf{NET}:Nomic-Embed-Text, \textbf{OAI}:OpenAI, \textbf{CC}:Christmas Carol \cite{dickens2003christmas}, \textbf{MS}:Microsoft, \textbf{DS}:DeepSeek} \\
\hline
\end{tabular}
\end{table*}
For initial benchmarking, we tested relatively small and open-source models (e.g., mistral:latest-4k, llama3.2:1b/3b) on an entry-level gaming laptop (Gigabyte G6) equipped with an Intel i7-13620HX CPU, 64GB RAM, and a power-limited NVIDIA GeForce RTX 4060 mobile GPU (8GB vRAM) and more moderate (HP Victus) with an Intel i7-13700HX, 64GB RAM, and a NVidia GeForce RTX 4060 mobile GPU (8GB vRAM). While this setup allowed for rapid iteration, it became clear that the limited vRAM and smaller context windows of the models that can fit into this vRAM were insufficient, especially during GraphRAG indexing. \par
For more demanding open-source models (e.g., mixtral:8x22B, llama3.3:70b, etc.), we leveraged our custom multi-GPU workstation (Craftnetics TensorCraft-Red) featuring an AMD EPYC CPU, 504GB RAM, and six AMD RDNA3 GPUs with a combined 144GB vRAM. This setup enabled better performance when handling larger architectural graphs and multi-level retrieval processes. However, compared to proprietary models with cloud-infrastructure, the processing time remained a concern, particularly in GraphRAG implementations requiring graph indexing and retrieval augmentation. \par
For proprietary models, we relied on OpenAI's cloud infrastructure through API access. The lack of transparency in hardware specifications was offset by significantly faster inference times and a more optimized tokenization process, particularly for RAG and GraphRAG tasks. Despite a non-negligible cost per request, OpenAI models consistently provided high-quality node extractions during the indexing step of the GraphRAG in a short notice. \par
During preliminary tests, we also evaluated embedding models to balance local and cloud-based computations. Specifically, we utilized OpenAI's text-embedding-3-small (te3s) against Nomic's nomic-embed-text (net) for vector storage and retrieval using ChromaDB. The combination of local LLMs with cloud-based embeddings helped offload GPU workload while improving retrieval consistency for resource-constrained hardware. Furthermore, the per-token costs (0.02 cents per million tokens for `text-embedding-3-small`) would be a limit for larger datasets. \par
To benchmark indexing and clustering performance, we tested a baseline dataset consisting of Charles Dickens's A Christmas Carol \cite{dickens2003christmas}. This text provides a controlled environment to compare model performance, retrieval efficiency, and indexing overhead. From these trials, we determined that smaller models lacked the necessary performance for GraphRAG-based knowledge graph extraction. This led us to exclude them from further analysis for the whole set of experiments with other methods. Instead, we focused on a refined selection of models that performed best in these tests, including mixtral-8x22b, llama-3.3:70b, deepseek-r1:14b, gpt-4-o, and gpt-4-turbo-preview, for the remainder of the study. \par
Additionally, we identified scalability constraints in deepseek-r1:32b, which, despite its promising architecture, suffered from long response times and server errors (500 timeout issues) in GraphRAG indexing tasks locally. As a result, we excluded deepseek-r1:32b from further experiments, prioritizing models that maintained a balance between performance and usability. Our thinking by selecting 14b over 32b is having a comparison between a relatively larger model without reasoning and showing the performance of a smaller model that can run on relatively resource-constrained hardware with reasoning. \par
\subsection{Some Remarks for the Experiments}
\label{App:remarks}
Here, during our experiments---especially for local LLMs,---one needs to pay close attention to the context window size and embedding length of the model that is used. The default context window in Ollama is purposefully small (2048) and needs to be changed, as we detailed in our repository. Otherwise, the results of the GraphRAG will be quite poor for the open-source models, and in some cases, a lot of errors occur. In addition to this, to have a proper entity detection from the text, the user has to pay attention to some of the parameters, especially `Number\_of\_gleanings` and `chunk\_size`. A small `chunk\_size` will identify the entities poorly since the LLM will only process within the provided `chunk\_size` and irrespective of the chunk size, as the number of gleanings (passes) increases the number of identified entities \cite{edge2024local}. \par
In the light of these foregoing points, we conduct the benchmark for the some variation of RAG experiments with appropriate setting of the parameters (chunk\_size=1200 and chunk\_overlap=300 for the RAG text splitter; chunk size 1200 with overlap 100 for gpt-4-turbo-preview and 150 for the Ollama-served models, and number of gleanings 1, for GraphRAG in `settings.yaml`; num\_ctx={maximum embedding size} for the corresponding Ollama models). Because of the required time and cost for the indexing step, we limited the GraphRAG experiments to two reference combinations chosen from the baseline RAG results available at the time of the experiments. Hence, we utilized R1-R2 and R2-R3 cases for GraphRAG in our evaluations. \par
We observed unsure outputs primarily in GPT-4o, and in some configurations of deepseek-r1:14b, llama3.3:70b, and mixtral:8x22b; gpt-4-turbo-preview did not produce unsure cells. GPT-4o showed 16.7\% unsure cells (30/180) in task \textit{i} CubeSat baseline LLM, 83.3\% (150/180 per configuration) across all task \textit{i} CubeSat RAG configurations, and 85.7\% (210/245 per configuration) across all power screwdriver RAG configurations in both tasks \textit{i} and \textit{ii}. In task \textit{ii}, GPT-4o's initial power screwdriver baseline experiment showed 73.5\% (180/245) and its initial CubeSat RAG experiment showed 83.3\% in five of the seven reference configurations; the repeated experiments produced no unsure cells. The GPT-4o task \textit{ii} power screwdriver baseline metrics reported in the tables (Tables~\ref{tab:case_ii_raw_ps_experiments} and~\ref{tab:case_ii_aligned_ps_experiments}) correspond to the repeated experiment, while all GPT-4o RAG configurations are excluded (na) due to these inconsistencies across the experiment set. Deepseek-r1:14b showed unsure percentages ranging from 1.7\% to 100\% in select configurations: task \textit{i} CubeSat R3 (36.7\%), R2-R3 (80\%); task \textit{i} power screwdriver R1/R2/R1-R2/R1-R2-R3 (20\%), R2-R3 (40\%), R3 (60\%), R1-R3 (80\%); task \textit{ii} CubeSat R1/R1-R3 (20\%), R3 (60\%), R1-R2-R3 (75\%), GraphRAG R1-R2/R2-R3 (1.7\%); task \textit{ii} power screwdriver R2 (40\%), R1-R3 (2.4\%), GraphRAG R2-R3 (100\%). Llama3.3:70b showed 85.7\% unsure in task \textit{ii} power screwdriver R3, and mixtral:8x22b showed 12.8\% (23/180) in task \textit{ii} CubeSat R1-R3. Unsure cells are never scored as confident predictions: for configurations with low unsure percentages, the affected cells are excluded from the cell-level metric denominators; for configurations dominated by unsure outputs, we exclude the entire runs from the reported results and mark them as not available (na), since the resulting metrics would not reflect confident predictions. One aligned cell (CubeSat, mixtral:8x22b with RAG R3) is also marked na because the alignment matched a single one of the six components, so its metrics would describe a $1 \times 1$ matrix. The full per-model, per-configuration unsure counts and percentages for both tasks and both use cases are provided in the accompanying repository~\cite{bankh2024xlm}. \par
For the baseline LLM approach, we evaluated both proprietary and open-source models. The proprietary models included gpt-4-turbo-preview and gpt-4-o, while the open-source models encompass llama3.3:70B, mixtral:8x22B, and deepSeek-R1:14B. \par
In the following tables, we provide complementary information to Fig. \ref{fig:8} and Fig \ref{fig:9}, where we present the best result out of other possible combinations of the references for \textit{i} and \textit{ii} tasks.

\setlength{\textfloatsep}{1.5\baselineskip}
\setlength{\intextsep}{\baselineskip}
\setlength{\dbltextfloatsep}{1.5\baselineskip}
\setlength{\dblfloatsep}{\baselineskip}
\setlength{\abovecaptionskip}{0pt plus 0pt minus 0pt}
\setlength{\belowcaptionskip}{0pt plus 0pt minus 0pt}
\captionsetup*[table]{skip=0pt plus 0pt minus 0pt}

\clearpage
\begin{table*} %
\captionsetup{width=\textwidth}
\scalebox{0.98}{\resizebox{\textwidth}{!}{%
\fontsize{5}{8}\selectfont
\begin{tabular}{|c|c|c|c|c|c|} %
\hline
 & \multicolumn{1}{c|}{\textbf{1.77T/128k/128k}} & \multicolumn{1}{c|}{\textbf{1.77T/128k/128k}} & \multicolumn{1}{c|}{\textbf{140.6B/65k/6k}} & \multicolumn{1}{c|}{\textbf{70.6B/131k/8k}} & \multicolumn{1}{c|}{\textbf{14.8B/131k/5k}} \\
\hline
 & \multicolumn{1}{c|}{\textbf{gpt4-turbo-preview}} & \multicolumn{1}{c|}{\textbf{gpt4o}} & \multicolumn{1}{c|}{\textbf{mixtral:8x22b}} & \multicolumn{1}{c|}{\textbf{llama3.3:70b}} & \multicolumn{1}{c|}{\textbf{deepseek-r1:14b}} \\
\hline
\multirow{5}{*}{\rotatebox[origin=c]{90}{\textbf{LLM}}} & 
\multirow{5}{*}{\rotatebox[origin=c]{90}{\makecell[c]{0.7821 $\pm$ 0.0256 \\ 1.0000 $\pm$ 0.0000 \\ 0.8775 $\pm$ 0.0158 \\ \textcolor{silver}{\textbf{0.8444 $\pm$ 0.0222}} \\ 5.6000 $\pm$ 0.8000 \\ 1.5163 $\pm$ 0.1270}}} &
\multirow{5}{*}{\rotatebox[origin=c]{90}{\makecell[c]{0.6838 $\pm$ 0.1047 \\ 1.0000 $\pm$ 0.0000 \\ 0.8075 $\pm$ 0.0761 \\ 0.6838 $\pm$ 0.1047 \\ 22.0000 $\pm$ 4.8990 \\ 3.9179 $\pm$ 0.5866}}} &
\multirow{5}{*}{\rotatebox[origin=c]{90}{\makecell[c]{0.7692 $\pm$ 0.0000 \\ 1.0000 $\pm$ 0.0000 \\ 0.8696 $\pm$ 0.0000 \\ 0.8333 $\pm$ 0.0000 \\ 6.0000 $\pm$ 0.0000 \\ 1.3409 $\pm$ 0.0000}}} &
\multirow{5}{*}{\rotatebox[origin=c]{90}{\makecell[c]{0.9091 $\pm$ 0.0000 \\ 1.0000 $\pm$ 0.0000 \\ 0.9524 $\pm$ 0.0000 \\ \textcolor{gold}{\textbf{0.9444 $\pm$ 0.0000}} \\ 2.0000 $\pm$ 0.0000 \\ 1.2042 $\pm$ 0.0000}}} &
\multirow{5}{*}{\rotatebox[origin=c]{90}{\makecell[c]{0.6957 $\pm$ 0.0121 \\ 0.9800 $\pm$ 0.0400 \\ 0.8131 $\pm$ 0.0065 \\ 0.7500 $\pm$ 0.0000 \\ 9.0000 $\pm$ 0.0000 \\ 1.5899 $\pm$ 0.1701}}} \\
&&&&&\\
&&&&&\\
&&&&&\\
&&&&&\\
\hline
\multirow{5}{*}{\rotatebox[origin=c]{90}{\textbf{RAG R1}}} & 
\multirow{5}{*}{\rotatebox[origin=c]{90}{\makecell[c]{0.7253 $\pm$ 0.0220 \\ 1.0000 $\pm$ 0.0000 \\ 0.8406 $\pm$ 0.0145 \\ 0.7889 $\pm$ 0.0222 \\ 7.6000 $\pm$ 0.8000 \\ 1.5205 $\pm$ 0.0339}}} &
\multirow{5}{*}{\rotatebox[origin=c]{90}{\makecell[c]{na \\ na \\ na \\ na \\ na \\ na}}} &
\multirow{5}{*}{\rotatebox[origin=c]{90}{\makecell[c]{0.5882 $\pm$ 0.0000 \\ 1.0000 $\pm$ 0.0000 \\ 0.7407 $\pm$ 0.0000 \\ 0.6111 $\pm$ 0.0000 \\ 14.0000 $\pm$ 0.0000 \\ 2.5285 $\pm$ 0.1371}}} &
\multirow{5}{*}{\rotatebox[origin=c]{90}{\makecell[c]{0.9091 $\pm$ 0.0000 \\ 1.0000 $\pm$ 0.0000 \\ 0.9524 $\pm$ 0.0000 \\ \textcolor{gold}{\textbf{0.9444 $\pm$ 0.0000}} \\ 2.0000 $\pm$ 0.0000 \\ 1.2042 $\pm$ 0.0000}}} &
\multirow{5}{*}{\rotatebox[origin=c]{90}{\makecell[c]{0.6111 $\pm$ 0.0278 \\ 1.0000 $\pm$ 0.0000 \\ 0.7582 $\pm$ 0.0220 \\ 0.6444 $\pm$ 0.0444 \\ 12.8000 $\pm$ 1.6000 \\ 2.3900 $\pm$ 0.4047}}} \\
&&&&&\\
&&&&&\\
&&&&&\\
&&&&&\\
\hline
\multirow{5}{*}{\rotatebox[origin=c]{90}{\textbf{RAG R2}}} & 
\multirow{5}{*}{\rotatebox[origin=c]{90}{\makecell[c]{0.7601 $\pm$ 0.0441 \\ 1.0000 $\pm$ 0.0000 \\ 0.8630 $\pm$ 0.0282 \\ 0.8222 $\pm$ 0.0416 \\ 6.4000 $\pm$ 1.4967 \\ 1.3483 $\pm$ 0.2960}}} &
\multirow{5}{*}{\rotatebox[origin=c]{90}{\makecell[c]{na \\ na \\ na \\ na \\ na \\ na}}} &
\multirow{5}{*}{\rotatebox[origin=c]{90}{\makecell[c]{0.7253 $\pm$ 0.0220 \\ 1.0000 $\pm$ 0.0000 \\ 0.8406 $\pm$ 0.0145 \\ 0.7889 $\pm$ 0.0222 \\ 7.6000 $\pm$ 0.8000 \\ 1.4006 $\pm$ 0.0169}}} &
\multirow{5}{*}{\rotatebox[origin=c]{90}{\makecell[c]{0.8333 $\pm$ 0.0000 \\ 1.0000 $\pm$ 0.0000 \\ 0.9091 $\pm$ 0.0000 \\ 0.8889 $\pm$ 0.0000 \\ 4.0000 $\pm$ 0.0000 \\ 1.2132 $\pm$ 0.0000}}} &
\multirow{5}{*}{\rotatebox[origin=c]{90}{\makecell[c]{0.7147 $\pm$ 0.0248 \\ 0.9500 $\pm$ 0.0316 \\ 0.8155 $\pm$ 0.0249 \\ 0.7611 $\pm$ 0.0333 \\ 8.6000 $\pm$ 1.2000 \\ 1.0628 $\pm$ 0.3183}}} \\
&&&&&\\
&&&&&\\
&&&&&\\
&&&&&\\
\hline
\multirow{5}{*}{\rotatebox[origin=c]{90}{\textbf{RAG R3}}} & 
\multirow{5}{*}{\rotatebox[origin=c]{90}{\makecell[c]{0.7692 $\pm$ 0.0000 \\ 1.0000 $\pm$ 0.0000 \\ 0.8696 $\pm$ 0.0000 \\ 0.8333 $\pm$ 0.0000 \\ 6.0000 $\pm$ 0.0000 \\ 1.4528 $\pm$ 0.0000}}} &
\multirow{5}{*}{\rotatebox[origin=c]{90}{\makecell[c]{na \\ na \\ na \\ na \\ na \\ na}}} &
\multirow{5}{*}{\rotatebox[origin=c]{90}{\makecell[c]{0.6667 $\pm$ 0.0833 \\ 1.0000 $\pm$ 0.0000 \\ 0.7972 $\pm$ 0.0559 \\ 0.7111 $\pm$ 0.0889 \\ 10.4000 $\pm$ 3.2000 \\ 1.9897 $\pm$ 0.3958}}} &
\multirow{5}{*}{\rotatebox[origin=c]{90}{\makecell[c]{0.9091 $\pm$ 0.0000 \\ 1.0000 $\pm$ 0.0000 \\ 0.9524 $\pm$ 0.0000 \\ \textcolor{gold}{\textbf{0.9444 $\pm$ 0.0000}} \\ 2.0000 $\pm$ 0.0000 \\ 1.2042 $\pm$ 0.0000}}} &
\multirow{5}{*}{\rotatebox[origin=c]{90}{\makecell[c]{0.6786 $\pm$ 0.3543 \\ 0.7500 $\pm$ 0.3768 \\ 0.7096 $\pm$ 0.3609 \\ 0.6889 $\pm$ 0.3555 \\ 23.6000 $\pm$ 20.9055 \\ 3.7412 $\pm$ 3.7821}}} \\
&&&&&\\
&&&&&\\
&&&&&\\
&&&&&\\
\hline
\multirow{5}{*}{\rotatebox[origin=c]{90}{\textbf{RAG R1-R2}}} & 
\multirow{5}{*}{\rotatebox[origin=c]{90}{\makecell[c]{0.6896 $\pm$ 0.0564 \\ 1.0000 $\pm$ 0.0000 \\ 0.8149 $\pm$ 0.0396 \\ 0.7444 $\pm$ 0.0667 \\ 9.2000 $\pm$ 2.4000 \\ 1.7806 $\pm$ 0.3338}}} &
\multirow{5}{*}{\rotatebox[origin=c]{90}{\makecell[c]{na \\ na \\ na \\ na \\ na \\ na}}} &
\multirow{5}{*}{\rotatebox[origin=c]{90}{\makecell[c]{0.6891 $\pm$ 0.0504 \\ 1.0000 $\pm$ 0.0000 \\ 0.8148 $\pm$ 0.0370 \\ 0.7444 $\pm$ 0.0667 \\ 9.2000 $\pm$ 2.4000 \\ 1.6758 $\pm$ 0.3921}}} &
\multirow{5}{*}{\rotatebox[origin=c]{90}{\makecell[c]{0.9091 $\pm$ 0.0000 \\ 1.0000 $\pm$ 0.0000 \\ 0.9524 $\pm$ 0.0000 \\ \textcolor{gold}{\textbf{0.9444 $\pm$ 0.0000}} \\ 2.0000 $\pm$ 0.0000 \\ 1.2042 $\pm$ 0.0000}}} &
\multirow{5}{*}{\rotatebox[origin=c]{90}{\makecell[c]{0.6039 $\pm$ 0.0314 \\ 1.0000 $\pm$ 0.0000 \\ 0.7526 $\pm$ 0.0237 \\ 0.6333 $\pm$ 0.0444 \\ 13.2000 $\pm$ 1.6000 \\ 2.4064 $\pm$ 0.1071}}} \\
&&&&&\\
&&&&&\\
&&&&&\\
&&&&&\\
\hline
\multirow{5}{*}{\rotatebox[origin=c]{90}{\textbf{RAG R1-R3}}} & 
\multirow{5}{*}{\rotatebox[origin=c]{90}{\makecell[c]{0.7363 $\pm$ 0.0269 \\ 1.0000 $\pm$ 0.0000 \\ 0.8478 $\pm$ 0.0177 \\ 0.8000 $\pm$ 0.0272 \\ 7.2000 $\pm$ 0.9798 \\ 1.5035 $\pm$ 0.0415}}} &
\multirow{5}{*}{\rotatebox[origin=c]{90}{\makecell[c]{na \\ na \\ na \\ na \\ na \\ na}}} &
\multirow{5}{*}{\rotatebox[origin=c]{90}{\makecell[c]{0.7143 $\pm$ 0.0000 \\ 1.0000 $\pm$ 0.0000 \\ 0.8333 $\pm$ 0.0000 \\ 0.7778 $\pm$ 0.0000 \\ 8.0000 $\pm$ 0.0000 \\ 1.4797 $\pm$ 0.0000}}} &
\multirow{5}{*}{\rotatebox[origin=c]{90}{\makecell[c]{0.9091 $\pm$ 0.0000 \\ 1.0000 $\pm$ 0.0000 \\ 0.9524 $\pm$ 0.0000 \\ \textcolor{gold}{\textbf{0.9444 $\pm$ 0.0000}} \\ 2.0000 $\pm$ 0.0000 \\ 1.2042 $\pm$ 0.0000}}} &
\multirow{5}{*}{\rotatebox[origin=c]{90}{\makecell[c]{0.5882 $\pm$ 0.0000 \\ 1.0000 $\pm$ 0.0000 \\ 0.7407 $\pm$ 0.0000 \\ 0.6111 $\pm$ 0.0000 \\ 14.0000 $\pm$ 0.0000 \\ 2.4599 $\pm$ 0.0000}}} \\
&&&&&\\
&&&&&\\
&&&&&\\
&&&&&\\
\hline
\multirow{5}{*}{\rotatebox[origin=c]{90}{\textbf{RAG R2-R3}}} & 
\multirow{5}{*}{\rotatebox[origin=c]{90}{\makecell[c]{0.7473 $\pm$ 0.0269 \\ 1.0000 $\pm$ 0.0000 \\ 0.8551 $\pm$ 0.0177 \\ 0.8111 $\pm$ 0.0272 \\ 6.8000 $\pm$ 0.9798 \\ 1.4866 $\pm$ 0.0415}}} &
\multirow{5}{*}{\rotatebox[origin=c]{90}{\makecell[c]{na \\ na \\ na \\ na \\ na \\ na}}} &
\multirow{5}{*}{\rotatebox[origin=c]{90}{\makecell[c]{0.7253 $\pm$ 0.0220 \\ 1.0000 $\pm$ 0.0000 \\ 0.8406 $\pm$ 0.0145 \\ 0.7889 $\pm$ 0.0222 \\ 7.6000 $\pm$ 0.8000 \\ 1.4006 $\pm$ 0.0169}}} &
\multirow{5}{*}{\rotatebox[origin=c]{90}{\makecell[c]{0.8333 $\pm$ 0.0000 \\ 1.0000 $\pm$ 0.0000 \\ 0.9091 $\pm$ 0.0000 \\ 0.8889 $\pm$ 0.0000 \\ 4.0000 $\pm$ 0.0000 \\ 1.2132 $\pm$ 0.0000}}} &
\multirow{5}{*}{\rotatebox[origin=c]{90}{\makecell[c]{na \\ na \\ na \\ na \\ na \\ na}}} \\
&&&&&\\
&&&&&\\
&&&&&\\
&&&&&\\
\hline
\multirow{5}{*}{\rotatebox[origin=c]{90}{\textbf{RAG R1-R2-R3}}} & 
\multirow{5}{*}{\rotatebox[origin=c]{90}{\makecell[c]{0.6964 $\pm$ 0.0357 \\ 1.0000 $\pm$ 0.0000 \\ 0.8205 $\pm$ 0.0256 \\ 0.7556 $\pm$ 0.0444 \\ 8.8000 $\pm$ 1.6000 \\ 1.6674 $\pm$ 0.2601}}} &
\multirow{5}{*}{\rotatebox[origin=c]{90}{\makecell[c]{na \\ na \\ na \\ na \\ na \\ na}}} &
\multirow{5}{*}{\rotatebox[origin=c]{90}{\makecell[c]{0.7253 $\pm$ 0.0220 \\ 1.0000 $\pm$ 0.0000 \\ 0.8406 $\pm$ 0.0145 \\ 0.7889 $\pm$ 0.0222 \\ 7.6000 $\pm$ 0.8000 \\ 1.4520 $\pm$ 0.0555}}} &
\multirow{5}{*}{\rotatebox[origin=c]{90}{\makecell[c]{0.9091 $\pm$ 0.0000 \\ 1.0000 $\pm$ 0.0000 \\ 0.9524 $\pm$ 0.0000 \\ \textcolor{gold}{\textbf{0.9444 $\pm$ 0.0000}} \\ 2.0000 $\pm$ 0.0000 \\ 1.2042 $\pm$ 0.0000}}} &
\multirow{5}{*}{\rotatebox[origin=c]{90}{\makecell[c]{0.5956 $\pm$ 0.0147 \\ 1.0000 $\pm$ 0.0000 \\ 0.7464 $\pm$ 0.0114 \\ 0.6222 $\pm$ 0.0222 \\ 13.6000 $\pm$ 0.8000 \\ 2.4487 $\pm$ 0.0224}}} \\
&&&&&\\
&&&&&\\
&&&&&\\
&&&&&\\
\hline
\multirow{5}{*}{\rotatebox[origin=c]{90}{\textbf{\makecell[c]{Graph RAG \\ R1-R2}}}} & 
\multirow{5}{*}{\rotatebox[origin=c]{90}{\makecell[c]{0.8434 $\pm$ 0.0436 \\ 0.7600 $\pm$ 0.0735 \\ 0.7988 $\pm$ 0.0567 \\ 0.7889 $\pm$ 0.0572 \\ 7.6000 $\pm$ 2.0591 \\ 1.6230 $\pm$ 0.2204}}} &
\multirow{5}{*}{\rotatebox[origin=c]{90}{\makecell[c]{na \\ na \\ na \\ na \\ na \\ na}}} &
\multirow{5}{*}{\rotatebox[origin=c]{90}{\makecell[c]{0.7143 $\pm$ 0.0000 \\ 1.0000 $\pm$ 0.0000 \\ 0.8333 $\pm$ 0.0000 \\ 0.7778 $\pm$ 0.0000 \\ 8.0000 $\pm$ 0.0000 \\ 1.4797 $\pm$ 0.0000}}} &
\multirow{5}{*}{\rotatebox[origin=c]{90}{\makecell[c]{0.7143 $\pm$ 0.0000 \\ 1.0000 $\pm$ 0.0000 \\ 0.8333 $\pm$ 0.0000 \\ 0.7778 $\pm$ 0.0000 \\ 8.0000 $\pm$ 0.0000 \\ 1.4797 $\pm$ 0.0000}}} &
\multirow{5}{*}{\rotatebox[origin=c]{90}{\makecell[c]{0.6667 $\pm$ 0.0000 \\ 1.0000 $\pm$ 0.0000 \\ 0.8000 $\pm$ 0.0000 \\ 0.7222 $\pm$ 0.0000 \\ 10.0000 $\pm$ 0.0000 \\ 2.0269 $\pm$ 0.0000}}} \\
&&&&&\\
&&&&&\\
&&&&&\\
&&&&&\\
\hline
\multirow{5}{*}{\rotatebox[origin=c]{90}{\textbf{\makecell[c]{Graph RAG \\ R2-R3}}}} & 
\multirow{5}{*}{\rotatebox[origin=c]{90}{\makecell[c]{0.9169 $\pm$ 0.0467 \\ 0.7500 $\pm$ 0.0000 \\ 0.8246 $\pm$ 0.0185 \\ 0.8222 $\pm$ 0.0222 \\ 6.4000 $\pm$ 0.8000 \\ 1.6396 $\pm$ 0.2673}}} &
\multirow{5}{*}{\rotatebox[origin=c]{90}{\makecell[c]{na \\ na \\ na \\ na \\ na \\ na}}} &
\multirow{5}{*}{\rotatebox[origin=c]{90}{\makecell[c]{0.7099 $\pm$ 0.0088 \\ 0.9800 $\pm$ 0.0400 \\ 0.8232 $\pm$ 0.0203 \\ 0.7667 $\pm$ 0.0222 \\ 8.4000 $\pm$ 0.8000 \\ 1.4743 $\pm$ 0.0108}}} &
\multirow{5}{*}{\rotatebox[origin=c]{90}{\makecell[c]{0.7143 $\pm$ 0.0000 \\ 1.0000 $\pm$ 0.0000 \\ 0.8333 $\pm$ 0.0000 \\ 0.7778 $\pm$ 0.0000 \\ 8.0000 $\pm$ 0.0000 \\ 1.4797 $\pm$ 0.0000}}} &
\multirow{5}{*}{\rotatebox[origin=c]{90}{\makecell[c]{0.6667 $\pm$ 0.0000 \\ 1.0000 $\pm$ 0.0000 \\ 0.8000 $\pm$ 0.0000 \\ 0.7222 $\pm$ 0.0000 \\ 10.0000 $\pm$ 0.0000 \\ 2.0269 $\pm$ 0.0000}}} \\
&&&&&\\
&&&&&\\
&&&&&\\
&&&&&\\
\hline
\end{tabular} %
}
}
\caption{Performance metrics of Cubesat for \textit{i} across methods and models. Each cell has from left to right (\textbf{Precision}, \textbf{Recall}, \textbf{F1-Score}, \textbf{Accuracy}, \textbf{Edit Distance}, and \textbf{Spectral Distance}). Values in each cell are the mean $\pm$ standard deviation over the repeated experiments. The total of 50 $\times$ 5 $=$ 250 experiments. (na: not available)}
\label{tab:case_i_cs_experiments}
\end{table*}

\clearpage
\begin{table*}
\captionsetup{width=\textwidth}
\scalebox{0.98}{\resizebox{\textwidth}{!}{%
\fontsize{5}{8}\selectfont
\begin{tabular}{|c|c|c|c|c|c|} %
\hline
 & \multicolumn{1}{c|}{\textbf{1.77T/128k/128k}} & \multicolumn{1}{c|}{\textbf{1.77T/128k/128k}} & \multicolumn{1}{c|}{\textbf{140.6B/65k/6k}} & \multicolumn{1}{c|}{\textbf{70.6B/131k/8k}} & \multicolumn{1}{c|}{\textbf{14.8B/131k/5k}} \\
\hline
  & \multicolumn{1}{c|}{\textbf{gpt4-turbo-preview}} & \multicolumn{1}{c|}{\textbf{gpt4o}} & \multicolumn{1}{c|}{\textbf{mixtral:8x22b}} & \multicolumn{1}{c|}{\textbf{llama3.3:70b}} & \multicolumn{1}{c|}{\textbf{deepseek-r1:14b}} \\
\hline
\multirow{5}{*}{\rotatebox[origin=c]{90}{\textbf{LLM}}} & 
\multirow{5}{*}{\rotatebox[origin=c]{90}{\makecell[c]{0.8075 $\pm$ 0.0228 \\ 0.7091 $\pm$ 0.0242 \\ 0.7547 $\pm$ 0.0164 \\ 0.6898 $\pm$ 0.0200 \\ 15.2000 $\pm$ 0.9798 \\ 0.8185 $\pm$ 0.2308}}} &
\multirow{5}{*}{\rotatebox[origin=c]{90}{\makecell[c]{0.8117 $\pm$ 0.0258 \\ 0.6242 $\pm$ 0.0707 \\ 0.7034 $\pm$ 0.0481 \\ 0.6490 $\pm$ 0.0416 \\ 17.2000 $\pm$ 2.0396 \\ 1.2926 $\pm$ 0.3513}}} &
\multirow{5}{*}{\rotatebox[origin=c]{90}{\makecell[c]{0.8258 $\pm$ 0.0240 \\ 0.8788 $\pm$ 0.1212 \\ 0.8486 $\pm$ 0.0743 \\ \textcolor{gold}{\textbf{0.7959 $\pm$ 0.0816}} \\ 10.0000 $\pm$ 4.000 \\ 0.9162 $\pm$ 0.1020}}} &
\multirow{5}{*}{\rotatebox[origin=c]{90}{\makecell[c]{0.7241 $\pm$ 0.0000 \\ 0.6364 $\pm$ 0.0000 \\ 0.6774 $\pm$ 0.0000 \\ 0.5918 $\pm$ 0.0000 \\ 20.0000 $\pm$ 0.000 \\ 0.8735 $\pm$ 0.0000}}} &
\multirow{5}{*}{\rotatebox[origin=c]{90}{\makecell[c]{0.8657 $\pm$ 0.0018 \\ 0.7818 $\pm$ 0.0121 \\ 0.8216 $\pm$ 0.0076 \\ \textcolor{silver}{\textbf{0.7714 $\pm$ 0.0082}} \\ 11.2000 $\pm$ 0.4000 \\ 1.0424 $\pm$ 0.0850}}} \\ 
&&&&&\\
&&&&&\\
&&&&&\\
&&&&&\\
\hline
\multirow{5}{*}{\rotatebox[origin=c]{90}{\textbf{RAG R1}}} & 
\multirow{5}{*}{\rotatebox[origin=c]{90}{\makecell[c]{0.7046 $\pm$ 0.0457 \\ 0.7333 $\pm$ 0.0485 \\ 0.7159 $\pm$ 0.0171 \\ 0.6082 $\pm$ 0.0327 \\ 19.2000 $\pm$ 1.6000 \\ 1.3264 $\pm$ 0.4612}}} &
\multirow{5}{*}{\rotatebox[origin=c]{90}{\makecell[c]{na \\ na \\ na \\ na \\ na \\ na}}} &
\multirow{5}{*}{\rotatebox[origin=c]{90}{\makecell[c]{0.7647 $\pm$ 0.0142 \\ 0.7455 $\pm$ 0.0242 \\ 0.7544 $\pm$ 0.0063 \\ 0.6735 $\pm$ 0.0000 \\ 16.0000 $\pm$ 0.0000 \\ 0.7625 $\pm$ 0.1447}}} &
\multirow{5}{*}{\rotatebox[origin=c]{90}{\makecell[c]{0.7391 $\pm$ 0.0000 \\ 0.5152 $\pm$ 0.0000 \\ 0.6071 $\pm$ 0.0000 \\ 0.5510 $\pm$ 0.0000 \\ 22.000 $\pm$ 0.000 \\ 1.9079 $\pm$ 0.0000}}} &
\multirow{5}{*}{\rotatebox[origin=c]{90}{\makecell[c]{0.6502 $\pm$ 0.3327 \\ 0.5879 $\pm$ 0.3334 \\ 0.6063 $\pm$ 0.3139 \\ 0.5592 $\pm$ 0.2899 \\ 24.8000 $\pm$ 20.4490 \\ 3.1968 $\pm$ 3.1978}}} \\
&&&&&\\
&&&&&\\
&&&&&\\
&&&&&\\
\hline
\multirow{5}{*}{\rotatebox[origin=c]{90}{\textbf{RAG R2}}} & 
\multirow{5}{*}{\rotatebox[origin=c]{90}{\makecell[c]{0.7758 $\pm$ 0.0218 \\ 0.7091 $\pm$ 0.0242 \\ 0.7404 $\pm$ 0.0124 \\ 0.6653 $\pm$ 0.0163 \\ 16.4000 $\pm$ 0.8000 \\ 0.8585 $\pm$ 0.1645}}} &
\multirow{5}{*}{\rotatebox[origin=c]{90}{\makecell[c]{na \\ na \\ na \\ na \\ na \\ na}}} &
\multirow{5}{*}{\rotatebox[origin=c]{90}{\makecell[c]{0.7980 $\pm$ 0.0575 \\ 0.6121 $\pm$ 0.0121 \\ 0.6924 $\pm$ 0.0288 \\ 0.6327 $\pm$ 0.0408 \\ 18.0000 $\pm$ 2.0000 \\ 2.1551 $\pm$ 0.2038}}} &
\multirow{5}{*}{\rotatebox[origin=c]{90}{\makecell[c]{0.6970 $\pm$ 0.000 \\ 0.6970 $\pm$ 0.000 \\ 0.6970 $\pm$ 0.0000 \\ 0.5918 $\pm$ 0.0000 \\ 20.0000 $\pm$ 0.000 \\ 1.3170 $\pm$ 0.0000}}} &
\multirow{5}{*}{\rotatebox[origin=c]{90}{\makecell[c]{0.6554 $\pm$ 0.3285 \\ 0.5273 $\pm$ 0.2694 \\ 0.5838 $\pm$ 0.2950 \\ 0.5388 $\pm$ 0.2731 \\ 25.8000 $\pm$ 19.7221 \\ 3.0330 $\pm$ 3.2685}}} \\
&&&&&\\
&&&&&\\
&&&&&\\
&&&&&\\
\hline
\multirow{5}{*}{\rotatebox[origin=c]{90}{\textbf{RAG R3}}} & 
\multirow{5}{*}{\rotatebox[origin=c]{90}{\makecell[c]{0.7900 $\pm$ 0.0061 \\ 0.6848 $\pm$ 0.0242 \\ 0.7335 $\pm$ 0.0168 \\ 0.6653 $\pm$ 0.0163 \\ 16.4000 $\pm$ 0.8000 \\ 0.9554 $\pm$ 0.0741}}} &
\multirow{5}{*}{\rotatebox[origin=c]{90}{\makecell[c]{na \\ na \\ na \\ na \\ na \\ na}}} &
\multirow{5}{*}{\rotatebox[origin=c]{90}{\makecell[c]{0.6848 $\pm$ 0.0147 \\ 0.6606 $\pm$ 0.0485 \\ 0.6721 $\pm$ 0.0316 \\ 0.5673 $\pm$ 0.0327 \\ 21.2000 $\pm$ 1.6000 \\ 1.1346 $\pm$ 0.0810}}} &
\multirow{5}{*}{\rotatebox[origin=c]{90}{\makecell[c]{0.6774 $\pm$ 0.0000 \\ 0.6364 $\pm$ 0.0000 \\ 0.6562 $\pm$ 0.0000 \\ 0.5510 $\pm$ 0.0000 \\ 22.000 $\pm$ 0.000 \\ 1.1751 $\pm$ 0.0000}}} &
\multirow{5}{*}{\rotatebox[origin=c]{90}{\makecell[c]{na \\ na \\ na \\ na \\ na \\ na}}} \\
&&&&&\\
&&&&&\\
&&&&&\\
&&&&&\\
\hline
\multirow{5}{*}{\rotatebox[origin=c]{90}{\textbf{RAG R1-R2}}} & 
\multirow{5}{*}{\rotatebox[origin=c]{90}{\makecell[c]{0.7279 $\pm$ 0.0388 \\ 0.7333 $\pm$ 0.0297 \\ 0.7291 $\pm$ 0.0107 \\ 0.6327 $\pm$ 0.0258 \\ 18.0000 $\pm$ 1.2649 \\ 0.9718 $\pm$ 0.2526}}} &
\multirow{5}{*}{\rotatebox[origin=c]{90}{\makecell[c]{na \\ na \\ na \\ na \\ na \\ na}}} &
\multirow{5}{*}{\rotatebox[origin=c]{90}{\makecell[c]{0.7108 $\pm$ 0.0623 \\ 0.6606 $\pm$ 0.0727 \\ 0.6847 $\pm$ 0.0681 \\ 0.5918 $\pm$ 0.0816 \\ 20.0000 $\pm$ 4.0000 \\ 0.8909 $\pm$ 0.0658}}} &
\multirow{5}{*}{\rotatebox[origin=c]{90}{\makecell[c]{0.7108 $\pm$ 0.0142 \\ 0.5636 $\pm$ 0.0242 \\ 0.6281 $\pm$ 0.0105 \\ 0.5510 $\pm$ 0.0000 \\ 22.000 $\pm$ 0.0000 \\ 1.1997 $\pm$ 0.3541}}} &
\multirow{5}{*}{\rotatebox[origin=c]{90}{\makecell[c]{0.5904 $\pm$ 0.3019 \\ 0.6364 $\pm$ 0.3325 \\ 0.6050 $\pm$ 0.3037 \\ 0.5265 $\pm$ 0.2656 \\ 26.4000 $\pm$ 19.3763 \\ 3.1958 $\pm$ 3.2034}}} \\
&&&&&\\
&&&&&\\
&&&&&\\
&&&&&\\
\hline
\multirow{5}{*}{\rotatebox[origin=c]{90}{\textbf{RAG R1-R3}}} & 
\multirow{5}{*}{\rotatebox[origin=c]{90}{\makecell[c]{0.7265 $\pm$ 0.0497 \\ 0.7212 $\pm$ 0.0297 \\ 0.7218 $\pm$ 0.0165 \\ 0.6245 $\pm$ 0.0400 \\ 18.4000 $\pm$ 1.9596 \\ 1.0816 $\pm$ 0.5881}}} &
\multirow{5}{*}{\rotatebox[origin=c]{90}{\makecell[c]{na \\ na \\ na \\ na \\ na \\ na}}} &
\multirow{5}{*}{\rotatebox[origin=c]{90}{\makecell[c]{0.7419 $\pm$ 0.0000 \\ 0.6970 $\pm$ 0.0000 \\ 0.7188 $\pm$ 0.0000 \\ 0.6327 $\pm$ 0.0000 \\ 18.0000 $\pm$ 0.000 \\ 0.9238 $\pm$ 0.0000}}} &
\multirow{5}{*}{\rotatebox[origin=c]{90}{\makecell[c]{0.7037 $\pm$ 0.0000 \\ 0.5758 $\pm$ 0.0000 \\ 0.6333 $\pm$ 0.0000 \\ 0.5510 $\pm$ 0.0000 \\ 22.000 $\pm$ 0.000 \\ 1.0226 $\pm$ 0.0000}}} &
\multirow{5}{*}{\rotatebox[origin=c]{90}{\makecell[c]{na \\ na \\ na \\ na \\ na \\ na}}} \\
&&&&&\\
&&&&&\\
&&&&&\\
&&&&&\\
\hline
\multirow{5}{*}{\rotatebox[origin=c]{90}{\textbf{RAG R2-R3}}} & 
\multirow{5}{*}{\rotatebox[origin=c]{90}{\makecell[c]{0.7624 $\pm$ 0.0251 \\ 0.6970 $\pm$ 0.0000 \\ 0.7280 $\pm$ 0.0114 \\ 0.6490 $\pm$ 0.02000 \\ 17.2000 $\pm$ 0.9798 \\ 0.9683 $\pm$ 0.0408}}} &
\multirow{5}{*}{\rotatebox[origin=c]{90}{\makecell[c]{na \\ na \\ na \\ na \\ na \\ na}}} &
\multirow{5}{*}{\rotatebox[origin=c]{90}{\makecell[c]{0.7980 $\pm$ 0.0575 \\ 0.6121 $\pm$ 0.0121 \\ 0.6924 $\pm$ 0.0288 \\ 0.6327 $\pm$ 0.0408 \\ 18.0000 $\pm$ 2.000 \\ 2.1441 $\pm$ 0.2258}}} &
\multirow{5}{*}{\rotatebox[origin=c]{90}{\makecell[c]{0.6970 $\pm$ 0.0000 \\ 0.6970 $\pm$ 0.0000 \\ 0.6970 $\pm$ 0.0000 \\ 0.5918 $\pm$ 0.0000 \\ 20.0000 $\pm$ 0.0000 \\ 1.3170 $\pm$ 0.0000}}} &
\multirow{5}{*}{\rotatebox[origin=c]{90}{\makecell[c]{na \\ na \\ na \\ na \\ na \\ na}}} \\
&&&&&\\
&&&&&\\
&&&&&\\
&&&&&\\
\hline
\multirow{5}{*}{\rotatebox[origin=c]{90}{\textbf{RAG R1-R2-R3}}} & 
\multirow{5}{*}{\rotatebox[origin=c]{90}{\makecell[c]{0.7277 $\pm$ 0.0489 \\ 0.7697 $\pm$ 0.0594 \\ 0.7469 $\pm$ 0.0457 \\ 0.6490 $\pm$ 0.0663 \\ 17.2000 $\pm$ 3.2496 \\ 1.1314 $\pm$ 0.5593}}} &
\multirow{5}{*}{\rotatebox[origin=c]{90}{\makecell[c]{na \\ na \\ na \\ na \\ na \\ na}}} &
\multirow{5}{*}{\rotatebox[origin=c]{90}{\makecell[c]{0.7218 $\pm$ 0.0404 \\ 0.7091 $\pm$ 0.0242 \\ 0.7139 $\pm$ 0.0097 \\ 0.6163 $\pm$ 0.0327 \\ 18.8000 $\pm$ 1.6000 \\ 1.0125 $\pm$ 0.1774}}} &
\multirow{5}{*}{\rotatebox[origin=c]{90}{\makecell[c]{0.7108 $\pm$ 0.0142 \\ 0.5636 $\pm$ 0.0242 \\ 0.6281 $\pm$ 0.0105 \\ 0.5510 $\pm$ 0.0000 \\ 22.000 $\pm$ 0.000 \\ 1.1997 $\pm$ 0.3541}}} &
\multirow{5}{*}{\rotatebox[origin=c]{90}{\makecell[c]{0.6643 $\pm$ 0.3335 \\ 0.5758 $\pm$ 0.2925 \\ 0.6146 $\pm$ 0.3077 \\ 0.5673 $\pm$ 0.2838 \\ 24.4000 $\pm$ 20.3037 \\ 2.7900 $\pm$ 3.3631}}} \\
&&&&&\\
&&&&&\\
&&&&&\\
&&&&&\\
\hline
\multirow{5}{*}{\rotatebox[origin=c]{90}{\textbf{\makecell[c]{Graph RAG \\ R1-R2}}}} & 
\multirow{5}{*}{\rotatebox[origin=c]{90}{\makecell[c]{0.7419 $\pm$ 0.0764 \\ 0.7091 $\pm$ 0.1504 \\ 0.7206 $\pm$ 0.1098 \\ 0.6408 $\pm$ 0.1275 \\ 17.600 $\pm$ 6.2482 \\ 1.0876 $\pm$ 0.2668}}} &
\multirow{5}{*}{\rotatebox[origin=c]{90}{\makecell[c]{na \\ na \\ na \\ na \\ na \\ na}}} &
\multirow{5}{*}{\rotatebox[origin=c]{90}{\makecell[c]{0.7521 $\pm$ 0.0094 \\ 0.8788 $\pm$ 0.0606 \\ 0.8092 $\pm$ 0.0185 \\ 0.7224 $\pm$ 0.0163 \\ 13.600 $\pm$ 0.800 \\ 2.0466 $\pm$ 0.2143}}} &
\multirow{5}{*}{\rotatebox[origin=c]{90}{\makecell[c]{0.8519 $\pm$ 0.0000 \\ 0.6970 $\pm$ 0.000 \\ 0.7667 $\pm$ 0.0000 \\ 0.7143 $\pm$ 0.0000 \\ 14.0000 $\pm$ 0.0000 \\ 1.0334 $\pm$ 0.0000}}} &
\multirow{5}{*}{\rotatebox[origin=c]{90}{\makecell[c]{0.8400 $\pm$ 0.000 \\ 0.6364 $\pm$ 0.000 \\ 0.7241 $\pm$ 0.000 \\ 0.6735 $\pm$ 0.000 \\ 16.000 $\pm$ 0.000 \\ 1.7049 $\pm$ 0.000}}} \\
&&&&&\\
&&&&&\\
&&&&&\\
&&&&&\\
\hline
\multirow{5}{*}{\rotatebox[origin=c]{90}{\textbf{\makecell[c]{Graph RAG \\ R2-R3}}}} & 
\multirow{5}{*}{\rotatebox[origin=c]{90}{\makecell[c]{0.8537 $\pm$ 0.0402 \\ 0.6970 $\pm$ 0.0000 \\ 0.7670 $\pm$ 0.0162 \\ 0.7143 $\pm$ 0.0258 \\ 14.0000 $\pm$ 1.2649 \\ 1.0194 $\pm$ 0.1150}}} &
\multirow{5}{*}{\rotatebox[origin=c]{90}{\makecell[c]{na \\ na \\ na \\ na \\ na \\ na}}} &
\multirow{5}{*}{\rotatebox[origin=c]{90}{\makecell[c]{0.7568 $\pm$ 0.0000 \\ 0.8485 $\pm$ 0.0000 \\ 0.8000 $\pm$ 0.0000 \\ 0.7143 $\pm$ 0.0000 \\ 14.0000 $\pm$ 0.0000 \\ 1.9394 $\pm$ 0.0000}}} &
\multirow{5}{*}{\rotatebox[origin=c]{90}{\makecell[c]{0.8519 $\pm$ 0.000 \\ 0.6970 $\pm$ 0.000 \\ 0.7667 $\pm$ 0.000 \\ 0.7143 $\pm$ 0.000 \\ 14.000 $\pm$ 0.000 \\ 1.0334 $\pm$ 0.000}}} &
\multirow{5}{*}{\rotatebox[origin=c]{90}{\makecell[c]{na \\ na \\ na \\ na \\ na \\ na}}} \\
&&&&&\\
&&&&&\\
&&&&&\\
&&&&&\\
\hline
\end{tabular} %
}
}
\caption{Performance metrics of power screwdriver for \textit{i} across methods and models. Each cell has from left to right (\textbf{Precision}, \textbf{Recall}, \textbf{F1-Score}, \textbf{Accuracy}, \textbf{Edit Distance}, and \textbf{Spectral Distance}). Values in each cell are the mean $\pm$ standard deviation over the repeated experiments. The total of 50 $\times$ 5 $=$ 250 experiments. (na: not available)}
\label{tab:case_i_ps_experiments}
\end{table*}

\clearpage
\begin{table*}
\captionsetup{width=\textwidth}
\scalebox{0.98}{\resizebox{\textwidth}{!}{%
\fontsize{5}{8}\selectfont
\begin{tabular}{|c|c|c|c|c|c|} %
\hline
 & \multicolumn{1}{c|}{\textbf{1.77T/128k/128k}} & \multicolumn{1}{c|}{\textbf{1.77T/128k/128k}} & \multicolumn{1}{c|}{\textbf{140.6B/65k/6k}} & \multicolumn{1}{c|}{\textbf{70.6B/131k/8k}} & \multicolumn{1}{c|}{\textbf{14.8B/131k/5k}} \\
\hline
  & \multicolumn{1}{c|}{\textbf{gpt4-turbo-preview}} & \multicolumn{1}{c|}{\textbf{gpt4o}} & \multicolumn{1}{c|}{\textbf{mixtral:8x22b}} & \multicolumn{1}{c|}{\textbf{llama3.3:70b}} & \multicolumn{1}{c|}{\textbf{deepseek-r1:14b}} \\
\hline
\multirow{5}{*}{\rotatebox[origin=c]{90}{\textbf{LLM}}} & 
\multirow{5}{*}{\rotatebox[origin=c]{90}{\makecell[c]{0.6168 $\pm$ 0.0160 \\ 0.9000 $\pm$ 0.0000 \\ 0.7319 $\pm$ 0.0112 \\ 0.6333 $\pm$ 0.0208 \\ 13.2000 $\pm$ 0.7483 \\ 1.4464 $\pm$ 0.1537}}} &
\multirow{5}{*}{\rotatebox[origin=c]{90}{\makecell[c]{0.5773 $\pm$ 0.0325 \\ 0.9800 $\pm$ 0.0400 \\ 0.7263 $\pm$ 0.0345 \\ 0.5889 $\pm$ 0.0567 \\ 14.8000 $\pm$ 2.0396 \\ 2.5534 $\pm$ 0.3398}}} &
\multirow{5}{*}{\rotatebox[origin=c]{90}{\makecell[c]{0.562 $\pm$ 0.000 \\ 0.900 $\pm$ 0.000 \\ 0.692 $\pm$ 0.000 \\ 0.556 $\pm$ 0.000 \\ 16.0000 $\pm$ 0.000 \\ 2.188 $\pm$ 0.000}}} &
\multirow{5}{*}{\rotatebox[origin=c]{90}{\makecell[c]{0.643 $\pm$ 0.0000 \\ 0.900 $\pm$ 0.0000 \\ 0.750 $\pm$ 0.0000 \\ 0.667 $\pm$ 0.0000 \\ 12.0000 $\pm$ 0.000 \\ 1.967 $\pm$ 0.0000}}} &
\multirow{5}{*}{\rotatebox[origin=c]{90}{\makecell[c]{0.6286 $\pm$ 0.0762 \\ 0.9000 $\pm$ 0.2000 \\ 0.7376 $\pm$ 0.1249 \\ 0.6611 $\pm$ 0.1222 \\ 12.2000 $\pm$ 4.4000 \\ 1.6060 $\pm$ 0.3896}}} \\
&&&&&\\
&&&&&\\
&&&&&\\
&&&&&\\
\hline
\multirow{5}{*}{\rotatebox[origin=c]{90}{\textbf{RAG R1}}} & 
\multirow{5}{*}{\rotatebox[origin=c]{90}{\makecell[c]{0.600 $\pm$ 0.000 \\ 0.900 $\pm$ 0.000 \\ 0.720 $\pm$ 0.000 \\ 0.611 $\pm$ 0.000 \\ 14.000 $\pm$ 0.000 \\ 2.027 $\pm$ 0.000}}} &
\multirow{5}{*}{\rotatebox[origin=c]{90}{\makecell[c]{na \\ na \\ na \\ na \\ na \\ na}}} &
\multirow{5}{*}{\rotatebox[origin=c]{90}{\makecell[c]{0.5857 $\pm$ 0.0050 \\ 0.9900 $\pm$ 0.0200 \\ 0.7360 $\pm$ 0.0095 \\ 0.6056 $\pm$ 0.0111 \\ 14.2000 $\pm$ 0.4000 \\ 2.4876 $\pm$ 0.0554}}} &
\multirow{5}{*}{\rotatebox[origin=c]{90}{\makecell[c]{0.778 $\pm$ 0.0000 \\ 0.700 $\pm$ 0.0000 \\ 0.737 $\pm$ 0.0000 \\ \textcolor{gold}{\textbf{0.722 $\pm$ 0.0000}} \\ 10.000 $\pm$ 0.000 \\ 1.560 $\pm$ 0.0000}}} &
\multirow{5}{*}{\rotatebox[origin=c]{90}{\makecell[c]{0.4444 $\pm$ 0.2222 \\ 0.8000 $\pm$ 0.4000 \\ 0.5714 $\pm$ 0.2857 \\ 0.4444 $\pm$ 0.2222 \\ 23.2000 $\pm$ 14.4000 \\ 4.2622 $\pm$ 2.1255}}} \\
&&&&&\\
&&&&&\\
&&&&&\\
&&&&&\\
\hline
\multirow{5}{*}{\rotatebox[origin=c]{90}{\textbf{RAG R2}}} & 
\multirow{5}{*}{\rotatebox[origin=c]{90}{\makecell[c]{0.5574 $\pm$ 0.0400 \\ 0.6700 $\pm$ 0.1166 \\ 0.6032 $\pm$ 0.0471 \\ 0.5167 $\pm$ 0.0515 \\ 17.4000 $\pm$ 1.8547 \\ 1.8623 $\pm$ 0.3496}}} &
\multirow{5}{*}{\rotatebox[origin=c]{90}{\makecell[c]{na \\ na \\ na \\ na \\ na \\ na}}} &
\multirow{5}{*}{\rotatebox[origin=c]{90}{\makecell[c]{0.625 $\pm$ 0.000 \\ 1.000 $\pm$ 0.000 \\ 0.769 $\pm$ 0.000 \\ 0.667 $\pm$ 0.000 \\ 12.0000 $\pm$ 0.000 \\ 2.188 $\pm$ 0.000}}} &
\multirow{5}{*}{\rotatebox[origin=c]{90}{\makecell[c]{0.5455 $\pm$ 0.0000 \\ 0.6000 $\pm$ 0.0000 \\ 0.5714 $\pm$ 0.0000 \\ 0.5000 $\pm$ 0.0000 \\ 18.0000 $\pm$ 0.0000 \\ 0.9358 $\pm$ 0.0000}}} &
\multirow{5}{*}{\rotatebox[origin=c]{90}{\makecell[c]{0.5773 $\pm$ 0.0220 \\ 0.9600 $\pm$ 0.0800 \\ 0.7206 $\pm$ 0.0403 \\ 0.5889 $\pm$ 0.0444 \\ 14.8000 $\pm$ 1.6000 \\ 2.3185 $\pm$ 0.2829}}} \\
&&&&&\\
&&&&&\\
&&&&&\\
&&&&&\\
\hline
\multirow{5}{*}{\rotatebox[origin=c]{90}{\textbf{RAG R3}}} & 
\multirow{5}{*}{\rotatebox[origin=c]{90}{\makecell[c]{0.5559 $\pm$ 0.0096 \\ 0.6400 $\pm$ 0.1319 \\ 0.5898 $\pm$ 0.0530 \\ 0.5167 $\pm$ 0.0222 \\ 17.4000 $\pm$ 0.8000 \\ 1.8006 $\pm$ 0.0777}}} &
\multirow{5}{*}{\rotatebox[origin=c]{90}{\makecell[c]{na \\ na \\ na \\ na \\ na \\ na}}} &
\multirow{5}{*}{\rotatebox[origin=c]{90}{\makecell[c]{0.500 $\pm$ 0.000 \\ 0.500 $\pm$ 0.000 \\ 0.500 $\pm$ 0.000 \\ 0.444 $\pm$ 0.000 \\ 20.000 $\pm$ 0.0000 \\ 1.182 $\pm$ 0.000}}} &
\multirow{5}{*}{\rotatebox[origin=c]{90}{\makecell[c]{0.667 $\pm$ 0.0000 \\ 0.600 $\pm$ 0.0000 \\ 0.632 $\pm$ 0.0000 \\ 0.611 $\pm$ 0.0000 \\ 14.000 $\pm$ 0.000 \\ 1.560 $\pm$ 0.0000}}} &
\multirow{5}{*}{\rotatebox[origin=c]{90}{\makecell[c]{na \\ na \\ na \\ na \\ na \\ na}}} \\
&&&&&\\
&&&&&\\
&&&&&\\
&&&&&\\
\hline
\multirow{5}{*}{\rotatebox[origin=c]{90}{\textbf{RAG R1-R2}}} & 
\multirow{5}{*}{\rotatebox[origin=c]{90}{\makecell[c]{0.5310 $\pm$ 0.0555 \\ 0.6800 $\pm$ 0.0400 \\ 0.5930 $\pm$ 0.0252 \\ 0.4778 $\pm$ 0.0711 \\ 18.8000 $\pm$ 2.5612 \\ 1.6863 $\pm$ 0.3543}}} &
\multirow{5}{*}{\rotatebox[origin=c]{90}{\makecell[c]{na \\ na \\ na \\ na \\ na \\ na}}} &
\multirow{5}{*}{\rotatebox[origin=c]{90}{\makecell[c]{0.625 $\pm$ 0.000 \\ 1.000 $\pm$ 0.000 \\ 0.769 $\pm$ 0.000 \\ 0.667 $\pm$ 0.000 \\ 12.0000 $\pm$ 0.0000 \\   2.188 $\pm$ 0.000}}} &
\multirow{5}{*}{\rotatebox[origin=c]{90}{\makecell[c]{0.562 $\pm$ 0.000 \\ 0.900 $\pm$ 0.000 \\ 0.692 $\pm$ 0.000 \\ 0.556 $\pm$ 0.0000 \\ 16.000 $\pm$ 0.0000 \\ 2.188 $\pm$ 0.000}}} &
\multirow{5}{*}{\rotatebox[origin=c]{90}{\makecell[c]{0.5308 $\pm$ 0.0154 \\ 0.7000 $\pm$ 0.0000 \\ 0.6036 $\pm$ 0.0101 \\ 0.4889 $\pm$ 0.0222 \\ 18.4000 $\pm$ 0.8000 \\ 1.4819 $\pm$ 0.0584}}} \\
&&&&&\\
&&&&&\\
&&&&&\\
&&&&&\\
\hline
\multirow{5}{*}{\rotatebox[origin=c]{90}{\textbf{RAG R1-R3}}} & 
\multirow{5}{*}{\rotatebox[origin=c]{90}{\makecell[c]{0.556 $\pm$ 0.000 \\ 1.000 $\pm$ 0.000 \\ 0.714 $\pm$ 0.000 \\ 0.556 $\pm$ 0.000 \\ 16.000 $\pm$ 0.000 \\ 3.199 $\pm$ 0.000}}} &
\multirow{5}{*}{\rotatebox[origin=c]{90}{\makecell[c]{na \\ na \\ na \\ na \\ na \\ na}}} &
\multirow{5}{*}{\rotatebox[origin=c]{90}{\makecell[c]{0.5937 $\pm$ 0.0109 \\ 1.0000 $\pm$ 0.0000 \\ 0.7450 $\pm$ 0.0085 \\ 0.6120 $\pm$ 0.0017 \\ 19.0000 $\pm$ 10.0000 \\ 3.3263 $\pm$ 1.7328}}} &
\multirow{5}{*}{\rotatebox[origin=c]{90}{\makecell[c]{0.778 $\pm$ 0.0000 \\ 0.700 $\pm$ 0.0000 \\ 0.737 $\pm$ 0.0000 \\ \textcolor{gold}{\textbf{0.722 $\pm$ 0.0000}} \\ 10.000 $\pm$ 0.000 \\ 1.560 $\pm$ 0.0000}}} &
\multirow{5}{*}{\rotatebox[origin=c]{90}{\makecell[c]{0.4444 $\pm$ 0.2222 \\ 0.8000 $\pm$ 0.4000 \\ 0.5714 $\pm$ 0.2857 \\ 0.4444 $\pm$ 0.2222 \\ 23.2000 $\pm$ 14.4000 \\ 4.2622 $\pm$ 2.1255}}} \\
&&&&&\\
&&&&&\\
&&&&&\\
&&&&&\\
\hline
\multirow{5}{*}{\rotatebox[origin=c]{90}{\textbf{RAG R2-R3}}} & 
\multirow{5}{*}{\rotatebox[origin=c]{90}{\makecell[c]{0.5674 $\pm$ 0.0142 \\ 0.5500 $\pm$ 0.0000 \\ 0.5585 $\pm$ 0.0069 \\ 0.5167 $\pm$ 0.0136 \\ 17.4000 $\pm$ 0.4899 \\ 1.7986 $\pm$ 0.0000}}} &
\multirow{5}{*}{\rotatebox[origin=c]{90}{\makecell[c]{na \\ na \\ na \\ na \\ na \\ na}}} &
\multirow{5}{*}{\rotatebox[origin=c]{90}{\makecell[c]{0.625 $\pm$ 0.000 \\ 1.000 $\pm$ 0.000 \\ 0.769 $\pm$ 0.000 \\ 0.667 $\pm$ 0.000 \\ 12.0000 $\pm$ 0.000 \\ 2.188 $\pm$ 0.000}}} &
\multirow{5}{*}{\rotatebox[origin=c]{90}{\makecell[c]{0.545 $\pm$ 0.0000 \\ 0.600 $\pm$ 0.0000 \\ 0.571 $\pm$ 0.0000 \\ 0.500 $\pm$ 0.0000 \\ 18.0000 $\pm$ 0.0000 \\ 0.936 $\pm$ 0.0000}}} &
\multirow{5}{*}{\rotatebox[origin=c]{90}{\makecell[c]{0.5773 $\pm$ 0.0220 \\ 0.9600 $\pm$ 0.0800 \\ 0.7206 $\pm$ 0.0403 \\ 0.5889 $\pm$ 0.0444 \\ 14.8000 $\pm$ 1.6000 \\ 2.3185 $\pm$ 0.2829}}} \\
&&&&&\\
&&&&&\\
&&&&&\\
&&&&&\\
\hline
\multirow{5}{*}{\rotatebox[origin=c]{90}{\textbf{RAG R1-R2-R3}}} & 
\multirow{5}{*}{\rotatebox[origin=c]{90}{\makecell[c]{0.5316 $\pm$ 0.0451 \\ 0.7600 $\pm$ 0.1594 \\ 0.6201 $\pm$ 0.0708 \\ 0.4944 $\pm$ 0.0733 \\ 18.2000 $\pm$ 2.6382 \\ 1.9084 $\pm$ 0.7942}}} &
\multirow{5}{*}{\rotatebox[origin=c]{90}{\makecell[c]{na \\ na \\ na \\ na \\ na \\ na}}} &
\multirow{5}{*}{\rotatebox[origin=c]{90}{\makecell[c]{0.625 $\pm$ 0.000 \\ 1.000 $\pm$ 0.000 \\ 0.769 $\pm$ 0.000 \\ 0.667 $\pm$ 0.000 \\ 12.000 $\pm$ 0.000 \\ 2.188 $\pm$ 0.000}}} &
\multirow{5}{*}{\rotatebox[origin=c]{90}{\makecell[c]{0.5864 $\pm$ 0.0236 \\ 0.6000 $\pm$ 0.0000 \\ 0.5929 $\pm$ 0.0124 \\ 0.5417 $\pm$ 0.0241 \\ 16.5000 $\pm$ 0.8660 \\ 1.2842 $\pm$ 0.0165}}} &
\multirow{5}{*}{\rotatebox[origin=c]{90}{\makecell[c]{na \\ na \\ na \\ na \\ na \\ na}}} \\
&&&&&\\
&&&&&\\
&&&&&\\
&&&&&\\
\hline
\multirow{5}{*}{\rotatebox[origin=c]{90}{\textbf{\makecell[c]{Graph RAG \\ R1-R2}}}} & 
\multirow{5}{*}{\rotatebox[origin=c]{90}{\makecell[c]{0.5783 $\pm$ 0.0290 \\ 0.9000 $\pm$ 0.1225 \\ 0.6983 $\pm$ 0.0202 \\ 0.5722 $\pm$ 0.0222 \\ 15.4000 $\pm$ 0.8000 \\ 2.3589 $\pm$ 1.0337}}} &
\multirow{5}{*}{\rotatebox[origin=c]{90}{\makecell[c]{na \\ na \\ na \\ na \\ na \\ na}}} &
\multirow{5}{*}{\rotatebox[origin=c]{90}{\makecell[c]{0.562 $\pm$ 0.000 \\ 0.900 $\pm$ 0.000 \\ 0.692 $\pm$ 0.000 \\ 0.556 $\pm$ 0.000 \\ 16.000 $\pm$ 0.000 \\ 2.188 $\pm$ 0.000}}} &
\multirow{5}{*}{\rotatebox[origin=c]{90}{\makecell[c]{0.615 $\pm$ 0.0000 \\ 0.800 $\pm$ 0.0000 \\ 0.696 $\pm$ 0.0000 \\ 0.611 $\pm$ 0.0000 \\ 14.0000 $\pm$ 0.0000 \\ 1.367 $\pm$ 0.0000}}} &
\multirow{5}{*}{\rotatebox[origin=c]{90}{\makecell[c]{0.581 $\pm$ 0.068 \\ 0.870 $\pm$ 0.214 \\ 0.693 $\pm$ 0.124 \\ 0.587 $\pm$ 0.110 \\ 15.800 $\pm$ 4.665 \\ 2.127 $\pm$ 0.506}}} \\
&&&&&\\
&&&&&\\
&&&&&\\
&&&&&\\
\hline
\multirow{5}{*}{\rotatebox[origin=c]{90}{\textbf{\makecell[c]{Graph RAG \\ R2-R3}}}} & 
\multirow{5}{*}{\rotatebox[origin=c]{90}{\makecell[c]{0.6030 $\pm$ 0.0583 \\ 0.9500 $\pm$ 0.1000 \\ 0.7314 $\pm$ 0.0343 \\ 0.6111 $\pm$ 0.0703 \\ 14.0000 $\pm$ 2.5298 \\ 2.4286 $\pm$ 0.9918}}} &
\multirow{5}{*}{\rotatebox[origin=c]{90}{\makecell[c]{na \\ na \\ na \\ na \\ na \\ na}}} &
\multirow{5}{*}{\rotatebox[origin=c]{90}{\makecell[c]{0.562 $\pm$ 0.000 \\ 0.900 $\pm$ 0.0000 \\ 0.692 $\pm$ 0.000 \\ 0.556 $\pm$ 0.000 \\ 16.0000 $\pm$ 0.000 \\ 2.188 $\pm$ 0.000}}} &
\multirow{5}{*}{\rotatebox[origin=c]{90}{\makecell[c]{0.615 $\pm$ 0.0000 \\ 0.800 $\pm$ 0.0000 \\ 0.696 $\pm$ 0.0000 \\ 0.611 $\pm$ 0.0000 \\ 14.0000 $\pm$ 0.0000 \\ 1.367 $\pm$ 0.0000}}} &
\multirow{5}{*}{\rotatebox[origin=c]{90}{\makecell[c]{0.581 $\pm$ 0.068 \\ 0.870 $\pm$ 0.214 \\ 0.693 $\pm$ 0.124 \\ 0.587 $\pm$ 0.110 \\ 15.800 $\pm$ 4.665 \\ 2.127 $\pm$ 0.506}}} \\
&&&&&\\
&&&&&\\
&&&&&\\
&&&&&\\
\hline
\end{tabular} %
}
}
\caption{Performance metrics of CubeSat for \textit{ii}-raw across methods and models. Each cell has from left to right (\textbf{Precision}, \textbf{Recall}, \textbf{F1-Score}, \textbf{Accuracy}, \textbf{Edit Distance}, and \textbf{Spectral Distance}). Values in each cell are the mean $\pm$ standard deviation over the repeated experiments. The total of 50 $\times$ 5 $=$ 250 experiments. (na: not available)}
\label{tab:case_ii_raw_cs_experiments}
\end{table*}

\clearpage
\begin{table*}
\captionsetup{width=\textwidth}
\scalebox{0.98}{\resizebox{\textwidth}{!}{%
\fontsize{5}{8}\selectfont
\begin{tabular}{|c|c|c|c|c|c|} %
\hline
 & \multicolumn{1}{c|}{\textbf{1.77T/128k/128k}} & \multicolumn{1}{c|}{\textbf{1.77T/128k/128k}} & \multicolumn{1}{c|}{\textbf{140.6B/65k/6k}} & \multicolumn{1}{c|}{\textbf{70.6B/131k/8k}} & \multicolumn{1}{c|}{\textbf{14.8B/131k/5k}} \\
\hline
  & \multicolumn{1}{c|}{\textbf{gpt4-turbo-preview}} & \multicolumn{1}{c|}{\textbf{gpt4o}} & \multicolumn{1}{c|}{\textbf{mixtral:8x22b}} & \multicolumn{1}{c|}{\textbf{llama3.3:70b}} & \multicolumn{1}{c|}{\textbf{deepseek-r1:14b}} \\
\hline
\multirow{5}{*}{\rotatebox[origin=c]{90}{\textbf{LLM}}} & 
\multirow{5}{*}{\rotatebox[origin=c]{90}{\makecell[c]{0.5888 $\pm$ 0.0106 \\ 0.8600 $\pm$ 0.0374 \\ 0.6989 $\pm$ 0.0199 \\ 0.5889 $\pm$ 0.0208 \\ 14.8000 $\pm$ 0.7483 \\ 1.4464 $\pm$ 0.1537}}} &
\multirow{5}{*}{\rotatebox[origin=c]{90}{\makecell[c]{0.5773 $\pm$ 0.0325 \\ 0.9800 $\pm$ 0.0400 \\ 0.7263 $\pm$ 0.0345 \\ 0.5889 $\pm$ 0.0567 \\ 14.8000 $\pm$ 2.0396 \\ 2.5534 $\pm$ 0.3398}}} &
\multirow{5}{*}{\rotatebox[origin=c]{90}{\makecell[c]{0.5000 $\pm$ 0.0000 \\ 0.8000 $\pm$ 0.0000 \\ 0.6154 $\pm$ 0.0000 \\ 0.4444 $\pm$ 0.0000 \\ 20.0000 $\pm$ 0.0000 \\ 2.1876 $\pm$ 0.0000}}} &
\multirow{5}{*}{\rotatebox[origin=c]{90}{\makecell[c]{0.684 $\pm$ 0.000 \\ 0.867 $\pm$ 0.0000 \\ 0.765 $\pm$ 0.0000 \\ 0.680 $\pm$ 0.0000 \\ 8.0000 $\pm$ 0.000 \\ 1.282 $\pm$ 0.0000}}} &
\multirow{5}{*}{\rotatebox[origin=c]{90}{\makecell[c]{0.5562 $\pm$ 0.0876 \\ 0.8000 $\pm$ 0.2000 \\ 0.6540 $\pm$ 0.1319 \\ 0.5500 $\pm$ 0.1222 \\ 16.2000 $\pm$ 4.4000 \\ 1.6060 $\pm$ 0.3896}}} \\
&&&&&\\
&&&&&\\
&&&&&\\
&&&&&\\
\hline
\multirow{5}{*}{\rotatebox[origin=c]{90}{\textbf{RAG R1}}} & 
\multirow{5}{*}{\rotatebox[origin=c]{90}{\makecell[c]{0.429 $\pm$ 0.000 \\ 1.000 $\pm$ 0.000 \\ 0.600 $\pm$ 0.000 \\ 0.500 $\pm$ 0.000 \\ 8.000 $\pm$ 0.000 \\ 1.937 $\pm$ 0.000}}} &
\multirow{5}{*}{\rotatebox[origin=c]{90}{\makecell[c]{na \\ na \\ na \\ na \\ na \\ na}}} &
\multirow{5}{*}{\rotatebox[origin=c]{90}{\makecell[c]{0.5736 $\pm$ 0.0292 \\ 0.9700 $\pm$ 0.0600 \\ 0.7209 $\pm$ 0.0397 \\ 0.5833 $\pm$ 0.0556 \\ 15.0000 $\pm$ 2.0000 \\ 2.4876 $\pm$ 0.0554}}} &
\multirow{5}{*}{\rotatebox[origin=c]{90}{\makecell[c]{0.2941 $\pm$ 0.0000 \\ 0.5556 $\pm$ 0.0000 \\ 0.3846 $\pm$ 0.0000 \\ 0.3600 $\pm$ 0.0000 \\ 16.0000 $\pm$ 0.0000 \\ 2.1648 $\pm$ 0.0000}}} &
\multirow{5}{*}{\rotatebox[origin=c]{90}{\makecell[c]{0.4444 $\pm$ 0.2222 \\ 0.8000 $\pm$ 0.4000 \\ 0.5714 $\pm$ 0.2857 \\ 0.4444 $\pm$ 0.2222 \\ 23.2000 $\pm$ 14.4000 \\ 4.2622 $\pm$ 2.1255}}} \\
&&&&&\\
&&&&&\\
&&&&&\\
&&&&&\\
\hline
\multirow{5}{*}{\rotatebox[origin=c]{90}{\textbf{RAG R2}}} & 
\multirow{5}{*}{\rotatebox[origin=c]{90}{\makecell[c]{0.7984 $\pm$ 0.0472 \\ 0.7714 $\pm$ 0.1143 \\ 0.7794 $\pm$ 0.0523 \\ 0.6667 $\pm$ 0.0703 \\ 3.0000 $\pm$ 0.6325 \\ 1.5633 $\pm$ 0.5724}}} &
\multirow{5}{*}{\rotatebox[origin=c]{90}{\makecell[c]{na \\ na \\ na \\ na \\ na \\ na}}} &
\multirow{5}{*}{\rotatebox[origin=c]{90}{\makecell[c]{0.778 $\pm$ 0.000 \\ 1.000 $\pm$ 0.000 \\ 0.875 $\pm$ 0.000 \\ \textcolor{gold}{\textbf{0.778 $\pm$ 0.000}} \\ 2.000 $\pm$ 0.000 \\ 1.231 $\pm$ 0.000}}} &
\multirow{5}{*}{\rotatebox[origin=c]{90}{\makecell[c]{0.3333 $\pm$ 0.0000 \\ 0.5556 $\pm$ 0.0000 \\ 0.4167 $\pm$ 0.0000 \\ 0.4400 $\pm$ 0.0000 \\ 14.0000 $\pm$ 0.0000 \\ 1.5291 $\pm$ 0.0000}}} &
\multirow{5}{*}{\rotatebox[origin=c]{90}{\makecell[c]{0.5639 $\pm$ 0.0486 \\ 0.9400 $\pm$ 0.1200 \\ 0.7046 $\pm$ 0.0723 \\ 0.5667 $\pm$ 0.0889 \\ 15.6000 $\pm$ 3.2000 \\ 2.3185 $\pm$ 0.2829}}} \\
&&&&&\\
&&&&&\\
&&&&&\\
&&&&&\\
\hline
\multirow{5}{*}{\rotatebox[origin=c]{90}{\textbf{RAG R3}}} & 
\multirow{5}{*}{\rotatebox[origin=c]{90}{\makecell[c]{0.6422 $\pm$ 0.0416 \\ 0.7300 $\pm$ 0.0927 \\ 0.6779 $\pm$ 0.0250 \\ 0.6167 $\pm$ 0.0369 \\ 13.8000 $\pm$ 1.3266 \\ 1.8006 $\pm$ 0.0777}}} &
\multirow{5}{*}{\rotatebox[origin=c]{90}{\makecell[c]{na \\ na \\ na \\ na \\ na \\ na}}} &
\multirow{5}{*}{\rotatebox[origin=c]{90}{\makecell[c]{na \\ na \\ na \\ na \\ na \\ na}}} &
\multirow{5}{*}{\rotatebox[origin=c]{90}{\makecell[c]{0.818 $\pm$ 0.0000 \\ 0.600 $\pm$ 0.0000 \\ 0.692 $\pm$ 0.0000 \\ 0.680 $\pm$ 0.0000 \\ 8.000 $\pm$ 0.000 \\ 1.001 $\pm$ 0.0000}}} &
\multirow{5}{*}{\rotatebox[origin=c]{90}{\makecell[c]{na \\ na \\ na \\ na \\ na \\ na}}} \\
&&&&&\\
&&&&&\\
&&&&&\\
&&&&&\\
\hline
\multirow{5}{*}{\rotatebox[origin=c]{90}{\textbf{RAG R1-R2}}} & 
\multirow{5}{*}{\rotatebox[origin=c]{90}{\makecell[c]{0.8103 $\pm$ 0.0963 \\ 0.7000 $\pm$ 0.0408 \\ 0.7451 $\pm$ 0.0205 \\ 0.6375 $\pm$ 0.0612 \\ 5.8000 $\pm$ 0.9798 \\ 1.9361 $\pm$ 1.0928}}} &
\multirow{5}{*}{\rotatebox[origin=c]{90}{\makecell[c]{na \\ na \\ na \\ na \\ na \\ na}}} &
\multirow{5}{*}{\rotatebox[origin=c]{90}{\makecell[c]{0.7778 $\pm$ 0.0000 \\ 1.0000 $\pm$ 0.0000 \\ 0.8750 $\pm$ 0.0000 \\ \textcolor{gold}{\textbf{0.7778 $\pm$ 0.0000}} \\ 2.0000 $\pm$ 0.0000 \\ 1.2307 $\pm$ 0.0000}}} &
\multirow{5}{*}{\rotatebox[origin=c]{90}{\makecell[c]{0.714 $\pm$ 0.000 \\ 1.000 $\pm$ 0.000 \\ 0.833 $\pm$ 0.000 \\ 0.760 $\pm$ 0.0000 \\ 6.000 $\pm$ 0.0000 \\ 1.411 $\pm$ 0.000}}} &
\multirow{5}{*}{\rotatebox[origin=c]{90}{\makecell[c]{0.5308 $\pm$ 0.0154 \\ 0.7000 $\pm$ 0.0000 \\ 0.6036 $\pm$ 0.0101 \\ 0.4889 $\pm$ 0.0222 \\ 18.4000 $\pm$ 0.8000 \\ 1.4819 $\pm$ 0.0584}}} \\
&&&&&\\
&&&&&\\
&&&&&\\
&&&&&\\
\hline
\multirow{5}{*}{\rotatebox[origin=c]{90}{\textbf{RAG R1-R3}}} & 
\multirow{5}{*}{\rotatebox[origin=c]{90}{\makecell[c]{0.556 $\pm$ 0.000 \\ 1.000 $\pm$ 0.000 \\ 0.714 $\pm$ 0.000 \\ 0.556 $\pm$ 0.000 \\ 16.000 $\pm$ 0.000 \\ 3.199 $\pm$ 0.000}}} &
\multirow{5}{*}{\rotatebox[origin=c]{90}{\makecell[c]{na \\ na \\ na \\ na \\ na \\ na}}} &
\multirow{5}{*}{\rotatebox[origin=c]{90}{\makecell[c]{0.5937 $\pm$ 0.0109 \\ 1.0000 $\pm$ 0.0000 \\ 0.7450 $\pm$ 0.0085 \\ 0.6120 $\pm$ 0.0017 \\ 19.0000 $\pm$ 10.0000 \\ 3.3263 $\pm$ 1.7328}}} &
\multirow{5}{*}{\rotatebox[origin=c]{90}{\makecell[c]{0.2941 $\pm$ 0.0000 \\ 0.5556 $\pm$ 0.0000 \\ 0.3846 $\pm$ 0.0000 \\ 0.3600 $\pm$ 0.0000 \\ 16.0000 $\pm$ 0.0000 \\ 2.1648 $\pm$ 0.0000}}} &
\multirow{5}{*}{\rotatebox[origin=c]{90}{\makecell[c]{0.4444 $\pm$ 0.2222 \\ 0.8000 $\pm$ 0.4000 \\ 0.5714 $\pm$ 0.2857 \\ 0.4444 $\pm$ 0.2222 \\ 23.2000 $\pm$ 14.4000 \\ 4.2622 $\pm$ 2.1255}}} \\
&&&&&\\
&&&&&\\
&&&&&\\
&&&&&\\
\hline
\multirow{5}{*}{\rotatebox[origin=c]{90}{\textbf{RAG R2-R3}}} & 
\multirow{5}{*}{\rotatebox[origin=c]{90}{\makecell[c]{0.6250 $\pm$ 0.0000 \\ 1.0000 $\pm$ 0.0000 \\ 0.7692 $\pm$ 0.0000 \\ 0.6667 $\pm$ 0.0000 \\ 3.0000 $\pm$ 0.0000 \\ 0.8740 $\pm$ 0.0000}}} &
\multirow{5}{*}{\rotatebox[origin=c]{90}{\makecell[c]{na \\ na \\ na \\ na \\ na \\ na}}} &
\multirow{5}{*}{\rotatebox[origin=c]{90}{\makecell[c]{0.714 $\pm$ 0.000 \\ 0.833 $\pm$ 0.000 \\ 0.769 $\pm$ 0.000 \\ 0.625 $\pm$ 0.000 \\ 6.0000 $\pm$ 0.000 \\ 0.506 $\pm$ 0.000}}} &
\multirow{5}{*}{\rotatebox[origin=c]{90}{\makecell[c]{0.333 $\pm$ 0.0000 \\ 0.556 $\pm$ 0.0000 \\ 0.417 $\pm$ 0.0000 \\ 0.440 $\pm$ 0.0000 \\ 14.0000 $\pm$ 0.0000 \\ 1.529 $\pm$ 0.0000}}} &
\multirow{5}{*}{\rotatebox[origin=c]{90}{\makecell[c]{0.5639 $\pm$ 0.0486 \\ 0.9400 $\pm$ 0.1200 \\ 0.7046 $\pm$ 0.0723 \\ 0.5667 $\pm$ 0.0889 \\ 15.6000 $\pm$ 3.2000 \\ 2.3185 $\pm$ 0.2829}}} \\
&&&&&\\
&&&&&\\
&&&&&\\
&&&&&\\
\hline
\multirow{5}{*}{\rotatebox[origin=c]{90}{\textbf{RAG R1-R2-R3}}} & 
\multirow{5}{*}{\rotatebox[origin=c]{90}{\makecell[c]{0.9111 $\pm$ 0.1089 \\ 0.8286 $\pm$ 0.1400 \\ 0.8500 $\pm$ 0.0204 \\ \textcolor{gold}{\textbf{0.7778 $\pm$ 0.0000}} \\ 2.0000 $\pm$ 0.0000 \\ 1.6923 $\pm$ 0.3769}}} &
\multirow{5}{*}{\rotatebox[origin=c]{90}{\makecell[c]{na \\ na \\ na \\ na \\ na \\ na}}} &
\multirow{5}{*}{\rotatebox[origin=c]{90}{\makecell[c]{0.3333 $\pm$ 0.0000 \\ 1.0000 $\pm$ 0.0000 \\ 0.5000 $\pm$ 0.0000 \\ 0.3333 $\pm$ 0.0000 \\ 6.0000 $\pm$ 0.0000 \\ 2.4495 $\pm$ 0.0000}}} &
\multirow{5}{*}{\rotatebox[origin=c]{90}{\makecell[c]{0.3529 $\pm$ 0.0340 \\ 0.6111 $\pm$ 0.0962 \\ 0.4471 $\pm$ 0.0527 \\ 0.4600 $\pm$ 0.0346 \\ 13.5000 $\pm$ 0.8660 \\ 1.6150 $\pm$ 0.1488}}} &
\multirow{5}{*}{\rotatebox[origin=c]{90}{\makecell[c]{na \\ na \\ na \\ na \\ na \\ na}}} \\
&&&&&\\
&&&&&\\
&&&&&\\
&&&&&\\
\hline
\multirow{5}{*}{\rotatebox[origin=c]{90}{\textbf{\makecell[c]{Graph RAG \\ R1-R2}}}} & 
\multirow{5}{*}{\rotatebox[origin=c]{90}{\makecell[c]{0.5000 $\pm$ 0.0000 \\ 1.0000 $\pm$ 0.0000 \\ 0.6667 $\pm$ 0.0000 \\ 0.5000 $\pm$ 0.0000 \\ 2.0000 $\pm$ 0.0000 \\ 1.4142 $\pm$ 0.0000}}} &
\multirow{5}{*}{\rotatebox[origin=c]{90}{\makecell[c]{na \\ na \\ na \\ na \\ na \\ na}}} &
\multirow{5}{*}{\rotatebox[origin=c]{90}{\makecell[c]{0.500 $\pm$ 0.000 \\ 0.800 $\pm$ 0.000 \\ 0.615 $\pm$ 0.000 \\ 0.444 $\pm$ 0.000 \\ 20.000 $\pm$ 0.000 \\ 2.188 $\pm$ 0.000}}} &
\multirow{5}{*}{\rotatebox[origin=c]{90}{\makecell[c]{0.647 $\pm$ 0.0000 \\ 0.733 $\pm$ 0.0000 \\ 0.688 $\pm$ 0.0000 \\ 0.600 $\pm$ 0.0000 \\ 10.0000 $\pm$ 0.0000 \\ 0.462 $\pm$ 0.0000}}} &
\multirow{5}{*}{\rotatebox[origin=c]{90}{\makecell[c]{0.534 $\pm$ 0.067 \\ 0.800 $\pm$ 0.200 \\ 0.637 $\pm$ 0.118 \\ 0.509 $\pm$ 0.088 \\ 18.6000 $\pm$ 3.323 \\ 2.127 $\pm$ 0.506}}} \\
&&&&&\\
&&&&&\\
&&&&&\\
&&&&&\\
\hline
\multirow{5}{*}{\rotatebox[origin=c]{90}{\textbf{\makecell[c]{Graph RAG \\ R2-R3}}}} & 
\multirow{5}{*}{\rotatebox[origin=c]{90}{\makecell[c]{0.6579 $\pm$ 0.0405 \\ 0.9400 $\pm$ 0.1200 \\ 0.7682 $\pm$ 0.0422 \\ 0.6500 $\pm$ 0.0500 \\ 5.6000 $\pm$ 0.8000 \\ 1.6905 $\pm$ 0.4315}}} &
\multirow{5}{*}{\rotatebox[origin=c]{90}{\makecell[c]{na \\ na \\ na \\ na \\ na \\ na}}} &
\multirow{5}{*}{\rotatebox[origin=c]{90}{\makecell[c]{0.625 $\pm$ 0.000 \\ 1.000 $\pm$ 0.0000 \\ 0.769 $\pm$ 0.000 \\ 0.625 $\pm$ 0.000 \\ 6.000 $\pm$ 0.000 \\ 2.036 $\pm$ 0.000}}} &
\multirow{5}{*}{\rotatebox[origin=c]{90}{\makecell[c]{0.647 $\pm$ 0.0000 \\ 0.733 $\pm$ 0.0000 \\ 0.688 $\pm$ 0.0000 \\ 0.600 $\pm$ 0.0000 \\ 10.0000 $\pm$ 0.0000 \\ 0.462 $\pm$ 0.0000}}} &
\multirow{5}{*}{\rotatebox[origin=c]{90}{\makecell[c]{0.534 $\pm$ 0.067 \\ 0.800 $\pm$ 0.200 \\ 0.637 $\pm$ 0.118 \\ 0.509 $\pm$ 0.088 \\ 18.600 $\pm$ 3.323 \\ 2.127 $\pm$ 0.506}}} \\
&&&&&\\
&&&&&\\
&&&&&\\
&&&&&\\
\hline
\end{tabular} %
}
}
\caption{Performance metrics of CubeSat for \textit{ii}-aligned across methods and models. Each cell has from left to right (\textbf{Precision}, \textbf{Recall}, \textbf{F1-Score}, \textbf{Accuracy}, \textbf{Edit Distance}, and \textbf{Spectral Distance}). Values in each cell are the mean $\pm$ standard deviation over the repeated experiments. The total of 50 $\times$ 5 $=$ 250 experiments. (na: not available)}
\label{tab:case_ii_aligned_cs_experiments}
\end{table*}

\clearpage
\begin{table*}
\captionsetup{width=\textwidth}
\scalebox{0.98}{\resizebox{\textwidth}{!}{%
\fontsize{5}{8}\selectfont
\begin{tabular}{|c|c|c|c|c|c|} %
\hline
 & \multicolumn{1}{c|}{\textbf{1.77T/128k/128k}} & \multicolumn{1}{c|}{\textbf{1.77T/128k/128k}} & \multicolumn{1}{c|}{\textbf{140.6B/65k/6k}} & \multicolumn{1}{c|}{\textbf{70.6B/131k/8k}} & \multicolumn{1}{c|}{\textbf{14.8B/131k/5k}} \\
\hline
  & \multicolumn{1}{c|}{\textbf{gpt4-turbo-preview}} & \multicolumn{1}{c|}{\textbf{gpt4o}} & \multicolumn{1}{c|}{\textbf{mixtral:8x22b}} & \multicolumn{1}{c|}{\textbf{llama3.3:70b}} & \multicolumn{1}{c|}{\textbf{deepseek-r1:14b}} \\
\hline
\multirow{5}{*}{\rotatebox[origin=c]{90}{\textbf{LLM}}} & 
\multirow{5}{*}{\rotatebox[origin=c]{90}{\makecell[c]{0.8304 $\pm$ 0.0037 \\ 0.8909 $\pm$ 0.0242 \\ 0.8595 $\pm$ 0.0131 \\ 0.8041 $\pm$ 0.0163 \\ 9.6000 $\pm$ 0.8000 \\ 1.1605 $\pm$ 0.0952}}} &
\multirow{5}{*}{\rotatebox[origin=c]{90}{\makecell[c]{0.9355 $\pm$ 0.0000 \\ 0.8788 $\pm$ 0.0000 \\ 0.9062 $\pm$ 0.0000 \\ \textcolor{gold}{\textbf{0.8776 $\pm$ 0.0000}} \\ 6.0000 $\pm$ 0.0000 \\ 0.9567 $\pm$ 0.0000}}} &
\multirow{5}{*}{\rotatebox[origin=c]{90}{\makecell[c]{0.6931 $\pm$ 0.0078 \\ 0.6848 $\pm$ 0.0242 \\ 0.6888 $\pm$ 0.0163 \\ 0.5837 $\pm$ 0.0163 \\ 20.4000 $\pm$ 0.8000 \\ 0.9858 $\pm$ 0.0145}}} &
\multirow{5}{*}{\rotatebox[origin=c]{90}{\makecell[c]{0.7037 $\pm$ 0.0000 \\ 0.5758 $\pm$ 0.0000 \\ 0.6333 $\pm$ 0.0000 \\ 0.5510 $\pm$ 0.0000 \\ 22.0000 $\pm$ 0.0000 \\ 0.7388 $\pm$ 0.0000}}} &
\multirow{5}{*}{\rotatebox[origin=c]{90}{\makecell[c]{0.7414 $\pm$ 0.0543 \\ 0.4667 $\pm$ 0.0242 \\ 0.5727 $\pm$ 0.0344 \\ 0.5306 $\pm$ 0.0408 \\ 23.0000 $\pm$ 2.0000 \\ 1.8567 $\pm$ 0.1851}}} \\
&&&&&\\
&&&&&\\
&&&&&\\
&&&&&\\
\hline
\multirow{5}{*}{\rotatebox[origin=c]{90}{\textbf{RAG R1}}} & 
\multirow{5}{*}{\rotatebox[origin=c]{90}{\makecell[c]{0.7926 $\pm$ 0.0513 \\ 0.6848 $\pm$ 0.0454 \\ 0.7337 $\pm$ 0.0375 \\ 0.6653 $\pm$ 0.0476 \\ 16.4000 $\pm$ 2.3324 \\ 1.3777 $\pm$ 0.2328}}} &
\multirow{5}{*}{\rotatebox[origin=c]{90}{\makecell[c]{na \\ na \\ na \\ na \\ na \\ na}}} &
\multirow{5}{*}{\rotatebox[origin=c]{90}{\makecell[c]{0.6106 $\pm$ 0.0489 \\ 0.5152 $\pm$ 0.0000 \\ 0.5580 $\pm$ 0.0192 \\ 0.4490 $\pm$ 0.0408 \\ 27.0000 $\pm$ 2.0000 \\ 1.0782 $\pm$ 0.1634}}} &
\multirow{5}{*}{\rotatebox[origin=c]{90}{\makecell[c]{0.6176 $\pm$ 0.0094 \\ 0.5879 $\pm$ 0.0242 \\ 0.6023 $\pm$ 0.0170 \\ 0.4776 $\pm$ 0.0163 \\ 25.6000 $\pm$ 0.8000 \\ 0.7731 $\pm$ 0.0830}}} &
\multirow{5}{*}{\rotatebox[origin=c]{90}{\makecell[c]{0.6353 $\pm$ 0.0506 \\ 0.5879 $\pm$ 0.0970 \\ 0.6066 $\pm$ 0.0626 \\ 0.4939 $\pm$ 0.0554 \\ 24.8000 $\pm$ 2.7129 \\ 1.7578 $\pm$ 0.7356}}} \\
&&&&&\\
&&&&&\\
&&&&&\\
&&&&&\\
\hline
\multirow{5}{*}{\rotatebox[origin=c]{90}{\textbf{RAG R2}}} & 
\multirow{5}{*}{\rotatebox[origin=c]{90}{\makecell[c]{0.8286 $\pm$ 0.0000 \\ 0.8788 $\pm$ 0.0000 \\ 0.8529 $\pm$ 0.0000 \\ \textcolor{gold}{\textbf{0.7959 $\pm$ 0.0000}} \\ 10.0000 $\pm$ 0.0000 \\ 1.2172 $\pm$ 0.0000}}} &
\multirow{5}{*}{\rotatebox[origin=c]{90}{\makecell[c]{na \\ na \\ na \\ na \\ na \\ na}}} &
\multirow{5}{*}{\rotatebox[origin=c]{90}{\makecell[c]{0.5329 $\pm$ 0.0308 \\ 0.5152 $\pm$ 0.0000 \\ 0.5235 $\pm$ 0.0144 \\ 0.3673 $\pm$ 0.0353 \\ 31.0000 $\pm$ 1.7321 \\ 0.8391 $\pm$ 0.0909}}} &
\multirow{5}{*}{\rotatebox[origin=c]{90}{\makecell[c]{0.8261 $\pm$ 0.0000 \\ 0.5758 $\pm$ 0.0000 \\ 0.6786 $\pm$ 0.0000 \\ 0.6327 $\pm$ 0.0000 \\ 18.0000 $\pm$ 0.0000 \\ 1.5726 $\pm$ 0.0000}}} &
\multirow{5}{*}{\rotatebox[origin=c]{90}{\makecell[c]{na \\ na \\ na \\ na \\ na \\ na}}} \\
&&&&&\\
&&&&&\\
&&&&&\\
&&&&&\\
\hline
\multirow{5}{*}{\rotatebox[origin=c]{90}{\textbf{RAG R3}}} & 
\multirow{5}{*}{\rotatebox[origin=c]{90}{\makecell[c]{0.7149 $\pm$ 0.0171 \\ 0.6970 $\pm$ 0.0383 \\ 0.7051 $\pm$ 0.0200 \\ 0.6082 $\pm$ 0.0200 \\ 19.2000 $\pm$ 0.9798 \\ 0.7041 $\pm$ 0.1892}}} &
\multirow{5}{*}{\rotatebox[origin=c]{90}{\makecell[c]{na \\ na \\ na \\ na \\ na \\ na}}} &
\multirow{5}{*}{\rotatebox[origin=c]{90}{\makecell[c]{0.7019 $\pm$ 0.0248 \\ 0.4545 $\pm$ 0.0000 \\ 0.5516 $\pm$ 0.0079 \\ 0.5020 $\pm$ 0.0163 \\ 24.4000 $\pm$ 0.8000 \\ 1.7682 $\pm$ 0.0120}}} &
\multirow{5}{*}{\rotatebox[origin=c]{90}{\makecell[c]{na \\ na \\ na \\ na \\ na \\ na}}} &
\multirow{5}{*}{\rotatebox[origin=c]{90}{\makecell[c]{0.6613 $\pm$ 0.0107 \\ 0.4848 $\pm$ 0.0000 \\ 0.5595 $\pm$ 0.0039 \\ 0.4857 $\pm$ 0.0082 \\ 25.2000 $\pm$ 0.4000 \\ 2.0021 $\pm$ 0.0035}}} \\
&&&&&\\
&&&&&\\
&&&&&\\
&&&&&\\
\hline
\multirow{5}{*}{\rotatebox[origin=c]{90}{\textbf{RAG R1-R2}}} & 
\multirow{5}{*}{\rotatebox[origin=c]{90}{\makecell[c]{0.8123 $\pm$ 0.0059 \\ 0.7879 $\pm$ 0.0303 \\ 0.7997 $\pm$ 0.0185 \\ 0.7347 $\pm$ 0.0204 \\ 13.0000 $\pm$ 1.0000 \\ 0.6247 $\pm$ 0.0787}}} &
\multirow{5}{*}{\rotatebox[origin=c]{90}{\makecell[c]{na \\ na \\ na \\ na \\ na \\ na}}} &
\multirow{5}{*}{\rotatebox[origin=c]{90}{\makecell[c]{0.7381 $\pm$ 0.0476 \\ 0.4242 $\pm$ 0.0606 \\ 0.5333 $\pm$ 0.0444 \\ 0.5061 $\pm$ 0.0082 \\ 24.2000 $\pm$ 0.4000 \\ 2.4896 $\pm$ 1.3047}}} &
\multirow{5}{*}{\rotatebox[origin=c]{90}{\makecell[c]{0.6129 $\pm$ 0.0000 \\ 0.5758 $\pm$ 0.0000 \\ 0.5938 $\pm$ 0.0000 \\ 0.4694 $\pm$ 0.0000 \\ 26.0000 $\pm$ 0.0000 \\ 0.8145 $\pm$ 0.0000}}} &
\multirow{5}{*}{\rotatebox[origin=c]{90}{\makecell[c]{0.6266 $\pm$ 0.0594 \\ 0.4970 $\pm$ 0.0309 \\ 0.5526 $\pm$ 0.0293 \\ 0.4571 $\pm$ 0.0493 \\ 26.6000 $\pm$ 2.4166 \\ 1.4582 $\pm$ 0.4459}}}  \\
&&&&&\\
&&&&&\\
&&&&&\\
&&&&&\\
\hline
\multirow{5}{*}{\rotatebox[origin=c]{90}{\textbf{RAG R1-R3}}} & 
\multirow{5}{*}{\rotatebox[origin=c]{90}{\makecell[c]{0.7553 $\pm$ 0.0198 \\ 0.7091 $\pm$ 0.0242 \\ 0.7312 $\pm$ 0.0160 \\ 0.6490 $\pm$ 0.0200 \\ 17.2000 $\pm$ 0.9798 \\ 1.3840 $\pm$ 0.1449}}} &
\multirow{5}{*}{\rotatebox[origin=c]{90}{\makecell[c]{na \\ na \\ na \\ na \\ na \\ na}}} &
\multirow{5}{*}{\rotatebox[origin=c]{90}{\makecell[c]{0.7591 $\pm$ 0.0374 \\ 0.4182 $\pm$ 0.0121 \\ 0.5392 $\pm$ 0.0196 \\ 0.5184 $\pm$ 0.0245 \\ 23.6000 $\pm$ 1.2000 \\ 2.3990 $\pm$ 0.0464}}} &
\multirow{5}{*}{\rotatebox[origin=c]{90}{\makecell[c]{0.6129 $\pm$ 0.0000 \\ 0.5758 $\pm$ 0.0000 \\ 0.5938 $\pm$ 0.0000 \\ 0.4694 $\pm$ 0.0000 \\ 26.0000 $\pm$ 0.0000 \\ 0.8145 $\pm$ 0.0000}}} &
\multirow{5}{*}{\rotatebox[origin=c]{90}{\makecell[c]{0.7221 $\pm$ 0.0162 \\ 0.5960 $\pm$ 0.0355 \\ 0.6523 $\pm$ 0.0208 \\ 0.5729 $\pm$ 0.0168 \\ 22.0000 $\pm$ 2.6077 \\ 1.2286 $\pm$ 0.3546}}} \\
&&&&&\\
&&&&&\\
&&&&&\\
&&&&&\\ 
\hline
\multirow{5}{*}{\rotatebox[origin=c]{90}{\textbf{RAG R2-R3}}} & 
\multirow{5}{*}{\rotatebox[origin=c]{90}{\makecell[c]{0.8286 $\pm$ 0.0000 \\ 0.8788 $\pm$ 0.0000 \\ 0.8529 $\pm$ 0.0000 \\ \textcolor{gold}{\textbf{0.7959 $\pm$ 0.0000}} \\ 10.0000 $\pm$ 0.0000 \\ 1.2172 $\pm$ 0.0000}}} &
\multirow{5}{*}{\rotatebox[origin=c]{90}{\makecell[c]{na \\ na \\ na \\ na \\ na \\ na}}} &
\multirow{5}{*}{\rotatebox[origin=c]{90}{\makecell[c]{0.7037 $\pm$ 0.0000 \\ 0.5758 $\pm$ 0.0000 \\ 0.6333 $\pm$ 0.0000 \\ 0.5510 $\pm$ 0.0000 \\ 22.0000 $\pm$ 0.0000 \\ 1.0144 $\pm$ 0.1622}}} &
\multirow{5}{*}{\rotatebox[origin=c]{90}{\makecell[c]{0.8261 $\pm$ 0.0000 \\ 0.5758 $\pm$ 0.0000 \\ 0.6786 $\pm$ 0.0000 \\ 0.6327 $\pm$ 0.0000 \\ 18.0000 $\pm$ 0.0000 \\ 1.5726 $\pm$ 0.0000}}} &
\multirow{5}{*}{\rotatebox[origin=c]{90}{\makecell[c]{0.7841 $\pm$ 0.0431 \\ 0.5697 $\pm$ 0.0445 \\ 0.6591 $\pm$ 0.0397 \\ 0.6041 $\pm$ 0.0420 \\ 19.4000 $\pm$ 2.0591 \\ 1.5518 $\pm$ 0.1979}}} \\
&&&&&\\
&&&&&\\
&&&&&\\
&&&&&\\
\hline
\multirow{5}{*}{\rotatebox[origin=c]{90}{\textbf{RAG R1-R2-R3}}} & 
\multirow{5}{*}{\rotatebox[origin=c]{90}{\makecell[c]{0.8309 $\pm$ 0.0040 \\ 0.8939 $\pm$ 0.0262 \\ 0.8611 $\pm$ 0.0142 \\ \textcolor{gold}{\textbf{0.8061 $\pm$ 0.0177}} \\ 9.5000 $\pm$ 0.8660 \\ 1.2075 $\pm$ 0.0167}}} &
\multirow{5}{*}{\rotatebox[origin=c]{90}{\makecell[c]{na \\ na \\ na \\ na \\ na \\ na}}} &
\multirow{5}{*}{\rotatebox[origin=c]{90}{\makecell[c]{0.7977 $\pm$ 0.0568 \\ 0.5394 $\pm$ 0.0727 \\ 0.6429 $\pm$ 0.0714 \\ 0.6000 $\pm$ 0.0653 \\ 19.6000 $\pm$ 3.2000 \\ 1.5165 $\pm$ 0.2719}}} &
\multirow{5}{*}{\rotatebox[origin=c]{90}{\makecell[c]{0.6129 $\pm$ 0.0000 \\ 0.5758 $\pm$ 0.0000 \\ 0.5938 $\pm$ 0.0000 \\ 0.4694 $\pm$ 0.0000 \\ 26.0000 $\pm$ 0.0000 \\ 0.8145 $\pm$ 0.0000}}} &
\multirow{5}{*}{\rotatebox[origin=c]{90}{\makecell[c]{0.6498 $\pm$ 0.0537 \\ 0.5697 $\pm$ 0.0675 \\ 0.6033 $\pm$ 0.0350 \\ 0.4980 $\pm$ 0.0420 \\ 24.6000 $\pm$ 2.0591 \\ 1.3945 $\pm$ 0.5032}}} \\
&&&&&\\
&&&&&\\
&&&&&\\
&&&&&\\
\hline
\multirow{5}{*}{\rotatebox[origin=c]{90}{\textbf{\makecell[c]{Graph RAG \\ R1-R2}}}} & 
\multirow{5}{*}{\rotatebox[origin=c]{90}{\makecell[c]{0.6970 $\pm$ 0.0000 \\ 0.6970 $\pm$ 0.0000 \\ 0.6970 $\pm$ 0.0000 \\ 0.5918 $\pm$ 0.0000 \\ 20.0000 $\pm$ 0.0000 \\ 0.8267 $\pm$ 0.0000}}} &
\multirow{5}{*}{\rotatebox[origin=c]{90}{\makecell[c]{na \\ na \\ na \\ na \\ na \\ na}}} &
\multirow{5}{*}{\rotatebox[origin=c]{90}{\makecell[c]{0.6237 $\pm$ 0.0119 \\ 0.5030 $\pm$ 0.0242 \\ 0.5568 $\pm$ 0.0198 \\ 0.4612 $\pm$ 0.0163 \\ 26.4000 $\pm$ 0.8000 \\ 1.0936 $\pm$ 0.1086}}} &
\multirow{5}{*}{\rotatebox[origin=c]{90}{\makecell[c]{0.6774 $\pm$ 0.0000 \\ 0.6364 $\pm$ 0.0000 \\ 0.6562 $\pm$ 0.0000 \\ 0.5510 $\pm$ 0.0000 \\ 22.0000 $\pm$ 0.0000 \\ 0.7825 $\pm$ 0.0000}}} &
\multirow{5}{*}{\rotatebox[origin=c]{90}{\makecell[c]{0.6566 $\pm$ 0.0540 \\ 0.4909 $\pm$ 0.0485 \\ 0.5573 $\pm$ 0.0187 \\ 0.4776 $\pm$ 0.0163 \\ 25.6000 $\pm$ 0.8000 \\ 1.8913 $\pm$ 0.2384}}} \\
&&&&&\\
&&&&&\\
&&&&&\\
&&&&&\\
\hline
\multirow{5}{*}{\rotatebox[origin=c]{90}{\textbf{\makecell[c]{Graph RAG \\ R2-R3}}}} & 
\multirow{5}{*}{\rotatebox[origin=c]{90}{\makecell[c]{0.5556 $\pm$ 0.0000 \\ 0.4545 $\pm$ 0.0000 \\ 0.5000 $\pm$ 0.0000 \\ 0.3878 $\pm$ 0.0000 \\ 30.0000 $\pm$ 0.0000 \\ 1.7721 $\pm$ 0.0000}}} &
\multirow{5}{*}{\rotatebox[origin=c]{90}{\makecell[c]{na \\ na \\ na \\ na \\ na \\ na}}} &
\multirow{5}{*}{\rotatebox[origin=c]{90}{\makecell[c]{0.6237 $\pm$ 0.0119 \\ 0.5030 $\pm$ 0.0242 \\ 0.5568 $\pm$ 0.0198 \\ 0.4612 $\pm$ 0.0163 \\ 26.4000 $\pm$ 0.8000 \\ 1.0936 $\pm$ 0.1086}}} &
\multirow{5}{*}{\rotatebox[origin=c]{90}{\makecell[c]{0.6774 $\pm$ 0.0000 \\ 0.6364 $\pm$ 0.0000 \\ 0.6562 $\pm$ 0.0000 \\ 0.5510 $\pm$ 0.0000 \\ 22.0000 $\pm$ 0.0000 \\ 0.7825 $\pm$ 0.0000}}} &
\multirow{5}{*}{\rotatebox[origin=c]{90}{\makecell[c]{na \\ na \\ na \\ na \\ na \\ na}}} \\
&&&&&\\
&&&&&\\
&&&&&\\
&&&&&\\
\hline
\end{tabular} %
}
}
\caption{Performance metrics of power screwdriver for \textit{ii}-raw across methods and models. Each cell has from left to right (\textbf{Precision}, \textbf{Recall}, \textbf{F1-Score}, \textbf{Accuracy}, \textbf{Edit Distance}, and \textbf{Spectral Distance}). Values in each cell are the mean $\pm$ standard deviation over the repeated experiments. The total of 50 $\times$ 5 $=$ 250 experiments. (na: not available)}
\label{tab:case_ii_raw_ps_experiments}
\end{table*}

\clearpage
\begin{table*}
\captionsetup{width=\textwidth}
\scalebox{0.98}{\resizebox{\textwidth}{!}{%
\fontsize{5}{8}\selectfont
\begin{tabular}{|c|c|c|c|c|c|} %
\hline
 & \multicolumn{1}{c|}{\textbf{1.77T/128k/128k}} & \multicolumn{1}{c|}{\textbf{1.77T/128k/128k}} & \multicolumn{1}{c|}{\textbf{140.6B/65k/6k}} & \multicolumn{1}{c|}{\textbf{70.6B/131k/8k}} & \multicolumn{1}{c|}{\textbf{14.8B/131k/5k}}  \\
\hline
  & \multicolumn{1}{c|}{\textbf{gpt4-turbo-preview}} & \multicolumn{1}{c|}{\textbf{gpt4o}} & \multicolumn{1}{c|}{\textbf{mixtral:8x22b}} & \multicolumn{1}{c|}{\textbf{llama3.3:70b}} & \multicolumn{1}{c|}{\textbf{deepseek-r1:14b}} \\
\hline
\multirow{5}{*}{\rotatebox[origin=c]{90}{\textbf{LLM}}} & 
\multirow{5}{*}{\rotatebox[origin=c]{90}{\makecell[c]{0.8187 $\pm$ 0.0340 \\ 0.9818 $\pm$ 0.0364 \\ 0.8927 $\pm$ 0.0329 \\ \textcolor{gold}{\textbf{0.8556 $\pm$ 0.0444}} \\ 5.2000 $\pm$ 1.6000 \\ 0.8595 $\pm$ 0.1307}}} &
\multirow{5}{*}{\rotatebox[origin=c]{90}{\makecell[c]{0.7692 $\pm$ 0.0000 \\ 0.9091 $\pm$ 0.0000 \\ 0.8333 $\pm$ 0.0000 \\ 0.7778 $\pm$ 0.0000 \\ 8.0000 $\pm$ 0.0000 \\ 0.8202 $\pm$ 0.0000}}} &
\multirow{5}{*}{\rotatebox[origin=c]{90}{\makecell[c]{0.7532 $\pm$ 0.0282 \\ 1.0000 $\pm$ 0.0000 \\ 0.8589 $\pm$ 0.0179 \\ 0.7760 $\pm$ 0.0320 \\ 5.6000 $\pm$ 0.8000 \\ 1.2250 $\pm$ 0.1718}}} &
\multirow{5}{*}{\rotatebox[origin=c]{90}{\makecell[c]{0.9000 $\pm$ 0.0000 \\ 0.8182 $\pm$ 0.0000 \\ 0.8571 $\pm$ 0.0000 \\ 0.8333 $\pm$ 0.0000 \\ 6.0000 $\pm$ 0.0000 \\ 0.3707 $\pm$ 0.0000}}} &
\multirow{5}{*}{\rotatebox[origin=c]{90}{\makecell[c]{0.7083 $\pm$ 0.0833 \\ 0.5636 $\pm$ 0.0364 \\ 0.6274 $\pm$ 0.0547 \\ 0.5889 $\pm$ 0.0667 \\ 14.8000 $\pm$ 2.4000 \\ 0.8486 $\pm$ 0.2310}}} \\
&&&&&\\
&&&&&\\
&&&&&\\
&&&&&\\
\hline
\multirow{5}{*}{\rotatebox[origin=c]{90}{\textbf{RAG R1}}} & 
\multirow{5}{*}{\rotatebox[origin=c]{90}{\makecell[c]{0.7193 $\pm$ 0.0271 \\ 0.7091 $\pm$ 0.1060 \\ 0.7121 $\pm$ 0.0650 \\ \textcolor{silver}{\textbf{0.6556 $\pm$ 0.0648}} \\ 12.4000 $\pm$ 2.3324 \\ 0.7966 $\pm$ 0.2268}}} &
\multirow{5}{*}{\rotatebox[origin=c]{90}{\makecell[c]{na \\ na \\ na \\ na \\ na \\ na}}} &
\multirow{5}{*}{\rotatebox[origin=c]{90}{\makecell[c]{0.6667 $\pm$ 0.0000 \\ 0.7091 $\pm$ 0.0364 \\ 0.6868 $\pm$ 0.0178 \\ 0.6056 $\pm$ 0.0111 \\ 14.2000 $\pm$ 0.4000 \\ 0.4400 $\pm$ 0.0796}}} &
\multirow{5}{*}{\rotatebox[origin=c]{90}{\makecell[c]{0.6824 $\pm$ 0.0198 \\ 0.8182 $\pm$ 0.0000 \\ 0.7440 $\pm$ 0.0120 \\ 0.6556 $\pm$ 0.0222 \\ 12.4000 $\pm$ 0.8000 \\ 1.1658 $\pm$ 0.1412}}} &
\multirow{5}{*}{\rotatebox[origin=c]{90}{\makecell[c]{0.7697 $\pm$ 0.0668 \\ 0.8091 $\pm$ 0.1532 \\ 0.7813 $\pm$ 0.0888 \\ 0.7333 $\pm$ 0.0853 \\ 9.6000 $\pm$ 3.0725 \\ 1.3642 $\pm$ 0.7536}}} \\
&&&&&\\
&&&&&\\
&&&&&\\
&&&&&\\
\hline
\multirow{5}{*}{\rotatebox[origin=c]{90}{\textbf{RAG R2}}} & 
\multirow{5}{*}{\rotatebox[origin=c]{90}{\makecell[c]{0.8462 $\pm$ 0.0000 \\ 1.0000 $\pm$ 0.0000 \\ 0.9167 $\pm$ 0.0000 \\ \textcolor{gold}{\textbf{0.8889 $\pm$ 0.0000}} \\ 4.0000 $\pm$ 0.0000 \\ 0.7532 $\pm$ 0.0000}}} &
\multirow{5}{*}{\rotatebox[origin=c]{90}{\makecell[c]{na \\ na \\ na \\ na \\ na \\ na}}} &
\multirow{5}{*}{\rotatebox[origin=c]{90}{\makecell[c]{0.7024 $\pm$ 0.0206 \\ 0.8636 $\pm$ 0.0787 \\ 0.7739 $\pm$ 0.0452 \\ 0.6944 $\pm$ 0.0481 \\ 11.0000 $\pm$ 1.7321 \\ 0.8772 $\pm$ 0.2754}}} &
\multirow{5}{*}{\rotatebox[origin=c]{90}{\makecell[c]{0.7273 $\pm$ 0.0000 \\ 0.7273 $\pm$ 0.0000 \\ 0.7273 $\pm$ 0.0000 \\ 0.6667 $\pm$ 0.0000 \\ 12.0000 $\pm$ 0.0000 \\ 0.6747 $\pm$ 0.0000}}} &
\multirow{5}{*}{\rotatebox[origin=c]{90}{\makecell[c]{na \\ na \\ na \\ na \\ na \\ na}}} \\
&&&&&\\
&&&&&\\
&&&&&\\
&&&&&\\
\hline
\multirow{5}{*}{\rotatebox[origin=c]{90}{\textbf{RAG R3}}} & 
\multirow{5}{*}{\rotatebox[origin=c]{90}{\makecell[c]{0.8303 $\pm$ 0.0061 \\ 0.8909 $\pm$ 0.0364 \\ 0.8593 $\pm$ 0.0206 \\ 0.8222 $\pm$ 0.0222 \\ 6.4000 $\pm$ 0.8000 \\ 0.8873 $\pm$ 0.0958}}} &
\multirow{5}{*}{\rotatebox[origin=c]{90}{\makecell[c]{na \\ na \\ na \\ na \\ na \\ na}}} &
\multirow{5}{*}{\rotatebox[origin=c]{90}{\makecell[c]{0.6923 $\pm$ 0.0000 \\ 0.6923 $\pm$ 0.0000 \\ 0.6923 $\pm$ 0.0000 \\ 0.6800 $\pm$ 0.0000 \\ 8.0000 $\pm$ 0.0000 \\ 1.4641 $\pm$ 0.0000}}} &
\multirow{5}{*}{\rotatebox[origin=c]{90}{\makecell[c]{na \\ na \\ na \\ na \\ na \\ na}}} &
\multirow{5}{*}{\rotatebox[origin=c]{90}{\makecell[c]{0.9857 $\pm$ 0.0286 \\ 0.7647 $\pm$ 0.0000 \\ 0.8611 $\pm$ 0.0112 \\ 0.8320 $\pm$ 0.0160 \\ 4.2000 $\pm$ 0.4000 \\ 0.9089 $\pm$ 0.0000}}} \\
&&&&&\\
&&&&&\\
&&&&&\\
&&&&&\\
\hline
\multirow{5}{*}{\rotatebox[origin=c]{90}{\textbf{RAG R1-R2}}} & 
\multirow{5}{*}{\rotatebox[origin=c]{90}{\makecell[c]{0.8333 $\pm$ 0.0000 \\ 0.9091 $\pm$ 0.0000 \\ 0.8696 $\pm$ 0.0000 \\ 0.8333 $\pm$ 0.0000 \\ 6.0000 $\pm$ 0.0000 \\ 0.4993 $\pm$ 0.0000}}} &
\multirow{5}{*}{\rotatebox[origin=c]{90}{\makecell[c]{na \\ na \\ na \\ na \\ na \\ na}}} &
\multirow{5}{*}{\rotatebox[origin=c]{90}{\makecell[c]{0.7288 $\pm$ 0.0731 \\ 0.5733 $\pm$ 0.0533 \\ 0.6360 $\pm$ 0.0137 \\ 0.6080 $\pm$ 0.0160 \\ 9.8000 $\pm$ 0.4000 \\ 1.2398 $\pm$ 0.9613}}} &
\multirow{5}{*}{\rotatebox[origin=c]{90}{\makecell[c]{0.7500 $\pm$ 0.0000 \\ 0.8182 $\pm$ 0.0000 \\ 0.7826 $\pm$ 0.0000 \\ 0.7222 $\pm$ 0.0000 \\ 10.0000 $\pm$ 0.0000 \\ 0.4993 $\pm$ 0.0000}}} &
\multirow{5}{*}{\rotatebox[origin=c]{90}{\makecell[c]{0.8243 $\pm$ 0.0403 \\ 0.7909 $\pm$ 0.1336 \\ 0.8039 $\pm$ 0.0886 \\ 0.7722 $\pm$ 0.0885 \\ 8.2000 $\pm$ 3.1875 \\ 0.9161 $\pm$ 0.3437}}} \\
&&&&&\\
&&&&&\\
&&&&&\\
&&&&&\\
\hline
\multirow{5}{*}{\rotatebox[origin=c]{90}{\textbf{RAG R1-R3}}} & 
\multirow{5}{*}{\rotatebox[origin=c]{90}{\makecell[c]{0.7493 $\pm$ 0.0133 \\ 0.8182 $\pm$ 0.0575 \\ 0.7817 $\pm$ 0.0336 \\ 0.7222 $\pm$ 0.0351 \\ 10.0000 $\pm$ 1.2649 \\ 0.9480 $\pm$ 0.2790}}} &
\multirow{5}{*}{\rotatebox[origin=c]{90}{\makecell[c]{na \\ na \\ na \\ na \\ na \\ na}}} &
\multirow{5}{*}{\rotatebox[origin=c]{90}{\makecell[c]{0.7673 $\pm$ 0.0655 \\ 0.5200 $\pm$ 0.0267 \\ 0.6197 $\pm$ 0.0406 \\ 0.6160 $\pm$ 0.0480 \\ 9.6000 $\pm$ 1.2000 \\ 1.4244 $\pm$ 0.1953}}} &
\multirow{5}{*}{\rotatebox[origin=c]{90}{\makecell[c]{0.7500 $\pm$ 0.0000 \\ 0.8182 $\pm$ 0.0000 \\ 0.7826 $\pm$ 0.0000 \\ 0.7222 $\pm$ 0.0000 \\ 10.0000 $\pm$ 0.0000 \\ 0.4993 $\pm$ 0.0000}}} &
\multirow{5}{*}{\rotatebox[origin=c]{90}{\makecell[c]{0.7516 $\pm$ 0.0577 \\ 1.0000 $\pm$ 0.0000 \\ 0.8570 $\pm$ 0.0379 \\ 0.8194 $\pm$ 0.0502 \\ 6.0000 $\pm$ 2.7568 \\ 1.6714 $\pm$ 0.6035}}} \\
&&&&&\\
&&&&&\\
&&&&&\\
&&&&&\\
\hline
\multirow{5}{*}{\rotatebox[origin=c]{90}{\textbf{RAG R2-R3}}} & 
\multirow{5}{*}{\rotatebox[origin=c]{90}{\makecell[c]{0.8462 $\pm$ 0.0000 \\ 1.0000 $\pm$ 0.0000 \\ 0.9167 $\pm$ 0.0000 \\ \textcolor{gold}{\textbf{0.8889 $\pm$ 0.0000}} \\ 4.0000 $\pm$ 0.0000 \\ 0.7532 $\pm$ 0.0000}}} &
\multirow{5}{*}{\rotatebox[origin=c]{90}{\makecell[c]{na \\ na \\ na \\ na \\ na \\ na}}} &
\multirow{5}{*}{\rotatebox[origin=c]{90}{\makecell[c]{0.6841 $\pm$ 0.0276 \\ 0.6364 $\pm$ 0.0000 \\ 0.6591 $\pm$ 0.0131 \\ 0.5972 $\pm$ 0.0241 \\ 14.5000 $\pm$ 0.8660 \\ 0.3907 $\pm$ 0.2256}}} &
\multirow{5}{*}{\rotatebox[origin=c]{90}{\makecell[c]{0.6923 $\pm$ 0.0000 \\ 0.8182 $\pm$ 0.0000 \\ 0.7500 $\pm$ 0.0000 \\ 0.6667 $\pm$ 0.0000 \\ 12.0000 $\pm$ 0.0000 \\ 1.0952 $\pm$ 0.0000}}} &
\multirow{5}{*}{\rotatebox[origin=c]{90}{\makecell[c]{0.8221 $\pm$ 0.0396 \\ 0.9846 $\pm$ 0.0308 \\ 0.8955 $\pm$ 0.0300 \\ 0.8800 $\pm$ 0.0358 \\ 3.0000 $\pm$ 0.8944 \\ 0.8589 $\pm$ 0.1588}}} \\
&&&&&\\
&&&&&\\
&&&&&\\
&&&&&\\
\hline
\multirow{5}{*}{\rotatebox[origin=c]{90}{\textbf{RAG R1-R2-R3}}} & 
\multirow{5}{*}{\rotatebox[origin=c]{90}{\makecell[c]{0.8310 $\pm$ 0.0262 \\ 1.0000 $\pm$ 0.0000 \\ 0.9075 $\pm$ 0.0159 \\ \textcolor{gold}{\textbf{0.8750 $\pm$ 0.0241}} \\ 4.5000 $\pm$ 0.8660 \\ 0.8239 $\pm$ 0.1226}}} &
\multirow{5}{*}{\rotatebox[origin=c]{90}{\makecell[c]{na \\ na \\ na \\ na \\ na \\ na}}} &
\multirow{5}{*}{\rotatebox[origin=c]{90}{\makecell[c]{0.7851 $\pm$ 0.0408 \\ 1.0000 $\pm$ 0.0000 \\ 0.8790 $\pm$ 0.0248 \\ 0.8560 $\pm$ 0.0320 \\ 3.6000 $\pm$ 0.8000 \\ 1.0402 $\pm$ 0.1573}}} &
\multirow{5}{*}{\rotatebox[origin=c]{90}{\makecell[c]{0.6923 $\pm$ 0.0000 \\ 0.8182 $\pm$ 0.0000 \\ 0.7500 $\pm$ 0.0000 \\ 0.6667 $\pm$ 0.0000 \\ 12.0000 $\pm$ 0.0000 \\ 1.0952 $\pm$ 0.0000}}} &
\multirow{5}{*}{\rotatebox[origin=c]{90}{\makecell[c]{0.8735 $\pm$ 0.0315 \\ 0.8000 $\pm$ 0.1026 \\ 0.8310 $\pm$ 0.0574 \\ 0.7840 $\pm$ 0.0599 \\ 5.4000 $\pm$ 1.4967 \\ 1.0616 $\pm$ 0.5129}}} \\
&&&&&\\
&&&&&\\
&&&&&\\
&&&&&\\
\hline
\multirow{5}{*}{\rotatebox[origin=c]{90}{\textbf{\makecell[c]{Graph RAG \\ R1-R2}}}} & 
\multirow{5}{*}{\rotatebox[origin=c]{90}{\makecell[c]{0.6970 $\pm$ 0.0000 \\ 0.6970 $\pm$ 0.0000 \\ 0.6970 $\pm$ 0.0000 \\ 0.5918 $\pm$ 0.0000 \\ 20.0000 $\pm$ 0.0000 \\ 0.8267 $\pm$ 0.0000}}} &
\multirow{5}{*}{\rotatebox[origin=c]{90}{\makecell[c]{na \\ na \\ na \\ na \\ na \\ na}}} &
\multirow{5}{*}{\rotatebox[origin=c]{90}{\makecell[c]{0.8067 $\pm$ 0.0133 \\ 0.7000 $\pm$ 0.0667 \\ 0.7485 $\pm$ 0.0424 \\ 0.6500 $\pm$ 0.0500 \\ 5.6000 $\pm$ 0.8000 \\ 0.4782 $\pm$ 0.2391}}} &
\multirow{5}{*}{\rotatebox[origin=c]{90}{\makecell[c]{0.8182 $\pm$ 0.0000 \\ 0.8182 $\pm$ 0.0000 \\ 0.8182 $\pm$ 0.0000 \\ 0.7778 $\pm$ 0.0000 \\ 8.0000 $\pm$ 0.0000 \\ 0.6747 $\pm$ 0.0000}}} &
\multirow{5}{*}{\rotatebox[origin=c]{90}{\makecell[c]{0.6800 $\pm$ 0.1600 \\ 0.5600 $\pm$ 0.0800 \\ 0.5943 $\pm$ 0.0114 \\ 0.5250 $\pm$ 0.0500 \\ 7.6000 $\pm$ 0.8000 \\ 1.3119 $\pm$ 0.5688}}} \\
&&&&&\\
&&&&&\\
&&&&&\\
&&&&&\\
\hline
\multirow{5}{*}{\rotatebox[origin=c]{90}{\textbf{\makecell[c]{Graph RAG \\ R2-R3}}}} & 
\multirow{5}{*}{\rotatebox[origin=c]{90}{\makecell[c]{0.3333 $\pm$ 0.0000 \\ 1.0000 $\pm$ 0.0000 \\ 0.5000 $\pm$ 0.0000 \\ 0.3333 $\pm$ 0.0000 \\ 6.0000 $\pm$ 0.0000 \\ 2.4495 $\pm$ 0.0000}}} &
\multirow{5}{*}{\rotatebox[origin=c]{90}{\makecell[c]{na \\ na \\ na \\ na \\ na \\ na}}} &
\multirow{5}{*}{\rotatebox[origin=c]{90}{\makecell[c]{0.8067 $\pm$ 0.0133 \\ 0.7000 $\pm$ 0.0667 \\ 0.7485 $\pm$ 0.0424 \\ 0.6500 $\pm$ 0.0500 \\ 5.6000 $\pm$ 0.8000 \\ 0.4782 $\pm$ 0.2391}}} &
\multirow{5}{*}{\rotatebox[origin=c]{90}{\makecell[c]{0.8182 $\pm$ 0.0000 \\ 0.8182 $\pm$ 0.0000 \\ 0.8182 $\pm$ 0.0000 \\ 0.7778 $\pm$ 0.0000 \\ 8.0000 $\pm$ 0.0000 \\ 0.6747 $\pm$ 0.0000}}} &
\multirow{5}{*}{\rotatebox[origin=c]{90}{\makecell[c]{na \\ na \\ na \\ na \\ na \\ na}}} \\
&&&&&\\
&&&&&\\
&&&&&\\
&&&&&\\
\hline
\end{tabular} %
}
}
\caption{Performance metrics of power screwdriver for \textit{ii}-aligned across methods and models. Each cell has from left to right (\textbf{Precision}, \textbf{Recall}, \textbf{F1-Score}, \textbf{Accuracy}, \textbf{Edit Distance}, and \textbf{Spectral Distance}). Values in each cell are the mean $\pm$ standard deviation over the repeated experiments. The total of 50 $\times$ 5 $=$ 250 experiments. (na: not available)}
\label{tab:case_ii_aligned_ps_experiments}
\end{table*}

\setlength{\abovecaptionskip}{4pt}
\setlength{\belowcaptionskip}{2pt}
\setlength{\dbltextfloatsep}{4pt plus 1pt minus 1pt}
\setlength{\dblfloatsep}{4pt plus 1pt minus 1pt}
\captionsetup{skip=4pt}

\clearpage
\begin{figure*}[htbp]
 \centering
 \caption{Best two aligned predictions of the models and methods for (\protect\textit{ii})-DSM of power screwdriver (gpt-4-turbo-preview with RAG R2 and deepseek-r1:14b with RAG R2-R3)}
 \label{fig:11}
 \includegraphics[height=0.88\textheight,keepaspectratio]{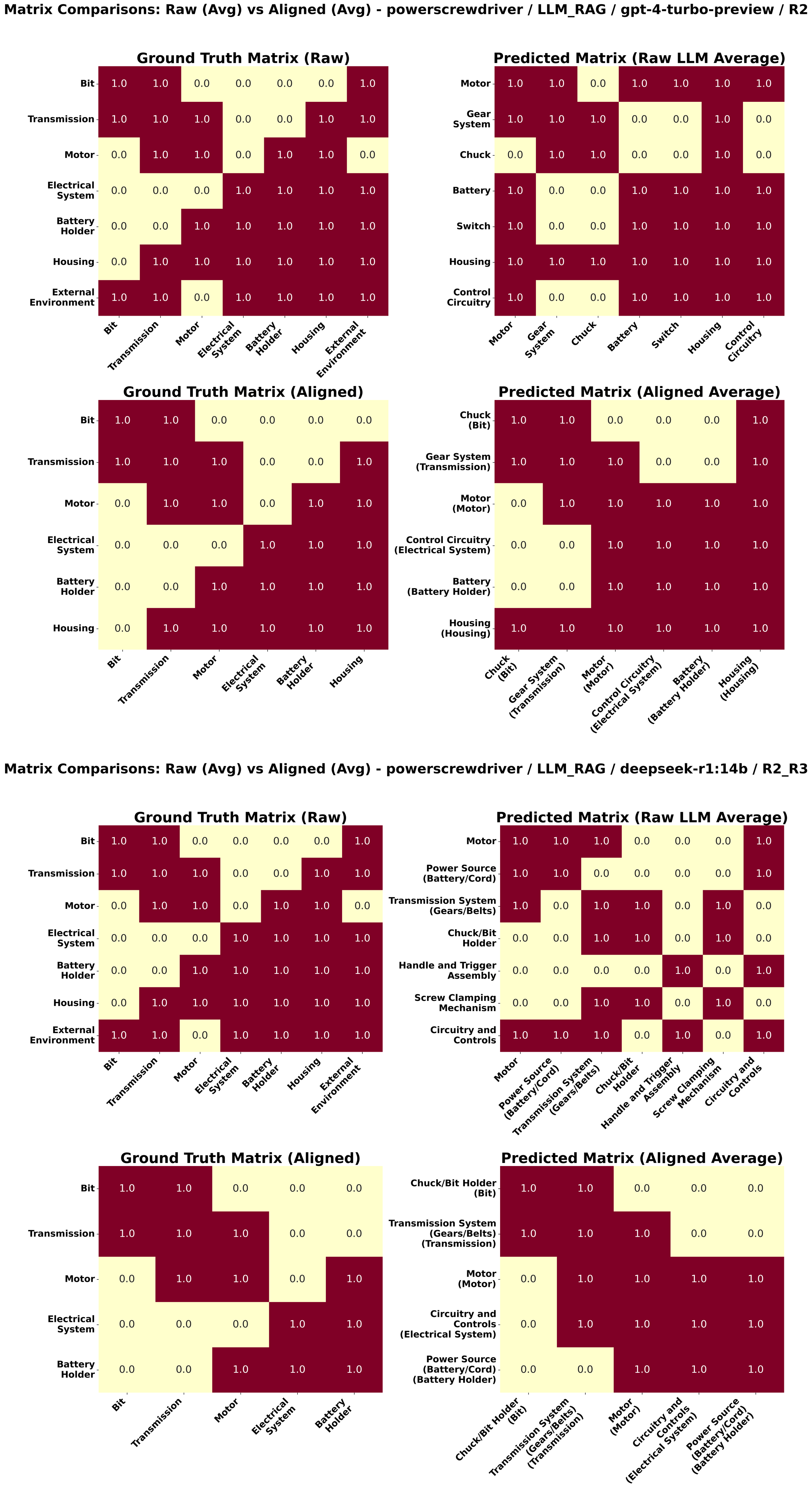}
\end{figure*}

\clearpage
\begin{figure*}[htbp]
 \centering
 \caption{Best two aligned predictions of the models and methods for (\protect\textit{ii})-DSM of CubeSat (llama3.3:70b with RAG R1-R2, the most accurate prediction with five of the six components aligned, and mixtral:8x22b with RAG R2)}
 \label{fig:12}
 \includegraphics[height=0.88\textheight,keepaspectratio]{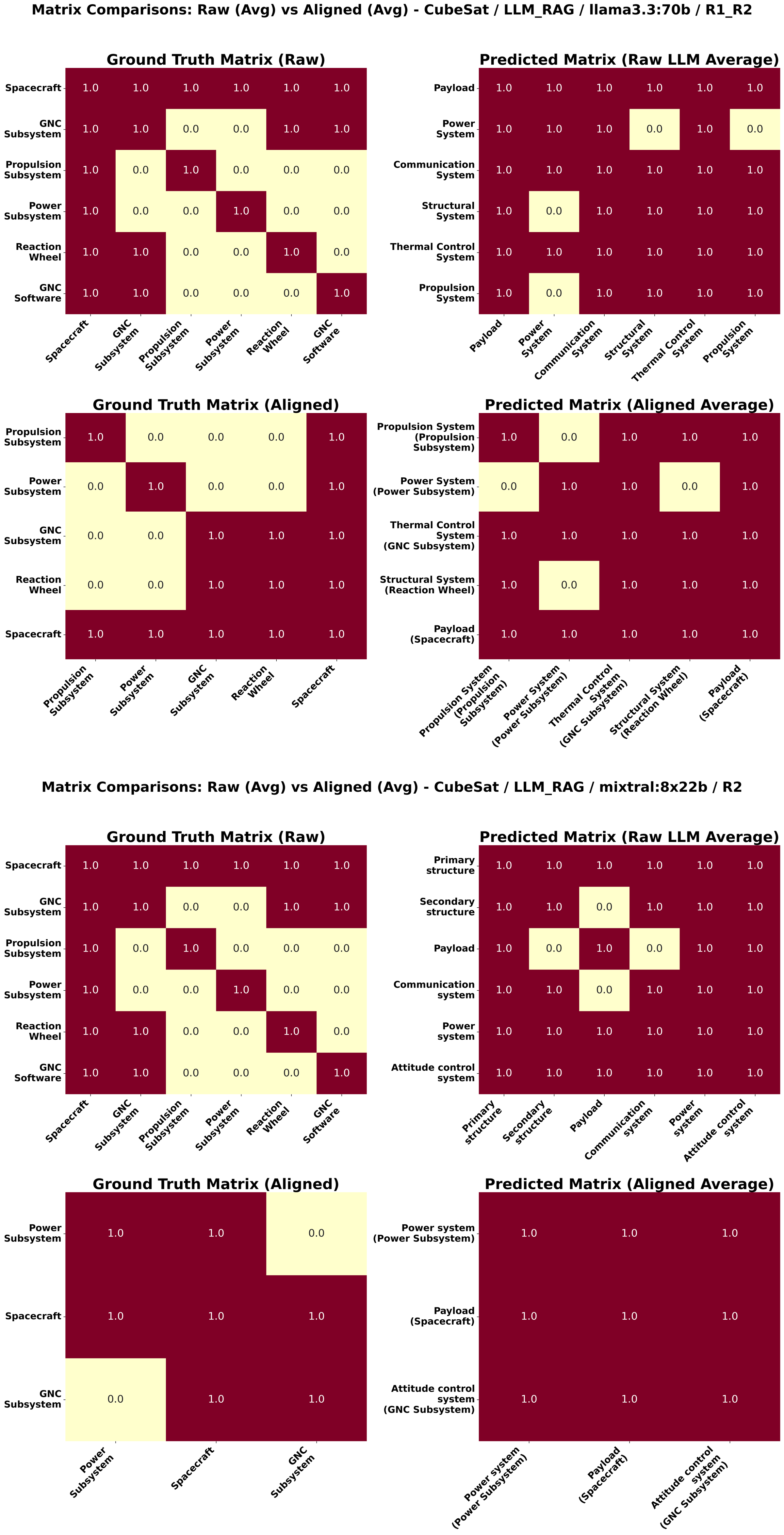}
\end{figure*}

\end{document}